\documentclass[12pt,letterpaper]{article}
\usepackage[top=0.6in,bottom=1.2in,left=0.75in,right=0.75in,footskip=0.75in,marginparwidth=1in]{geometry}
\date{}

\usepackage{url,hyperref,microtype}
\usepackage[onehalfspacing]{setspace}
\usepackage{listings}
\usepackage{booktabs}
\usepackage{tikz}
\usepackage{siunitx}
\usepackage{amsmath}
\usepackage{setspace}
\usepackage[colorinlistoftodos]{todonotes}

\usetikzlibrary{fadings}
\tikzfading[name=fade up,top color=red, bottom color=black]

\usepackage{collcell}
\usepackage{tabularx}
\usepackage{xstring}
\usepackage{xcolor,colortbl}
\usepackage{verbatim}
\usepackage{multirow}
\usepackage{changes}
\usepackage[symbol]{footmisc}

\newcommand{\citet}[1]{\cite{#1}}
\newcommand{\citep}[1]{\cite{#1}}

\newcommand*{\MinNumber}{-0.01}%
\newcommand*{\MidNumber}{0.0} %
\newcommand*{\MaxNumber}{0.035}%

\colorlet{MinColor}{blue} 
\colorlet{MidColor}{white} 
\colorlet{MaxColor}{red} 

\newcommand{\ApplyGradient}[1]{%
        \ifdim #1 pt > \MidNumber pt
            \pgfmathsetmacro{\PercentColor}{int(round(max(min(100.0*(#1 - \MidNumber)/(\MaxNumber-\MidNumber),100.0),0.00)))} %
            \edef\x{\noexpand\cellcolor{MaxColor!\PercentColor!MidColor}}\x#1
        \else
            \pgfmathsetmacro{\PercentColor}{int(round(max(min(100.0*(\MidNumber - #1)/(\MidNumber-\MinNumber),100.0),0.00)))} %
            \edef\x{\noexpand\cellcolor{MinColor!\PercentColor!MidColor}}\x #1
        \fi
}

\newcommand{\ColorIfDecimal}[1]{
	\noexpandarg
	\IfDecimal{#1}{\ApplyGradient{#1}}
	{
		\IfInteger{#1}{\ApplyGradient{#1}}
		{
			#1
		}
	}	
}

\newcolumntype{R}{>{\collectcell\ColorIfDecimal}c<{\endcollectcell}}

\newcommand{\tikzdrawcircle}[2][black,fill=black]{\tikz[baseline=-0.5ex]\draw[#1,radius=#2] (0,0) circle ;}%

\def\xshift{4.45em}
\def\yshift{1.8em}
\definecolor{PHY}{HTML}{0068B4} 
\definecolor{PPROP}{HTML}{137F1A}
\definecolor{FPROP}{HTML}{7F1212}
\definecolor{SENSOR}{HTML}{A5B217}

\tikzstyle{corners}=[rounded corners=0.5em]
\tikzstyle{linewidth} = [line width=0.1em]
\tikzstyle{element} = [draw, fill=PHY, minimum height=\yshift, minimum width=\xshift, text width=0.9*\xshift, corners, align=center, text=white]
\tikzstyle{edge} = [draw, linewidth, -latex]
\tikzstyle{every node}=[font=\footnotesize] 
\tikzstyle{elementDatasetFramework} = [element, minimum height=0.7*\yshift]
\def\alpha{20}

\makeatletter
\renewcommand*{\@fnsymbol}[1]{\ensuremath{\ifcase#1 \or \dagger \or \dagger \or 1 \or 2\or 3 \or 4 \or \|\or **\or \dagger\dagger
		\or \ddagger\ddagger \else\@ctrerr\fi}}
\makeatother

\title{\LARGE \textbf{From Multi-modal Property Dataset \\ to Robot-centric Conceptual Knowledge \\ About Household Objects}}

\author{Madhura Thosar\thanks{Authors have made equal contributions and share first-authorship.}\ \thanks{Institute for Intelligent Cooperating Systems, Faculty of Computer Science, \\ \indent \hspace{0.1cm} Otto-von-Guericke University Magdeburg, Germany}\ , Christian A. Mueller$^{\dagger}$\thanks{Robotics, Computer Science \& Electrical Engineering Department, Jacobs University, Bremen, Germany}\ , Georg Jaeger$^{\dagger}$\thanks{Institute for Computer Science, Technische Universit\"at Bergakademie Freiberg, Germany}\ , Johannes Schleiss$^{1}$, \\
	Narender Pulugu$^{1}$, Ravi Mallikarjun Chennaboina$^{1}$, \\ Sai Vivek Jeevangekar$^{1}$, Andreas Birk$^{2}$,  Max Pfingsthorn\thanks{OFFIS Institute for Information Technology, Oldenburg, Germany}\ , Sebastian Zug$^{3}$
} %

\begin{document}

\maketitle
\begin{abstract}
Tool-use applications in robotics require conceptual knowledge about objects for informed decision making and object interactions. State-of-the-art methods employ hand-crafted symbolic knowledge which is defined from a human perspective and grounded into sensory data afterwards. However, due to different sensing and acting capabilities of robots, their conceptual understanding of objects must be generated from a robot's perspective entirely, which asks for robot-centric conceptual knowledge about objects. 

With this goal in mind, this article motivates that such knowledge should be based on \emph{physical} and \emph{functional} properties of objects. Consequently, a selection of ten properties is defined and corresponding extraction methods are proposed. This multi-modal property extraction forms the basis on which our second contribution, a robot-centric knowledge generation is build on. It employs unsupervised clustering methods to transform numerical property data into symbols, and Bivariate Joint Frequency Distributions and Sample Proportion to generate conceptual knowledge about objects using the robot-centric symbols. 

A preliminary implementation of the proposed framework is employed to acquire a dataset comprising \emph{physical} and \emph{functional} property data of 110 houshold objects. This \textbf{Ro}bot-\textbf{C}entric data\textbf{S}et (RoCS)
is used to evaluate the framework regarding the property extraction methods, the semantics of the considered properties within the dataset and its usefulness in real-world applications such as tool substitution.

\end{abstract}

\section{Motivation}
Knowledge about the world and objects within it are a necessary prerequisite for performing informed actions.
For robotics applications, this knowledge is usually hand-crafted in a variety of ways, e.g. in semantic networks of knowledge bases or implicitly in designing action and state spaces for reinforcement learning.
As an alternative to hand-crafting, we propose a robot-centric data-driven knowledge acquisition framework which allows the robot to gain conceptual knowledge about objects and their properties using its own sensors.
This makes the arduous step of symbol grounding obsolete.

Consider a scenario where a robot has to select between a rock and a plastic bottle such that it can afford a hammering action.
One way of going about it is to interact with each object and perform the action. 
The downside, however, of such interactions is, it is time consuming and it may damage objects in the process that can not afford the action.
\citet{Baber2003_1} postulated that a non-invasive tool selection in humans or animals alike is facilitated by conceptual knowledge about objects, especially, knowledge about their physical and functional properties and relationship between them. 
For instance, knowledge about what physical properties of a hammer enable a hammering action can facilitate the decision between a rock and a plastic bottle.
Conceptual knowledge about objects, in this case, is considered as a common sense understanding of objects \textit{generalized} over our observations and daily interactions with them.
Therefore, based on our observations and interactions with various instances of a cup, a conceptual understanding of a \textit{cup} embodies an object that has a handle, is hollow and can contain hot liquid (c.f. Fig.~\ref{fig:multi_layer}).
Such conceptual knowledge about objects is desired in varies robotic scenarios (from household to industrial robotics) in order to efficiently perform tasks when dealing with objects (e.g., tools) in uncertain environments (e.g., home service, factory of the future, inspection); for instance in a toy scenario, a robot may have to choose between a rock and a plastic bottle as a substitute for a missing hammer.
The challenge, however, is how to acquire such conceptual knowledge.

In order to acquire conceptual knowledge about objects (e.g. in a household environment) which can primarily, but not only be used for tool selection purposes, the following questions need to be answered: \\
1) What kind of knowledge constitutes a conceptual knowledge about objects? 
This question is primarily concerned with the contents of the knowledge and the granularity of the knowledge; \\
2) How to represent the acquired knowledge? 
This is one of the fundamental question in knowledge representation for robotic applications that should be addressed.
It mainly concerns with what representation formalism is suitable for representing the conceptual knowledge;\\
3) How to ground conceptual knowledge in robot's sensory data? 
It is one of the essential challenges that should be addressed when symbolic artificial intelligence in general and knowledge representation methodologies more specifically are employed in robotic applications.
This question primarily deals with creating a bridge between robot's sensory reality of the world and an abstract symbolic conceptual knowledge about the world. \\
These questions forms a basis for three requirements we have targeted in our work, namely: \emph{knowledge contents}, \emph{knowledge representation}, and \emph{robot-centric}.
\begin{enumerate}
    \item \textbf{Knowledge Contents:} One of the key components in successful tool use is the knowledge about physical properties of a tool and their relationship to the tool's various functionalities \citep{RuizSantos2013}.  
    Accordingly, conceptual knowledge about objects is \textit{required} to consist of measurements of their physical and functional properties. %
    Physical properties usually describe the physicality of an object. For instance, shape, size, \emph{hollowness}, \emph{flatness}, or surface \emph{roughness} are appearance related properties of an object that describe its overall structure whereas properties such as \emph{heaviness}, \emph{rigidity}, and strength describe mechanical aspects of an object.
    On the other hand, the functional properties, widely known as affordances, are the functional abilities of an object. For example, an object's ability to contain something, to be moved by something or someone, to support something or someone are such properties.
    It should, however, be noted that a property observed in an object can not be expressed in binary terms.
    For instance, \textit{rigidity} is neither present nor absent in a single object. Instead it is present in various degrees in different objects.
    As a result, in addition to the properties observed in the objects 
    the degree with which a property is observed should also be considered in the conceptual knowledge.

    \item \textbf{Knowledge Representation:} %
    According to \citep{Davis1993}, a representation formalism is a medium where knowledge can be organized such that it allows for efficient reasoning.
    The choice of a specific formalism is driven by a desired outcome and a world in which a robot is operating \citep{Sloman1985}.
    Given the dynamic and uncertain nature of the real world, a formalism will have to deal with the uncertainty caused by large number of factors.
    
    Our focus is especially on the uncertainty which is characterized by the diversity notable within physicalities of the instances of any object.
    Therefore, it is prudent that a formalism is \textit{capable} of representing such diversity.
    Moreover,  as knowledge is to be represented as a set of generalized observations regarding the properties reflected in the instances of an object, a formalism should be able to incorporate the degree with which a property is observed in an object.

    \item \textbf{Robot-Centric:} In the context of this work, the term robot-centric knowledge refers to the knowledge that is generated on the basis of \textit{What I sense is What I know}.
    Traditionally, conceptual knowledge is portrayed by a symbolic representation formalism. %
    While such symbolic representations are effective for performing abstract reasoning or plan generation, the sensory-motor processes enable a robot to perceive the world and act on it.
    The symbol grounding process bridges the gap between symbolic and non-symbolic level by creating a correspondence between them.
    This correspondence either refers to a physical entity in the real-world or assigns a meaning to a symbol by means of a respective sensory-motor process \citep{Coradeschi2003, Harnad1990}. Therefore, it is \textit{paramount} that the symbols in the conceptual knowledge used to represent objects and properties should be grounded in the robot's perception of the world.
    
    We focus on grounding symbols by construction through their acquisition based on robot sensor data.
    
\end{enumerate} 

\subsection{Related Work in Knowledge Bases}
\label{subsec:related_work_KB}
Since the demand for knowledge bases has been increasing in robotic applications, the development of knowledge bases has been undertaken by the researchers around the world.
While there exists a multitude of knowledge bases, the question is how many existing knowledge bases about objects conform to the above mentioned requirements?

In \citet{Thosar2018}, we reviewed existing knowledge bases and their underlying acquisition system developed for service robotics to address this question.
We selected 20 papers covering 9 knowledge bases about household objects on the basis of the contents of the paper with respect to the above mentioned requirements 
and overall impact of the paper on the basis of the number of citations (refer Table \ref{tab:ListOfPapers}). 
Our review resulted in the following conclusions with respect to each requirement:

\begin{table*}[tb]
  \caption{List of selected knowledge bases and their names originally appeared in \citep{Thosar2018}}
  \label{tab:ListOfPapers}
  \centering
  \begin{tabular}{p{10.1cm}p{6.7cm}}
  & \\
   \textbf{Knowledge Base} &  \textbf{Acronym} \\
   \hline
  \small Knowledge processing system for Robots & \small KNOWROB \citep{Tenorth} \\
  \small Knowledge Base using Markov Logic Network & \small MLN-KB \citep{Zhu2014} \\
  \small Non-Monotonic Knowledge-Base & \small NMKB \citep{Pineda2017} \\
  \small Open Mind Indoor Common Sense & \small OMICS\citep{Gupta2004} \\
  \small Ontology-based Multi-layered Robot Knowledge Framework & \small OMRKF \citep{Suh2007} \\
  \small OpenRobots Ontology & \small ORO \citep{Lemaignan2010} \\
  \small Ontology-based Unified Robot Knowledge & \small OUR-K \citep{Lim2011} \\
  \small Physically Embedded Intelligent Systems & \small PEIS \citep{Daoutis2009} \\
  \small Knowledge Engine for Robots & \small RoboBrain \citep{Saxena2014} \\
  \end{tabular}
  \end{table*}

\noindent \textbf{Knowledge Contents:} We noted that the majority of the knowledge bases relied on the external human-made commonsense knowledge bases such as ConceptNet \citep{Liu:2004:CMP:1031314.1031373}, WordNet \citep{Fellbaum1998} (KnowRob, MLN-KB, OMICS, RoboBrain), Cyc \citep{Lenat1995} (PEIS-KB), OpenCyc \citep{Lenat1995} (KnowRob, ORO, RoboBrain) and the rest either relied on the hand-coded knowledge (OMRKF, OUR-K) or on knowledge acquired by human-robot interaction (NMKB), for the symbolic conceptual knowledge about objects.
As the required primary contents of the conceptual knowledge are the properties and degree with which they are observed in the objects, the external human-made knowledge bases did not include this requirement.
For instance, a \textit{cup} is described in WordNet as \texttt{a small open container usually used for drinking; usually has a handle}.
This description, however, does not mention how  hollow a cup \textit{usually} is or how rigid it may be, if at all.

\noindent \textbf{Knowledge Representation: } Logic based representation formalisms were overwhelmingly used by majority of the knowledge bases: OWL-RDF (KnowRob, OMRKF, ORO, OUR-K), Markov Logic Network (MLN-KB), Prolog - Horn Clause (NMKB), Second Order Predicate Logic (PEIS), while database inspired formalisms were used by RoboBrain (Graph Database) and OMICS (Relational Database).
However, none of the formalisms modeled the degree with which a property is reflected in an object.

Various techniques were deployed to represent uncertainty, such as probabilistic models, statistical relational models or Bayesian inference. 
The uncertainty  was used to characterize noisy sensor information, incomplete knowledge, unknown objects or environment, and inconsistent knowledge.
While all the above uncertainty factors are significant, the knowledge bases did not take into account uncertainty caused by the intra-class variations in the observed properties within the same object class.
For instance, when we think of a cup, although at the abstract level, it is a type of container, the degree of containability is different in a cup for expresso coffee and a cup for tea.
Such variation in the containability is not reflected in the knowledge base.

\noindent \textbf{Robot-Centric:} Almost all of the knowledge bases (except for OMICS) addressed the problem of symbol grounding.
While the object labels, appearance related properties (shape, size etc.), and functional properties (KnowRob, MLN-KB, NMKB, PEIS) were grounded in the robot's perception, the reliance on human-made knowledge did pose a disadvantage.
Since the commonsense knowledge bases are fully human-made, the depth and breadth of the knowledge
is not perceivable by a robot due to its limited perception
and manipulation capabilities. 
As a consequence, while the low portion of human-centric knowledge is grounded into robot's limited perception, majority of the portion of the knowledge base remain non-grounded.

\subsection{Contribution}
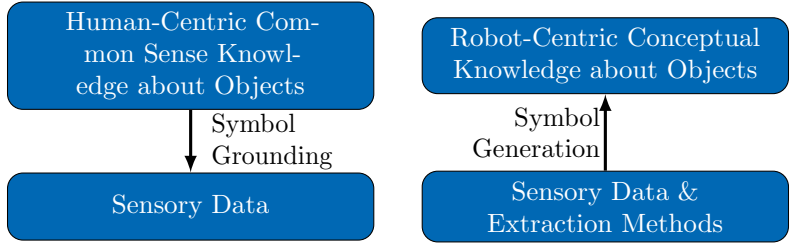
\begin{figure}[tb]
   \centering
   \scalebox{1.0}{
\begin{tikzpicture}

\node[element,text width=2.5*\xshift, minimum height=1.2*\yshift] (COMSENSOR) at (0,0) {Sensory Data};
\node[element,text width=2.5*\xshift, minimum height=1.2*\yshift] (COMKNOWLEDGE) at (0, 2.7*\yshift) {Human-Centric Common Sense Knowledge about Objects};

\node[element,text width=2.5*\xshift, minimum height=1.2*\yshift] (RCSENSOR) at (3*\xshift, 0.0*\yshift) {Sensory Data \& \\Extraction Methods};
\node[element,text width=2.5*\xshift, minimum height=1.2*\yshift] (RCCON) at (3*\xshift, 2.7*\yshift) {Robot-Centric Conceptual Knowledge about Objects};

\path[edge] (COMKNOWLEDGE) edge node[xshift=0.4*\xshift, text width=0.5*\xshift] {Symbol \\Grounding} (COMSENSOR);

\path[edge] (RCSENSOR) edge node[xshift=-0.5*\xshift, align=right] {Symbol \\Generation} (RCCON);

\end{tikzpicture}
     }
   \caption{Symbol grounding approach comparison: the typical approach vs. proposed approach to grounding the knowledge}
  \label{fig:grounding_approach}
\end{figure}

The perception and manipulation capabilities of robots and humans are different. Consequently, the representation of an object and its conceptual understanding differs from human to robot. But in contrast to this fact, many approaches intent to merge both worlds by mapping human-centric, symbolic knowledge on sensory data (Fig.~\ref{fig:grounding_approach} -- left side).
In contrast, we propose a bottom-up approach to symbol grounding, a robot-centric symbol generation approach, see right side of Fig.~\ref{fig:grounding_approach}.  

\noindent \textbf{Contribution 1) Multi-modal property extraction:} 
Our primary objective is to generate robot-centric conceptual knowledge about the properties of objects that is grounded in the robot's sensory data.
In order to realize a bottom-up approach to generate symbolic knowledge, we propose an extensible property extraction framework wherein multiple property extraction methods can be used to measure various physical properties. These include appearance based properties and mechanical properties.
Based on these, functional properties are acquired on the basis of the acquired physical properties. 

\begin{figure}[tb]
   \centering
   \scalebox{1.1}{
\includegraphics[width=0.7\linewidth]{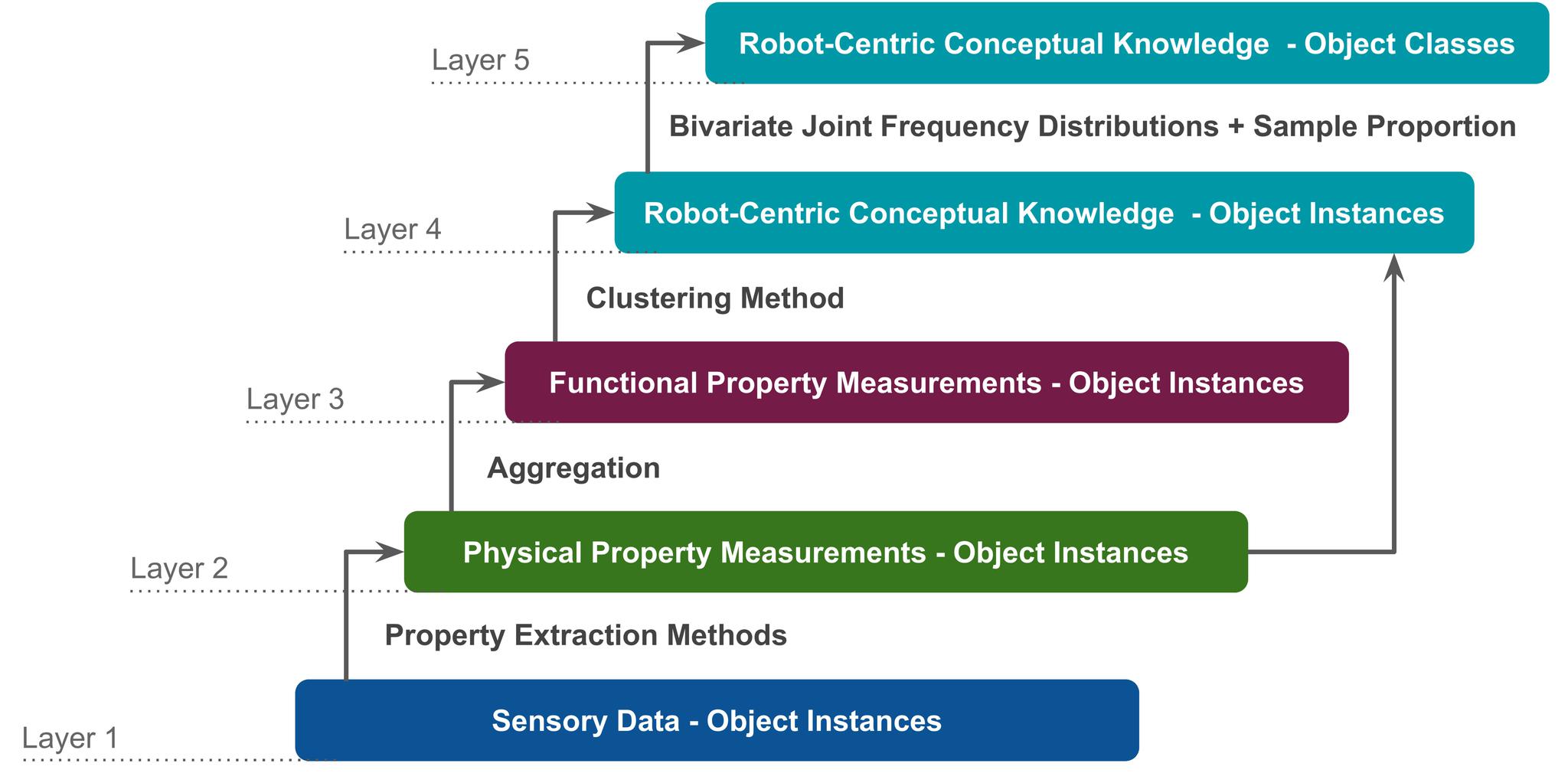}
}
\caption{The contribution of ROCS can be separated on two layers while considering the whole processing concept. The first one is focused on the extraction and aggregation of property data (Layer 1 - 3). The second contribution addresses the following Layers 4 and 5.}
  \label{fig:multi_layer}
\end{figure}

Moreover, our framework currently incorporates various light-weight physical property extraction methods which require minimalistic experimental set-ups. 
The obtained property data is then used to generate robot-centric symbolic knowledge. 
Consequently, the proposed framework can be used to create a multi-layered dataset and knowledge base about household objects where the layers denote the different levels of abstraction (Fig. \ref{fig:multi_layer}).

\noindent \textbf{Contribution 2) Robot-centric knowledge generation:} The robot-centric knowledge generation approach is proposed where a clustering method is used to generate the symbols to represent the qualitative measures or the degree with which a property is observed in an object. 
Instead of pre-defining the symbols and grounding them afterwards, we use sensory data and robot-centric extraction methods to generate qualitative data for each object property.
The statistical methods Bivariate Joint Frequency Distribution and Sample Proportion are used to model the intra-class variations in an object class (Fig. \ref{fig:multi_layer}). 
At the end, an \textit{Attribute-Value Pair} based formalism is used to represent conceptual knowledge about objects. 
The primary reason to use \textit{Attribute-Value Pair} based formalism is to allow the developers to adapt a representation language that is best suitable for their approach as opposed to adapt an approach according to the representation language of the knowledge base.

\section{Robot-Centric Property Acquisition -- \\ A Conceptual Framework}
\label{sec:framework}

At its core, the idea of the proposed framework is to enable a robot to abstract conceptual knowledge about objects from its sensory data such that the generated conceptual knowledge is inherently grounded into the robot's perception. 
Central for generating such robot-centric knowledge are the (physical and functional) properties used by the robot to describe objects. 

In this article we will discuss the underlying building blocks to realize the extraction framework where the discussion will primarily focus on the property definitions and their extraction methods. 
The notion of the physical and functional properties is inspired by the following literature on the tool use in humans and animals: \citep{Baber2003}, \citep{Vaesen2012}, \citep{Biro2013}, \citep{Vauclair1994}, \citep{Susi2005}, \citep{Hernik2009}, and \citep{Sanz2013}.
The research work done in this area on properties is considered in our work.

One of the pressing issues, however, is interpreting the meaning of the properties.
The meaning can be complex where various facets of a property and their relationship to the various parts of an object are perceived and interpreted accordingly or it can be primitive or simplistic.
In either case, the meaning or definition of a property forms a basis for designing a hardware set-up and a subsequent extraction method. 
In this work, when interpreting the properties, simplistic definitions of the properties were formed which allowed for a minimalistic set-up and a light-weight extraction method. 
The intend behind a simplistic approach is that it allows the use of a simple mobile manipulator whose limited capabilities can be exploited.

In the following section, we first discuss their selection, definitions, and methods for acquiring numeric representations of these object properties for a robotic platform. 
Given the acquired object properties, we propose a robot-centric conceptual knowledge generation approach in Section~\ref{sec:knowledge_generation}.
Section~\ref{sec:eval} discusses the extraction of the dataset from 110 household objects using the proposed framework (Section~\ref{sec:eval:ROCS_dataset}) followed by evaluation of the quality of the acquired dataset (Section~\ref{sec:eval:property_extraction_methods}), the semantics of property measurements (Section~\ref{sec:eval:property_semantics}), and knowledge base generation and its application in tool substitution scenario (Section~\ref{sec:eval:tool_substitution}).

\subsection{Property Acquisition}
\label{subsec:property_acquisition}
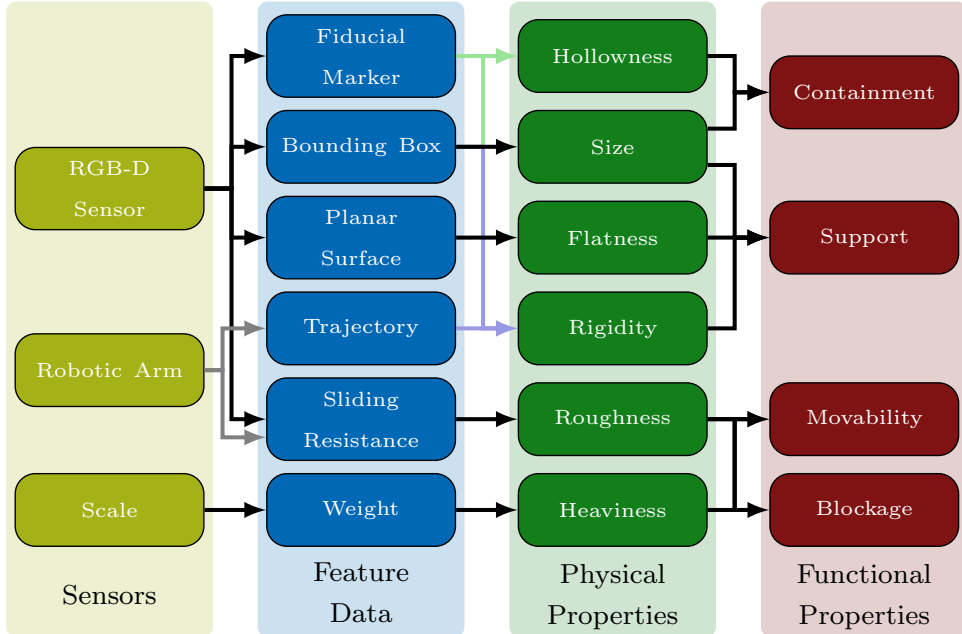
\begin{figure}[tb]
\centering
\scalebox{1.3}{
\def\xfactor{1.4}
\definecolor{blue2}{RGB}{50,50,200}
\definecolor{green2}{RGB}{50,200,50}
\definecolor{red2}{RGB}{200,100,100}

\begin{tikzpicture}

\node[fill=SENSOR!20, minimum height=8.75*\yshift, minimum width=1.15*\xshift,rounded corners=0.25em,label={[yshift=-8.5*\yshift, text width=1.15*\xshift, align=center]\scriptsize Sensors}] (FEATURE) at (-1*\xfactor*\xshift, -3.625*\yshift) {};

\node[fill=PHY!20, minimum height=8.75*\yshift, minimum width=1.15*\xshift,rounded corners=0.25em,label={[yshift=-8.75*\yshift, text width=1.15*\xshift, align=center]\scriptsize Feature\\Data}] (FEATURE) at (0*\xfactor*\xshift, -3.625*\yshift) {};

\node[fill=PPROP!20, minimum height=8.75*\yshift, minimum width=1.15*\xshift,rounded corners=0.25em,label={[yshift=-8.85*\yshift, text width=1.15*\xshift, align=center]\scriptsize Physical\\ Properties}] (PHYPROP) at (1*\xfactor*\xshift, -3.625*\yshift) {};

\node[fill=FPROP!20, minimum height=8.75*\yshift, minimum width=1.15*\xshift,rounded corners=0.25em,label={[yshift=-8.85*\yshift, text width=1.15*\xshift, align=center]\scriptsize Functional \\Properties}] (FUNPROP) at (2*\xfactor*\xshift, -3.625*\yshift) {};

\node[element, fill=SENSOR] (SEN1) at (-1*\xfactor*\xshift,-1.825*\yshift) {\tiny RGB-D Sensor};
\node[element, fill=SENSOR] (SEN2) at (-1*\xfactor*\xshift,-4.325*\yshift) {\tiny Robotic Arm};
\node[element, fill=SENSOR] (SEN3) at (-1*\xfactor*\xshift,-6.25*\yshift) {\tiny Scale};

\node[element, fill=PHY] (PHY1) at (0,0) {\tiny Fiducial Marker};
\node[element, fill=PHY] (PHY2) at (0,-1.25*\yshift) {\tiny Bounding Box};
\node[element, fill=PHY] (PHY3) at (0,-2.5*\yshift) { \tiny Planar Surface};
\node[element, fill=PHY] (PHY4) at (0,-3.75*\yshift) {\tiny Trajectory}; 
\node[element, fill=PHY] (PHY5) at (0,-5*\yshift) {\tiny Sliding Resistance};
\node[element, fill=PHY] (PHY6) at (0,-6.25*\yshift) {\tiny Weight};

\node[element, fill=PPROP] (PPROP1) at (1*\xfactor*\xshift,0) {\tiny Hollowness}; 
\node[element, fill=PPROP] (PPROP2) at (1*\xfactor*\xshift,-1.25*\yshift) {\tiny Size};
\node[element, fill=PPROP] (PPROP3) at (1*\xfactor*\xshift,-2.5*\yshift) {\tiny Flatness};
\node[element, fill=PPROP] (PPROP4) at (1*\xfactor*\xshift,-3.75*\yshift) {\tiny Rigidity};
\node[element, fill=PPROP] (PPROP5) at (1*\xfactor*\xshift,-5*\yshift) {\tiny Roughness};
\node[element, fill=PPROP] (PPROP6) at (1*\xfactor*\xshift,-6.25*\yshift) {\tiny Heaviness};

\node[element, fill=FPROP] (FPROP1) at (2*\xfactor*\xshift,-0.5*\yshift) {\tiny Containment}; 
\node[element, fill=FPROP] (FPROP2) at (2*\xfactor*\xshift,-2.5*\yshift) {\tiny Support};
\node[element, fill=FPROP] (FPROP3) at (2*\xfactor*\xshift,-5*\yshift) {\tiny Movability};
\node[element, fill=FPROP] (FPROP4) at (2*\xfactor*\xshift,-6.25*\yshift) {\tiny Blockage};

\path[edge,draw=black!100] (SEN1.east) -- ++(0.15*\xshift,0)|- (PHY1.west);
\path[edge,draw=black!100] (SEN1.east) -- ++(0.15*\xshift,0)|- (PHY2.west);
\path[edge,draw=black!100] (SEN1.east) -- ++(0.15*\xshift,0)|- (PHY3.west);
\path[edge,draw=black!100] (SEN1.east) -- ++(0.15*\xshift,0)|- (PHY5.west);

\path[edge,draw=black!50] (SEN2.east) -- ++(0.1*\xshift,0)|- (PHY4.west);
\path[edge,draw=black!50] (SEN2.east) -- ++(0.1*\xshift,0)|- ([yshift=-0.25*\yshift]PHY5.west);

\path[edge,draw=black] (SEN3.east) -- ++(0.15*\xshift,0)|- (PHY6.west);

\path[edge,draw=green2!50] (PHY1) -> (PPROP1);
\path[edge,draw=green2!50] (PHY2.east) -- ++(0.15*\xshift,0)|- (PPROP1.west);

\path[edge,draw=blue2!50] (PHY2.east) -- ++(0.15*\xshift,0)|- (PPROP4.west);
\path[edge] (PHY2) -> (PPROP2);

\path[edge] (PHY3) -> (PPROP3);
\path[edge,draw=blue2!50] (PHY4) -> (PPROP4);
\path[edge] (PHY5) -> (PPROP5);
\path[edge] (PHY6) -> (PPROP6);

\path[edge] (PPROP1.east) -- ++(0.15*\xshift,0)|- (FPROP1.west);
\path[edge] ([yshift=0.25*\yshift]PPROP2.east) -- ++(0.15*\xshift,0)|- (FPROP1.west);

\path[edge] ([yshift=-0.25*\yshift]PPROP2.east) -- ++(0.15*\xshift,0)|- (FPROP2.west);
\path[edge] (PPROP3.east) -- ++(0.15*\xshift,0)|- (FPROP2.west);
\path[edge] (PPROP4.east) -- ++(0.15*\xshift,0)|- (FPROP2.west);

\path[edge] (PPROP5.east) -- ++(0.15*\xshift,0)|- (FPROP3.west);
\path[edge] (PPROP5.east) -- ++(0.15*\xshift,0)|- (FPROP4.west);
\path[edge] (PPROP6.east) -- ++(0.15*\xshift,0)|- (FPROP3.west);

\end{tikzpicture}
      }
\caption{Proposed property hierarchy and their dependencies (arrow colors chosen to visually distinguish dependencies). }%
\label{fig:property_hierarchy} 
\end{figure}

Our understanding of "robot-centric" considers two aspects. As already mentioned, we have to distinguish between human conceptual knowledge and machine perspective. Secondly, the wording references to the specific capabilities of a robot determining its individual, abstract understanding of a certain object. In this way, the methods for sensory data generation and property extraction varies for different robot configurations. On the other hand, the set of required properties depends on the actual tasks and environmental conditions. 
In order to cover a broad bandwidth of applications and scenarios, we selected ten properties determining the capability of an object to substitute a tool in general.
These ten properties are hierarchically organized as illustrated in Fig.~\ref{fig:property_hierarchy}.
Therein, we mainly distinguish between \emph{physical} (\emph{hollowness}, \emph{size}, \emph{flatness}, \emph{rigidity}, \emph{roughness}, \emph{heaviness}) and \emph{functional} (\emph{containment}, \emph{support}, \emph{movability}, \emph{blockage}) properties.
These six \emph{physical} properties  form the basis from which the remaining four \emph{functional} properties emerge.
For the sake of completeness in Fig.~\ref{fig:property_hierarchy} \emph{sensory} and \emph{feature data} are introduced prior to properties as these initial abstraction levels are used to infer the characteristics of the high-level properties.

\begin{figure}[tb]
	\centering
	\def \figh {3.75cm}
\begin{tabular}{cccc}
\includegraphics[height=\figh]{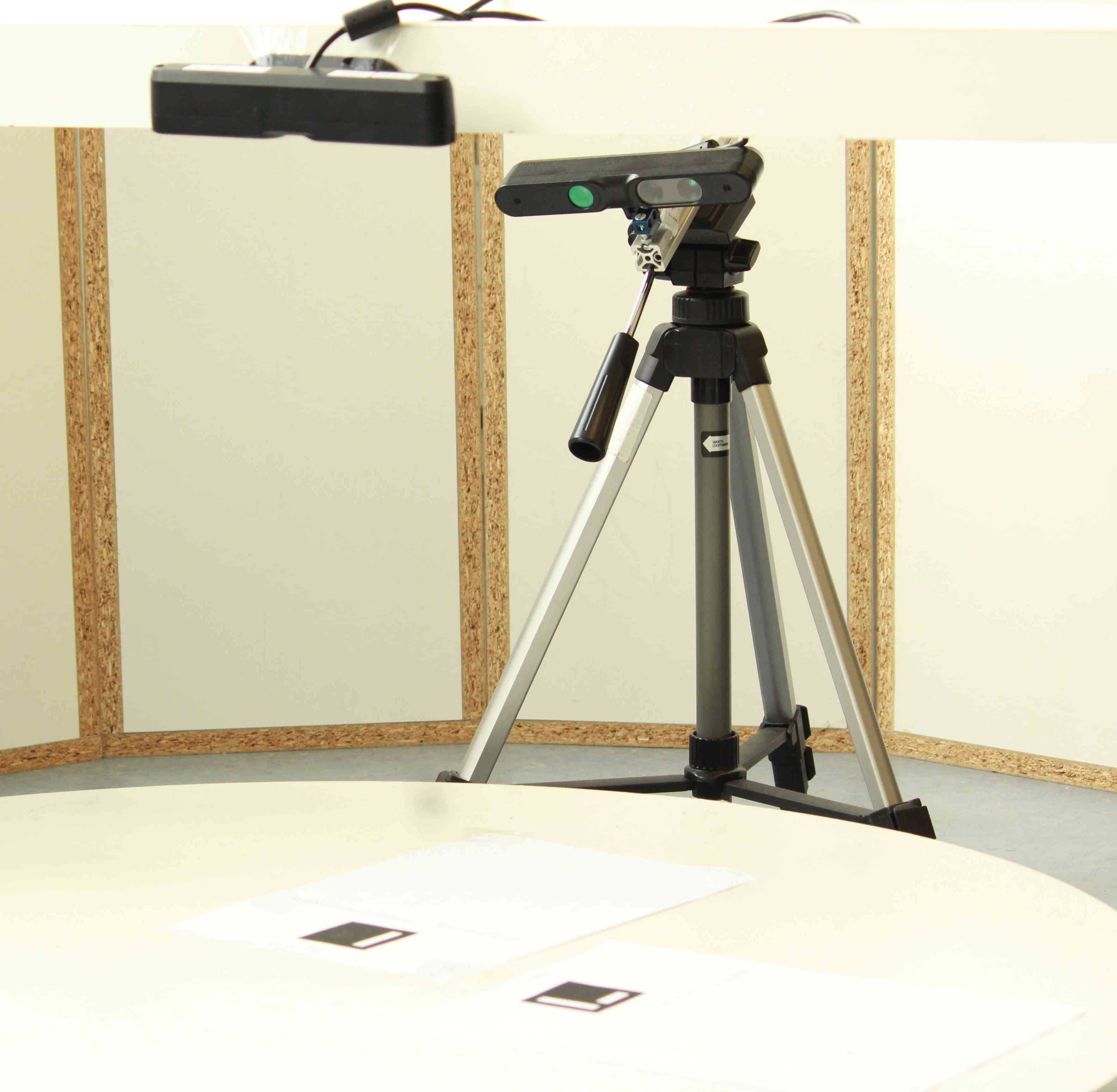} & \includegraphics[height=\figh]{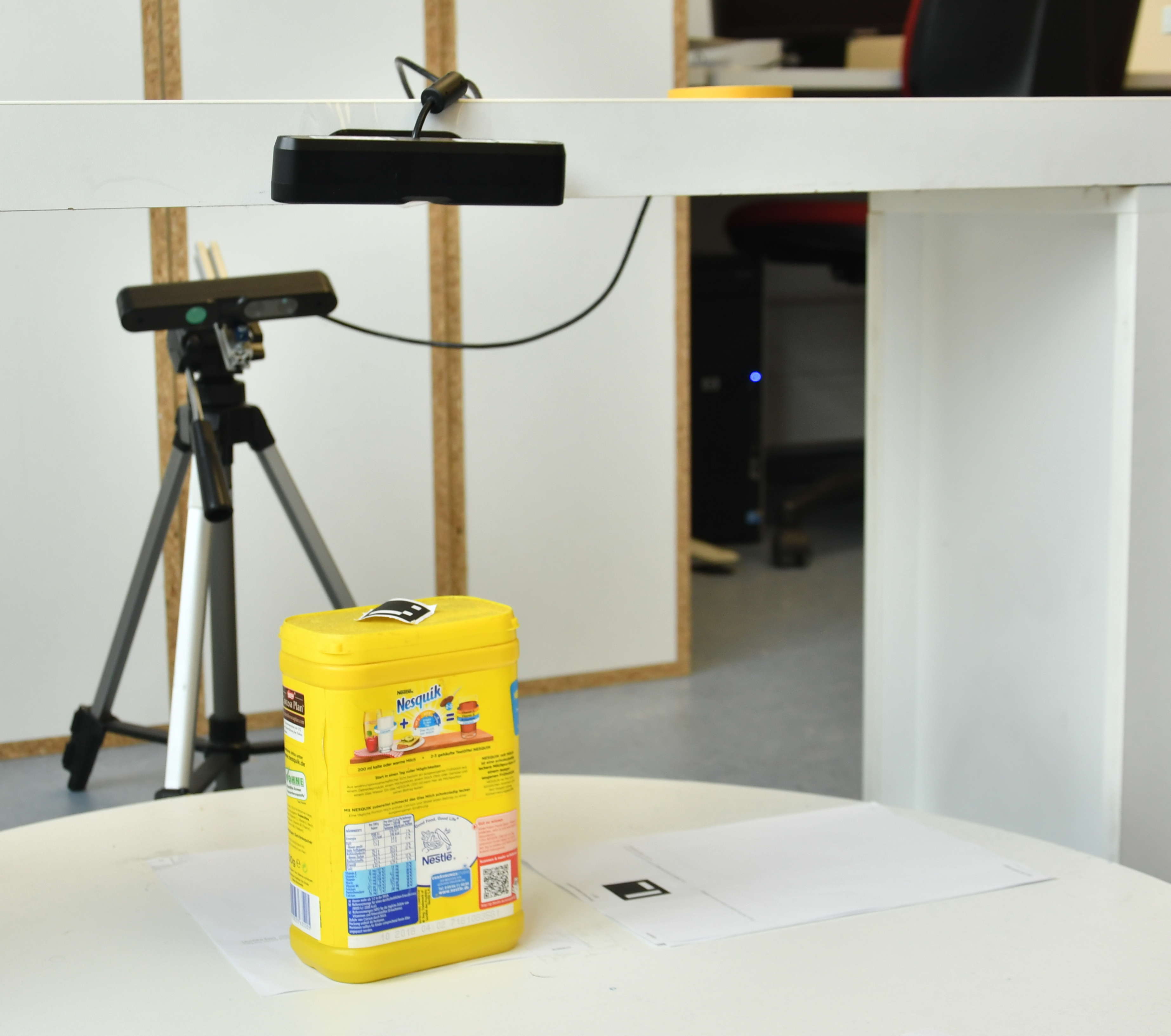} & \includegraphics[height=\figh]{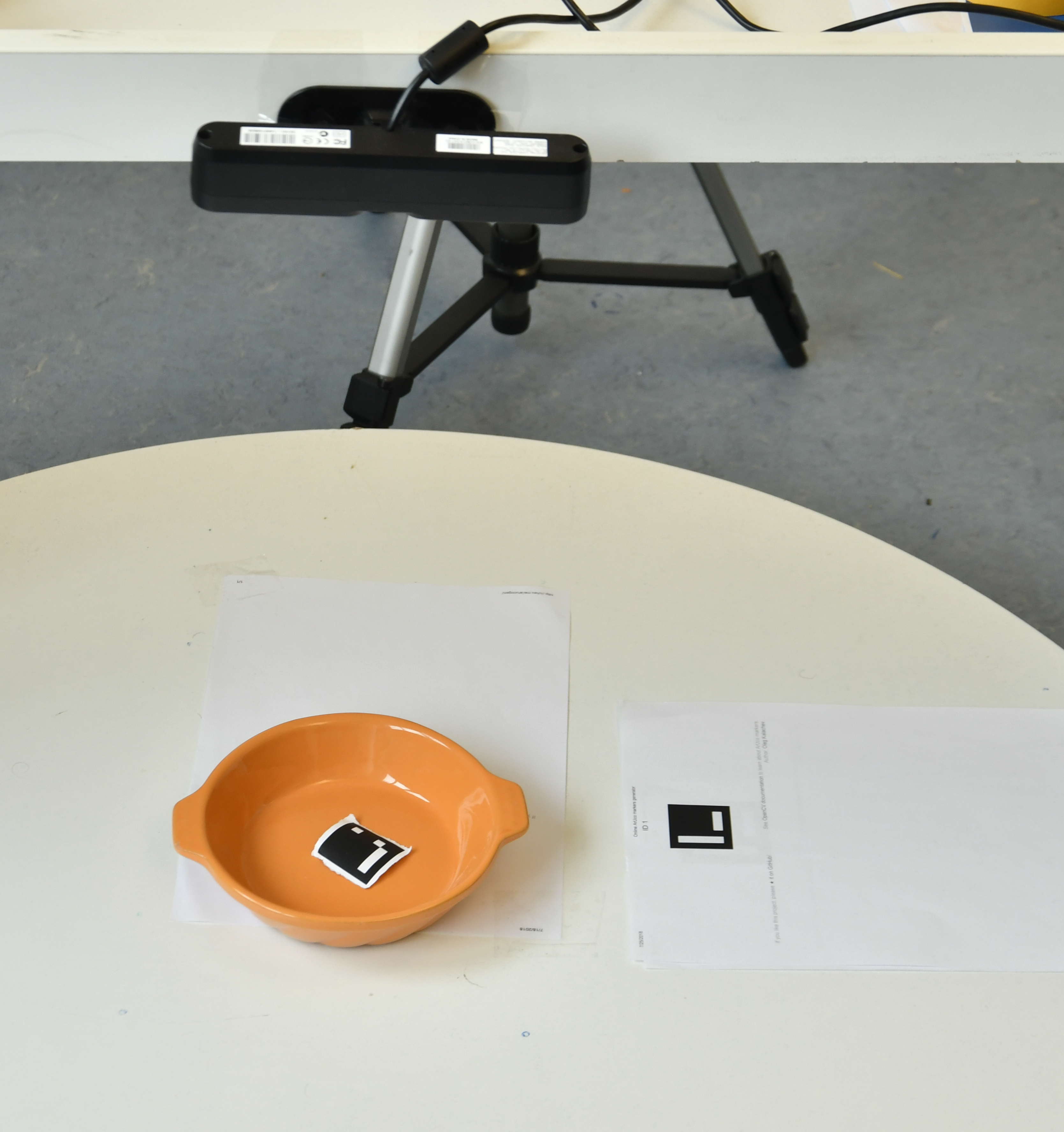} & \includegraphics[height=\figh]{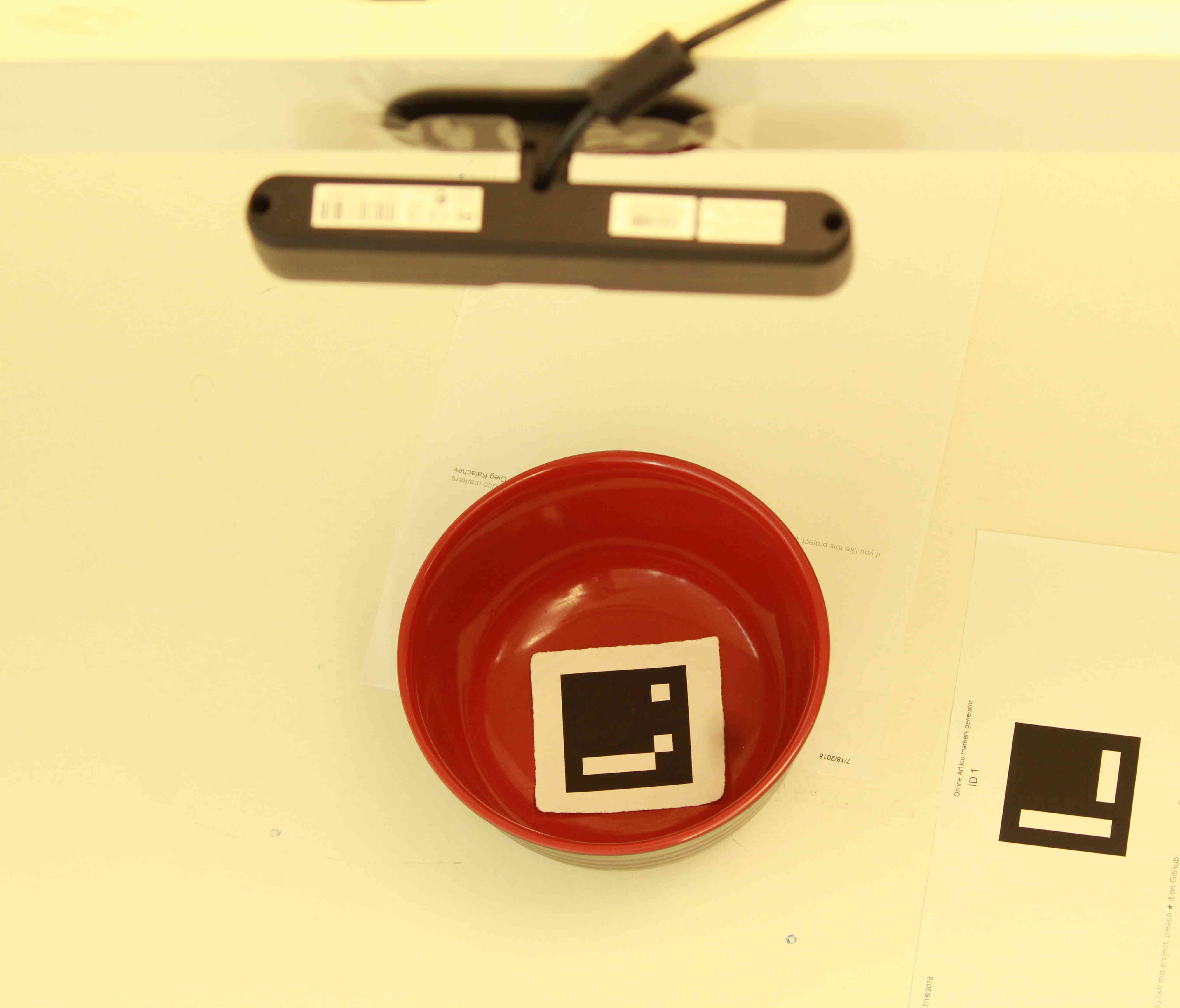}\\
\end{tabular}
	\caption{Light-weight experimental setup consisting of a two cameras and fiducial markers~\citep{GarridoJurado2014}, for acquiring physical properties.}
	\label{fig:setup:physical_property}
\end{figure}

In the following, each property is described in a two-fold manner.
First, for each property a \emph{definition} is provided in a generic manner, i.e. we aim for a \emph{simplistic} and \emph{intuitive} characterization for each property.
Second, for each property an \emph{extraction} method is proposed.
Note that, although the property definitions are formulated from a human perspective, we aim at enabling a robot to assemble its own understanding about objects.
Hence, we have derived extraction methods allowing a robot to interpret its \emph{sensory data}~(Fig.~\ref{fig:property_hierarchy}) for generating numeric representations of \emph{physical} and \emph{functional} properties. 
While the presented methods consider features of our robotic platform (Kuka youBot~\citep{Bischoff2011} (Fig.~\ref{fig:setup:physical_property_rigidity_cam_robot}) and a Asus Xtion Pro depth sensor~\citep{Swoboda2014}, we aim to propose a light-weight setup (Fig.~\ref{fig:setup:physical_property}) and methods that are transferable and adoptable to other robotic platforms by considering common hardware interfaces and data representations such as images, point clouds or joint states of robotic manipulators.

\subsection{Physical Properties}
\label{subsec:physical_properties}

\textit{Humans tend to conceptualize tools in terms of their function, i.e., the outcome that a given kind of artifact, due to its designed physical structure, helps to bring about when used in goal-directed actions} \citep{Hernik2009}.

In other words, in order to enable any functionality in an object, a certain assemblage of physical properties are essential prerequisites \citep{Baber2003_5}.
For humans, the first step towards understanding this causal relationship is by assessing the physical properties of an object and examining the functionalities enabled by them \citep{Hernik2009}.
In this work, we have selected \textit{flatness}, \textit{hollowness}, \textit{size}, \textit{roughness}, \textit{rigidity} and \textit{heaviness} as physical properties given their significance reported in the literature on tool use in humans and animals.
The main inspiration behind selecting these properties was the prominent roles these properties played in various tool use scenarios in humans and animals alike as widely reported in the literature to demonstrate the tool-use abilities.
For instance, the human infants begin exploring their abilities to use any object by studying and interacting with it to understand its \textit{weight}, \textit{texture}, and \textit{shape} \citep{Vauclair1994}. 
While designing and manufacturing a tool, humans and animals alike pay closer attention to the properties such as \emph{shape}, \emph{size}, \emph{rigidity}, \emph{roughness}, and \emph{heaviness} \citep{Baber2003_6}.
It has been observed that wild animals select the tools based on the \textit{size}, \textit{shape} or mechanical properties such as \textit{strength}, \textit{hardness} \citep{Biro2013}.
For example, otters have been observed carrying \textit{flat} rock on their chest which they use to break the shellfish \citep{Emery2013_4}.
On the other hand, researchers found that the monkeys are able to select the hardness of the stone with respect to the hardness of the nut they want to cut open \citep{Boesch2013_2}.

In the following, we provide a \emph{definition} for each physical property and subsequently an \emph{extraction} method is proposed for each property.
Note that, across all extraction methods, we assume that an object is placed in its most natural position, for instance, a cup is most commonly placed in such a way, that its opening points upwards.
Furthermore, we aim for a bounded property value, i.e. an extracted property value is bounded $[0,1]$ in order to enable a subsequent unbiased property analysis that is not affected by object-specific characteristics or scales.
Note that, as a prerequisite, each object is segmented a priori through  \emph{table-top} object segmentation procedure, particularly for the \emph{size}, \emph{flatness} and \emph{hollowness} property. 

\subsubsection{Size Property}
\label{sec:property_size}

\noindent \textbf{Definition:} \emph{Size} of an object is defined intuitively by the object's spatial dimensionality in form of \emph{length}, \emph{width} and \emph{height}.

\noindent \textbf{Extraction:}
The size of an object is defined by the length, width and height. As it therefore can be estimated by determining an object's bounding box, we use an RGB-D sensor to obtain point clouds of the object from an lateral perspective. 
Using marker detection to define a region of interest (ROI), we segment the object and transform its point cloud to an axis-normal representation, i.e. the z-axis is aligned with the object's height. 
Subsequently, an axis-aligned bounding box is approximated given the extracted object point cloud. 
The $\text{size}\mathrm{=}[\text{length},\text{width},\text{height}]$ of an object is directly derived from the object point cloud as distances between the minimal and maximal value in each spatial dimension of the bounding box.
In order to retrieve a bounded property value range $[0,1]$ for the property \emph{size} ($si$), each spatial dimension of $\text{size}$~$[\text{length},\text{width},\text{height}]$  is normalized according to Eq.~\ref{eq:property_size}.
As a result, $si$ is defined as a three dimensional property.
\begin{equation}
\label{eq:property_size}
si = \left[l=\frac{\text{length}}{\max(\text{size})},w=\frac{\text{width}}{\max(\text{size})},h=\frac{\text{height}}{\max(\text{size})}\right]
\end{equation}

\subsubsection{Flatness Property}
\label{sec:property_flatness}

\noindent\textbf{Definition:}
As \emph{flatness} describes a particular aspect of an object's shape, we define it as the ratio between the area of an object's greatest plane and its overall surface area. For instance, a sheet of paper features an upper bound of \emph{flatness} while a ball an lower bound of \emph{flatness}. 

\noindent\textbf{Extraction:}
The \emph{flatness} value of an object is extracted similarly to its \emph{size}: We firstly observe the object from above (Fig.~\ref{fig:setup:physical_property}) and extract its top-level plane using RANSAC (RAndom SAmple Consensus). 
In order to increase the confidence, a candidate plane is only selected if at least 95\% of the surface normal vectors of the plane points are directed in the same  direction.
In this manner, round surfaces (as they may be observed in \emph{balls}) are rejected and subsequently a \emph{flatness} value of zero is assigned to the considered object. 
Furthermore, if the candidate plane $p$ is accepted, the plane size $|p|$, i.e the number of object points corresponding to $p$, is divided by the total number of points representing the observed object $o$ in order to obtain a bounded numeric measure of its \emph{flatness} $fl$ (Eq.~\ref{eq:property_flatness}).
Consequently, the retrieved \emph{flatness} property is bounded within a value range of $[0,1]$.
\begin{equation}
\label{eq:property_flatness}
fl = \frac{|p|}{|o|}
\end{equation}

\subsubsection{Hollowness Property}

\noindent \textbf{Definition:}
\emph{Hollowness} is the amount of visible cavity or empty space within an object's enclosed volume. 
It contrasts \emph{flatness} as it focuses on a further particular aspect of an object's shape.

\noindent \textbf{Extraction:}
\label{sec:method:hollowness}
\emph{Hollowness} contributes to the characterization of object shape. 
According to its definition, an object may enclose a volume which is not filled. 
For the sake of simplicity, we measure the internal depth $d$, which resembles the enclosed volume, and height $h$ of an object $o$: the ratio defines the \emph{hollowness} value.
In order to retrieve a reasonable measure of object's depth and height, a two camera and fiducial marker~\citep{GarridoJurado2014} setup is introduced as illustrated in Fig.~\ref{fig:setup:physical_property}.
Given the side camera view, the height $h$ of an object can be obtained by estimating the respective bounding box (see Section~\ref{sec:property_size}).
In order to retrieve depth, two fiducial markers $\{m_r, m_h\}$ are introduced (see samples in Fig.~\ref{fig:setup:physical_property}): $m_r$ serves as global reference and is placed next to the object; $m_h$ is placed inside the hollow volume of the object.
Exploiting the top camera $c_t$ perpendicularly pointed to the object, the distances $d_r=\|m_r-c_t\|$ and $d_h=\|m_h-c_t\|$ can be obtained.
Given object height $h$ and the distances $d_r$ and $d_h$, hollowness $ho$ can be approximated as shown in Eq.~\ref{eq:property_hollowness}, where $b$ (Eq.~\ref{eq:property_hollowness_object_thickness}) is introduced to consider the base height of the object, i.e. distance between the table (global reference plane) and the bottom inside the object's hollow volume.
\begin{subequations}
\begin{align}
b &= d_r - d_h \label{eq:property_hollowness_object_thickness}\\
ho &= \frac{h - b}{h} \label{eq:property_hollowness} 
\end{align}
\end{subequations}
Note that, $ho$ is inherently bounded within the interval $[0,1]$.
Furthermore the proposed method may be susceptible to noise originated in the point clouds from which the bounding box was approximated to infer the object's height $h$. 
Hence, if the difference between an object's height $h$ and distance $d_h$ (fiducial marker inside the object) is smaller than 1\si{\cm} it is cumbersome to differentiate between sensor noise and the actual \emph{hollowness} due to the low signal-to-noise ratio. To sanitize the property in such situations (particularly in case of flat objects), default value of zero is assigned.

\subsubsection{Heaviness Property}

\noindent \textbf{Definition:}
Following our basic premise of using straight forward definitions, we borrow the definition of \emph{heaviness} from physics: the object's \emph{heaviness} is the force acting on its mass within a gravitational field. 

\noindent\textbf{Extraction:}
\emph{Heaviness} $he$ of an object $o$ can be directly derived by weighing an object with a \emph{scale} (Eq.~\ref{eq:property_heaviness}); a scale with a resolution of 1\si{\g} provides an adequate precision for our scenario.
Note that, $he$ is not bounded as it is so to say bounded by physics. 
\begin{equation}
he = scale(o) \label{eq:property_heaviness}
\end{equation}
While it may require additional hardware, a robot may lift an object and calculate the \emph{heaviness} by converting the efforts observed during the process in each of its joints.

\subsubsection{Rigidity Property}
\label{sec:property_rigidity}

\noindent \textbf{Definition:}
\emph{Rigidity} of an object is defined as the degree of deformation caused by an external force vertically operating on it. 

\noindent \textbf{Extraction:}
\emph{Rigidity} of an object is extracted using an robotic arm to \emph{employ} the properties definition: equipped with a planar end-effector, we use a robotic arm to vertically exert a force onto an object until predefined efforts in the arm's joints are exceeded, see Fig.~\ref{fig:setup:physical_property_rigidity_cam_robot}. During the process we record the trajectory $tr(t)$ of the arm as well as the efforts in all of its joints. By analyzing them using an adaptive threshold-checking, we detect the first contact of the end-effector with the object $o$ at time $t_0$. Using the final position of the arm when the efforts are exceeded at $t_1$, we can calculate the deformation $def$ of an object as the vertical movement of the end-effector, that is, its movement along the $z$-axis between $t_0$ and $t_1$:
\begin{figure}[tb]
	\centering
	\def \figh {5.5cm}
	\begin{tabular}{ccc}
    \includegraphics[width=\figh]{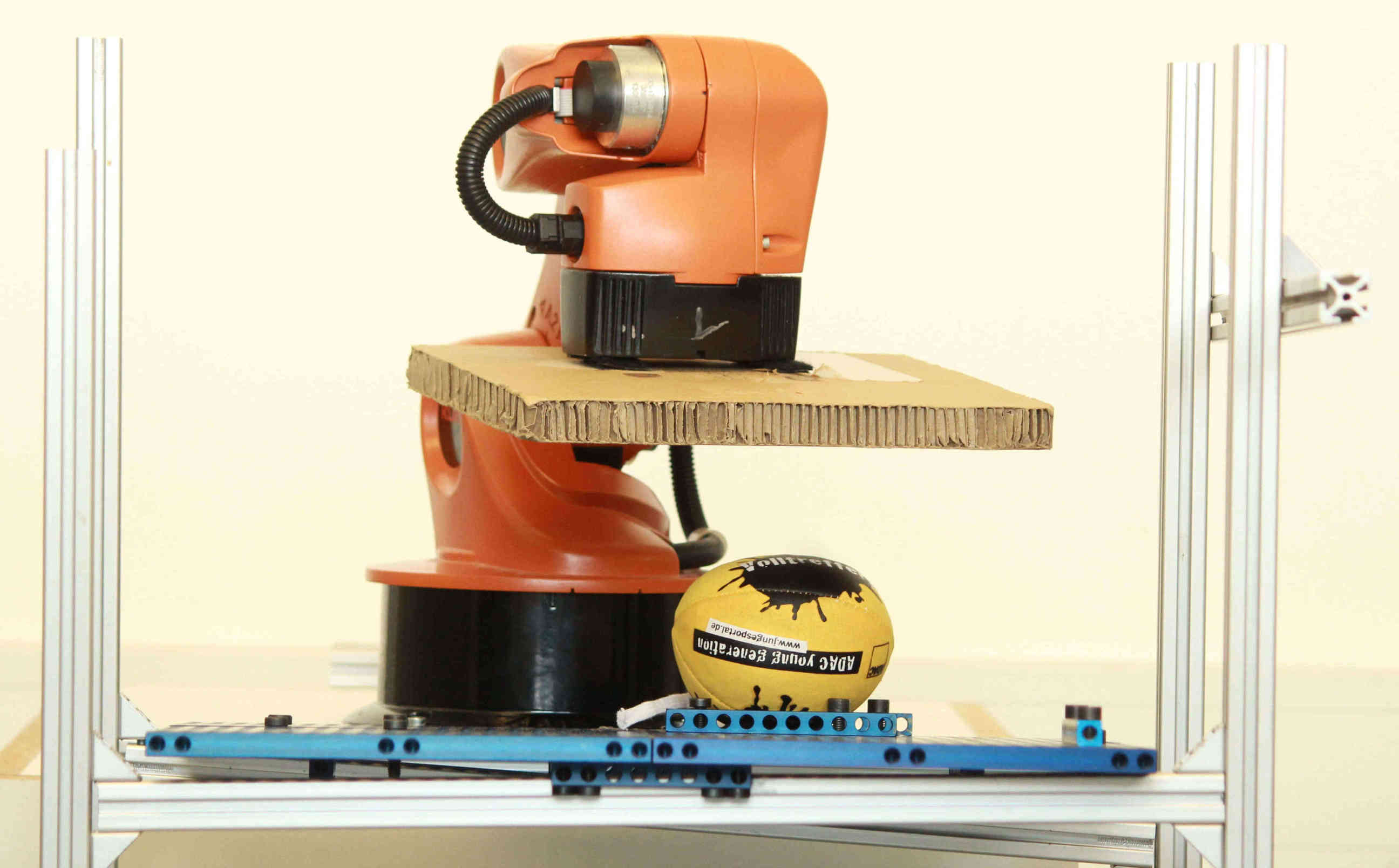} & \includegraphics[width=\figh]{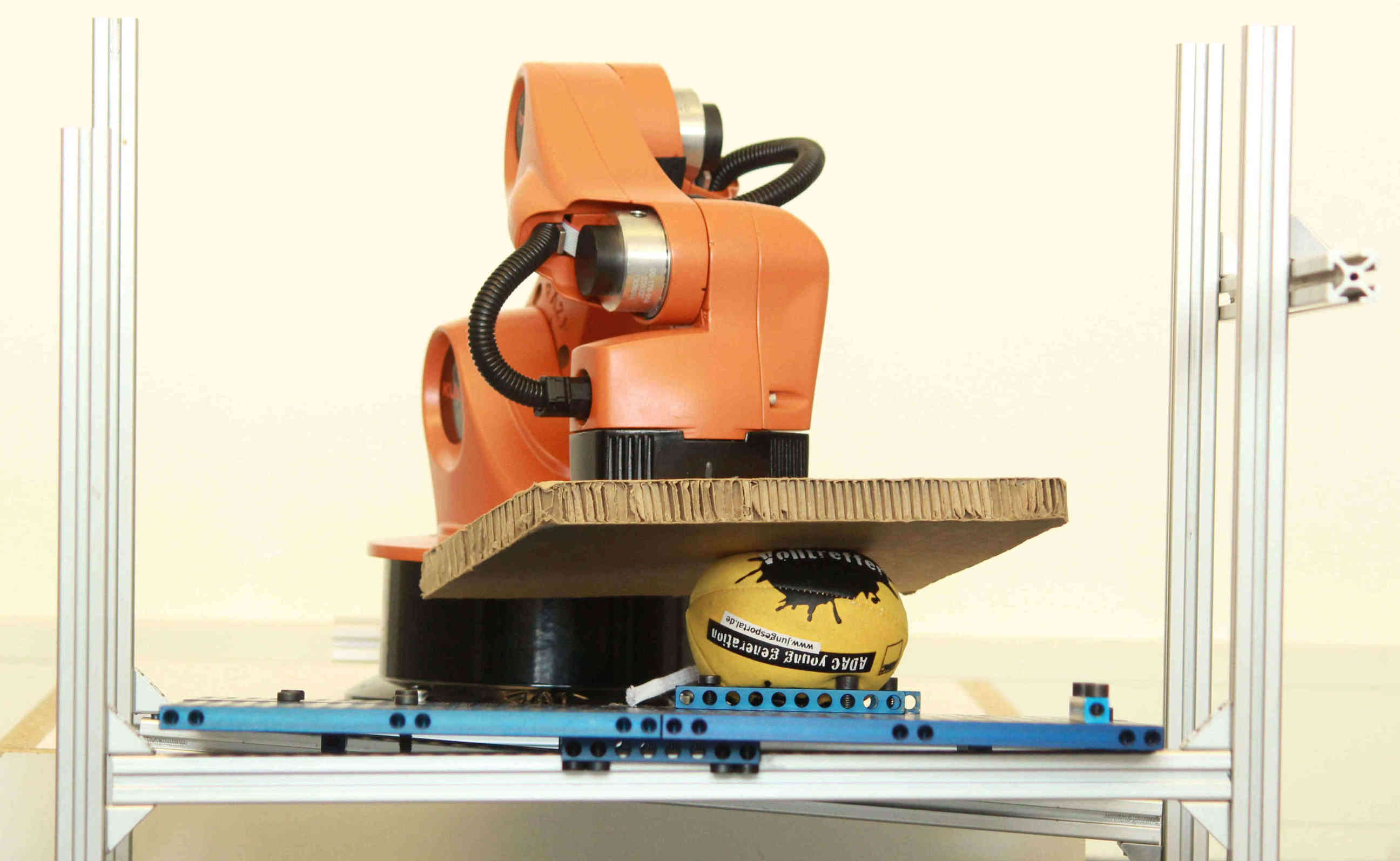} &   \includegraphics[width=\figh]{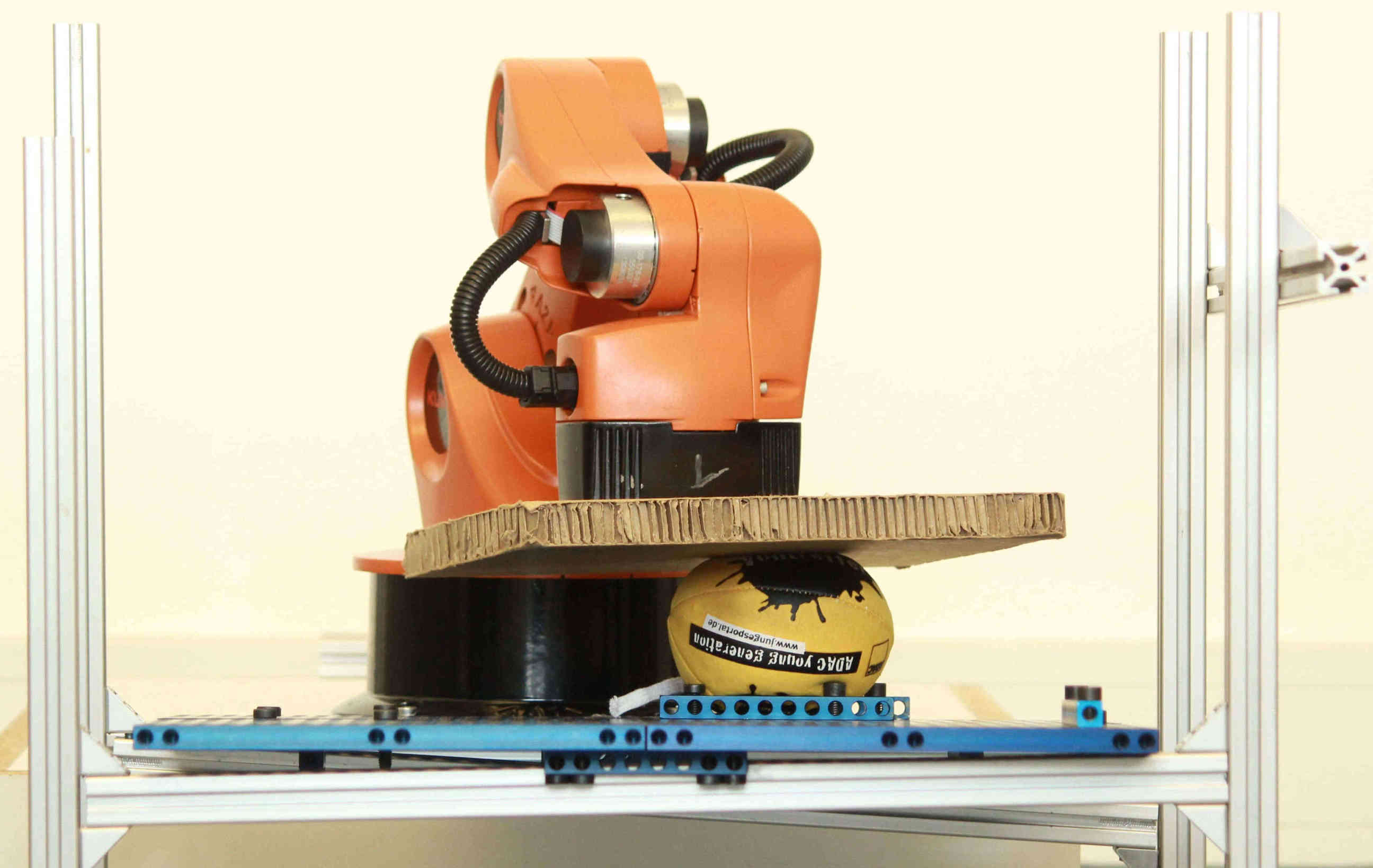} \\
\small Rigidity extraction at time $t\mathrm{=}1$ & \small Rigidity extraction at time $t\mathrm{=}2$ & \small Rigidity extraction at time $t\mathrm{=}3$ \\
 \includegraphics[width=\figh]{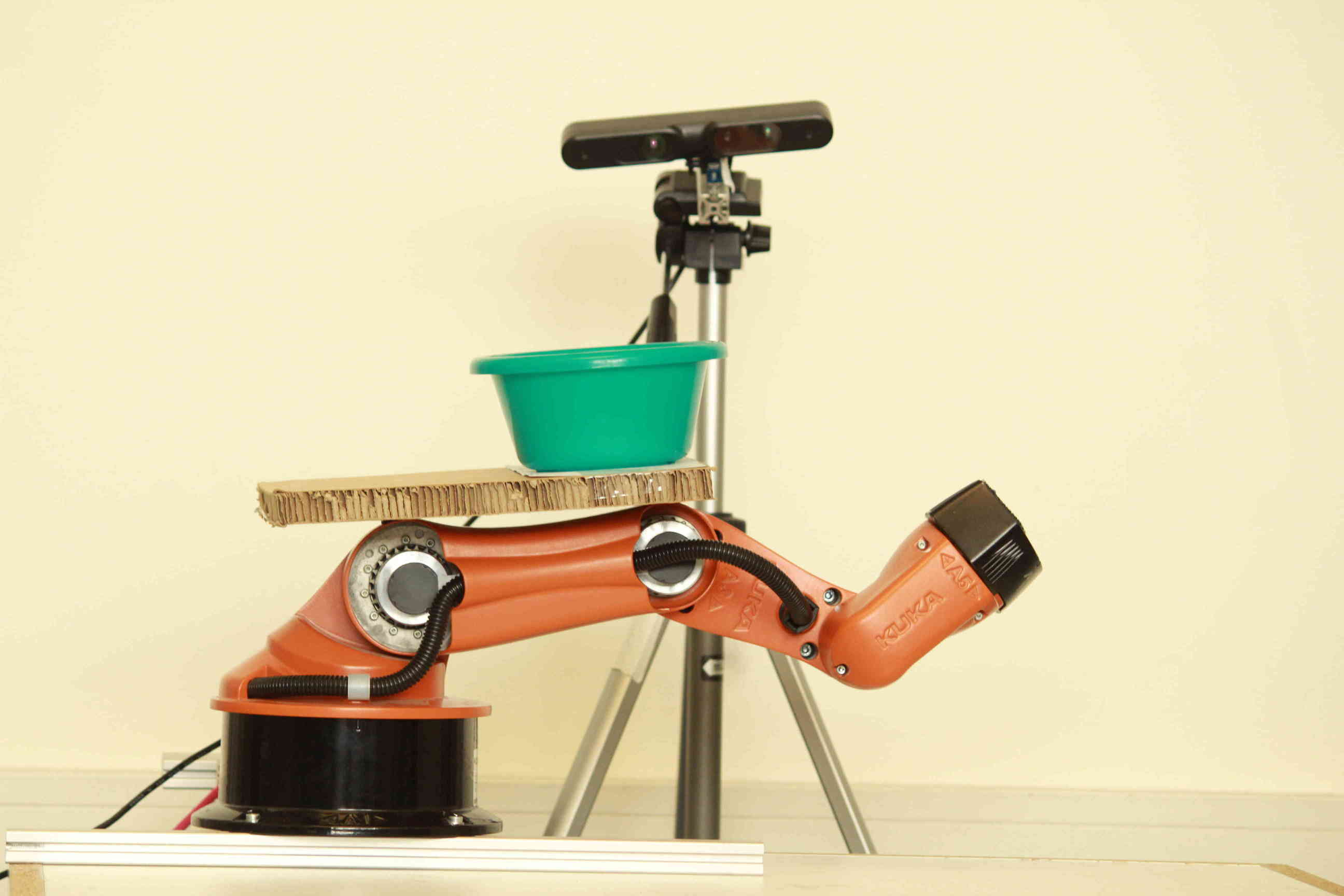} & \includegraphics[width=\figh]{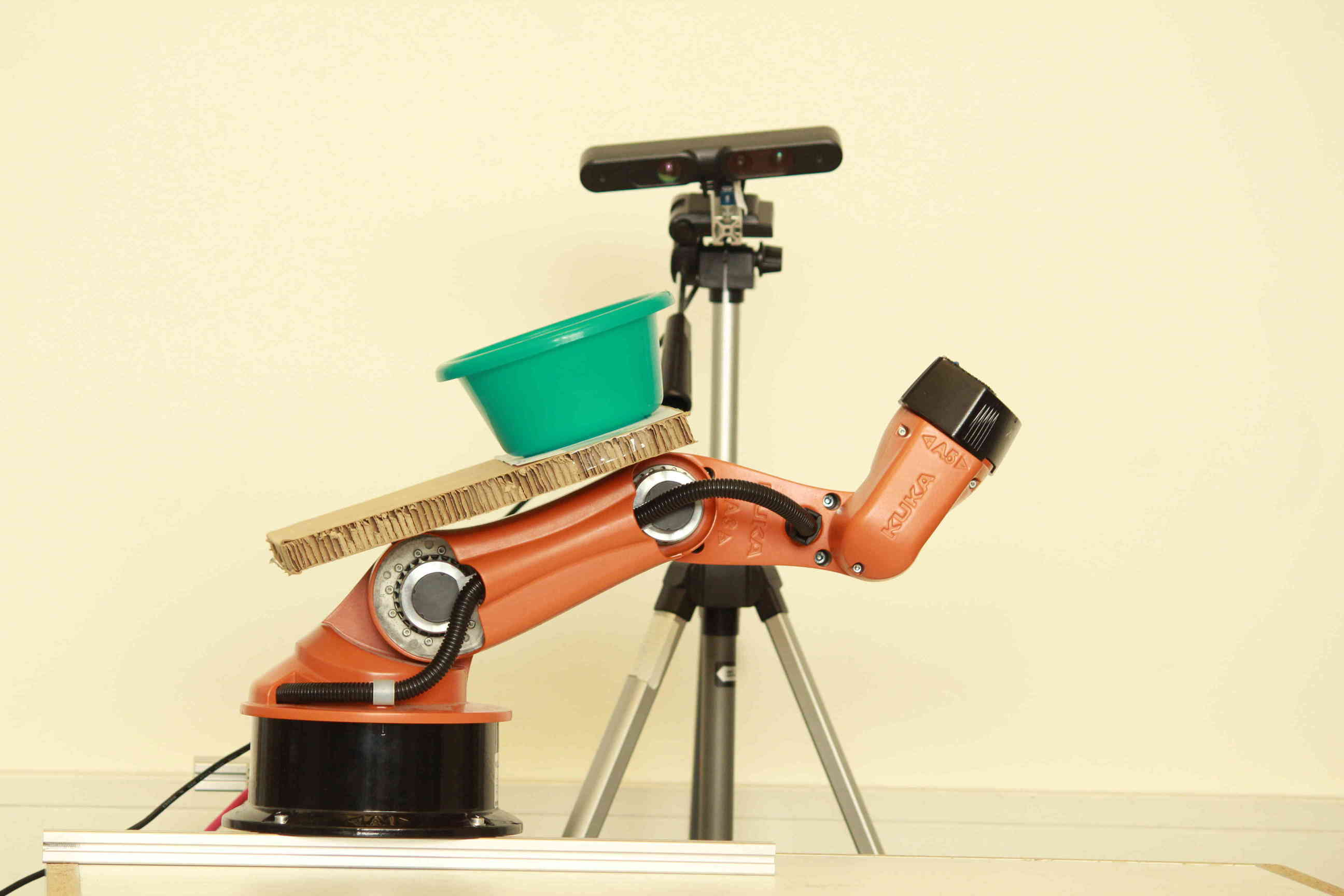} &   \includegraphics[width=\figh]{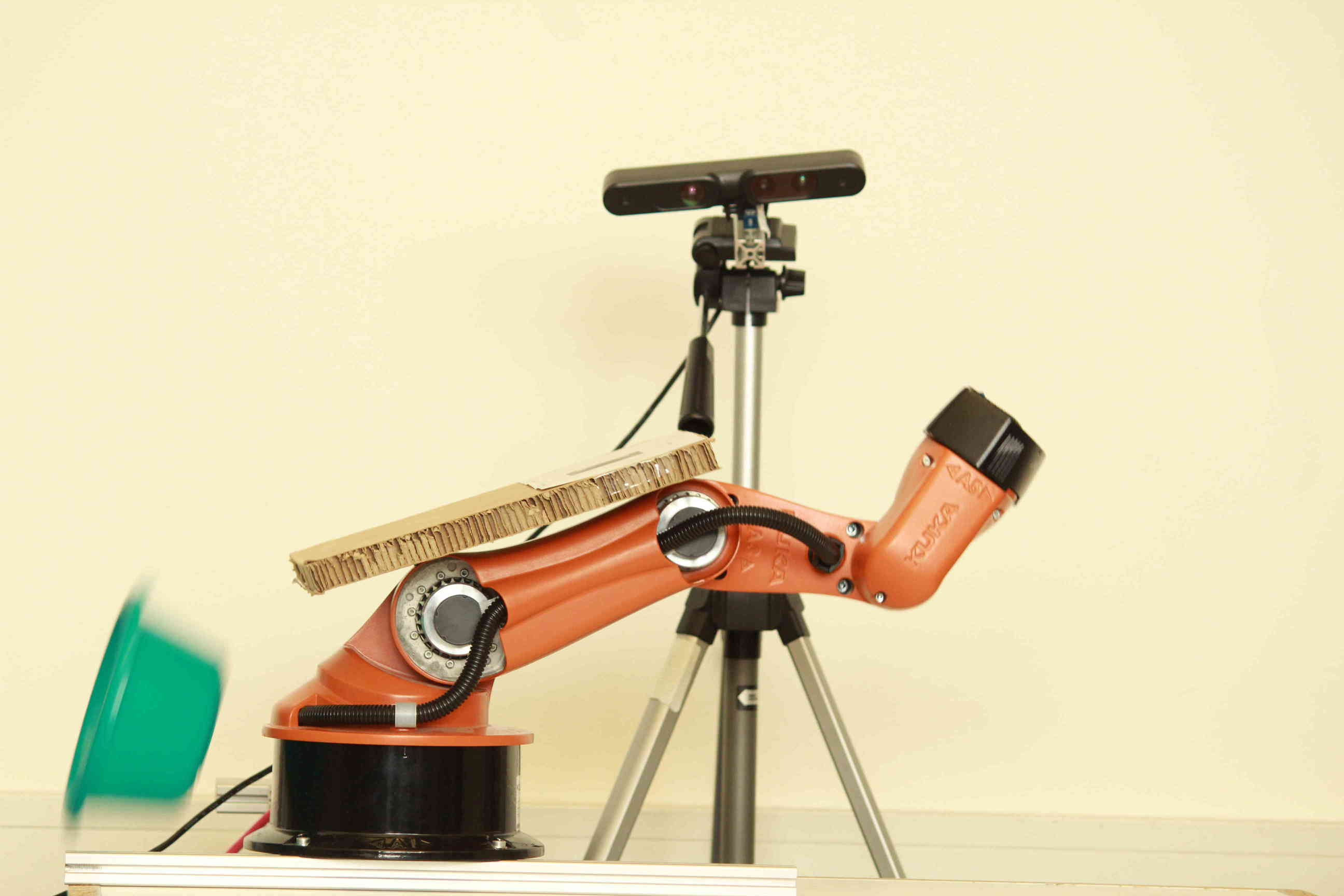} \\
\small Roughness extraction at time $t\mathrm{=}1$& \small Roughness extraction at time $t\mathrm{=}2$& \small Roughness extraction at time $t\mathrm{=}3$\\
\end{tabular}
    \caption{Light-weight experimental setup consisting of a camera-manipulator combination, for acquiring physical property \emph{rigidity} (top row) and \emph{roughness} (bottom row).}
	\label{fig:setup:physical_property_rigidity_cam_robot}
\end{figure}

\begin{subequations}
\begin{align}
	def(o) &= tr_z(t_0) - tr_z(t_1) \\
	ri &= \frac{def(o)}{h}
\end{align}
\end{subequations}
In that way, the deformation $def(o)$ is nothing but the distance the arm pushed into the considered object. For rigid objects, this deformation is zero while it is increased continuously for non-rigid objects. Finally, we normalize the deformation by the height $h$ of the object to obtain its \emph{rigidity} value $ri$.
As we use a distance as a measure of an object's deformation, $def(o)$ will always be positive. Furthermore, as an object may not be deformed more than its own height, the value of $ri$ is naturally bound to the interval of $[0,1]$.

\subsubsection{Roughness Property}
\label{sec:property_roughness}

\noindent \textbf{Definition:}
\emph{Roughness} provides information about an object's surface. Therefore, we simplify the physical idea of friction and define \emph{roughness} as an object's resistance to sliding. 

\noindent \textbf{Extraction:}
\emph{Roughness} $ro$ requires interaction as well to measure an object's resistance to sliding. 
The robotic arm is exploited to act as a ramp on which the considered object is placed, see Fig.~\ref{fig:setup:physical_property_rigidity_cam_robot}.
Starting horizontally, with an initial angle of $a_i=\ang{0}$, the ramp's angle is increased and thereby causes an increasing gravitational force pulling the object down the ramp.
When the object begins sliding, a fiducial marker that is a priori placed underneath the object, is unveiled and subsequently detected. %
As this means that the object's sliding resistance is exceeded, the ramps' angle $a_r$ is observed and exploited as a measure of \emph{roughness} as shown in Eq.~\ref{eq:property_roughness}.
In this setup, a \ang{90} ($\frac{\pi}{2}$) ramp angle represents the upper bound that induces an object to slide. Hence, it is used to normalize \emph{roughness} value $ro$ within $[0,1]$.
\begin{equation}
\label{eq:property_roughness}
ro = \frac{|a_i-a_r|}{\frac{\pi}{2}}
\end{equation}

\subsection{Functional Properties}
\label{subsec:functional_properties}
In contrast to physical properties, functional properties describe the functional capabilities or affordances \citep{Gibson} of objects.
It is proposed that functional properties do not exist in isolation, rather certain physical properties are required to enable them \citep{Baber2003_5}.
In tool use, functional properties play an important role especially when perceiving an object as a possible tool since humans in general characterize an object in terms of its functional properties rather than its physicality \citep{Hartson2003,Gibson}.
However, the term functional property (affordance) is used generally in much more broader sense  \citep{Susi2005} which raises a question: How does a functional property or affordance emerge? 
In other terms, what are the required qualifications for an ability to be recognized as a functional property or an affordance? 
A theory proposed by \citet{Kuhn2007} suggests that \emph{image schema} (such as \emph{LINK}, \emph{CONTAINER}, \emph{SUPPORT}, and \emph{PATH}) capture the necessary abstractions to model affordances. 
\textit{Image schema} is a theory proposed in psychology and cognitive linguistics and it concerns a recurring pattern abstracted from the perceptual and motor processes.
Some of the examples of image schemas are  \textit{containment}, 
\textit{support}, 
\textit{path}, and
\textit{blockage}.
These form the basis for functional abilities to \emph{contain}, \emph{support}, \emph{move}, and \emph{block} (Fig.~\ref{fig:property_hierarchy}).

To \emph{contain} is the ability of objects to hold within themselves other objects.
It is one of the most investigated image schemas and it appears in different levels of specification \citep{BennettCialone14}. 
The system presented in this paper takes a straight forward interpretation of containment as either full or partly enclosed.
\emph{Support} is an essential object relation for many objects. 
For example, \emph{support} appears as a necessary functional property for objects such as tables and trays that has the main function of carrying/supporting other objects.
\emph{move} is one of the most fundamental \citep{MANDLER2014} functional properties of any object derived from the image schema \emph{path}.
The last functional property is \emph{block}, which captures the notion of hindered movement of one object. 
While \emph{block} is derived from the image schema \emph{blockage}, the schema itself is a type of an abstract image schema called \emph{force}.
Like \emph{path}, it is also considered as one of the most fundamental schema.
 
\citet{Kuhn2007} suggested that image schema could model functional properties and that in (some) cases they can model the essential physical properties of objects.
The proposed system follows the same suit, where each functional property is defined in terms of the associated essential physical properties. 
For example, it is essential for a `coffee cup' to have the capacity to \emph{contain}, and for a `stool' to have the \emph{support} function. 
Here the functional properties are connected to the physical properties of the involved object. 
For example, for \emph{contain} the \textit{size} is relevant as it is in concrete situations not possible for an object to contain a larger object than itself. 
Likewise, \textit{rigidity} and \textit{weight} are essential properties for the \emph{support} function, as the rigidity of a supporting object needs to (on a physical level) correspond to the weight of the object being supported. 
In the following, we provide the definitions of the functional properties and their corresponding extraction methodology.

\subsubsection{Support Property}

\noindent \textbf{Definition:}
\emph{Support} describes an object's capability to support, i.e. to carry another object. 
Therefore, an object is attributed with \emph{support}, if other objects can be stably placed on top of the supporting object.
Consequently, the physical properties \emph{size}, \emph{flatness} and \emph{rigidity} are enabler of \emph{support}.

\noindent \textbf{Extraction:}
\emph{Support} requires to consider three aspects of an object. 
Firstly, the considered object needs to be rigid. 
Secondly, for carrying another object, the sizes of both may feature similar spatial proportions. 
Thirdly, the object's shape needs to be sufficiently flat in order to enable the placing of another object on top of it. 
Consequently, \emph{size}, \emph{flatness} and \emph{rigidity} are considered as core elements of the \emph{support} property, Eq.~\ref{eq:property_support}.
\begin{equation}
\label{eq:property_support}
su=[si,fl,ri]
\end{equation}

\subsubsection{Containment Property}

\noindent \textbf{Definition:}
An object is attributed with \emph{containment} if it is capable to enclose another object to a certain degree. 
This property is enabled by \emph{size} and \emph{hollowness}.

\noindent \textbf{Extraction:}
\emph{Containment} property requires to consider two aspects. 
In order to contain something, an object needs to be hollow. On the other hand, it's \emph{size} itself needs to be respected when considering whether it can contain another object. 
Thus, the value of the object's \emph{containment} $co$ property is formed by combining its \emph{size} and \emph{hollowness} property values, Eq.~\ref{eq:property_containment}.

\begin{equation}
\label{eq:property_containment}
co=[si,ho]
\end{equation}

\subsubsection{Movability Property}	

\noindent \textbf{Definition:} \emph{Movability} describes the required effort to move an object. 
The physical properties \emph{roughness} and \emph{heaviness} affect the \emph{movability} of an object.
As a result, we may interpret that \emph{movability} is enabled by these physical properties.

\noindent \textbf{Extraction:}
\emph{Movability} is based on a robot's primary ways of moving objects: either by lifting or pushing.
In both cases, \emph{heaviness} of an object is affects the \emph{movability} of an object. 
Additionally, when pushing an object, its sliding resistance expressed in form of \emph{roughness} (see Fig.~\ref{fig:setup:physical_property_rigidity_cam_robot}), needs to be considered as well.
Therefore \emph{movability} property $mo$ constitutes of \emph{heaviness} and \emph{roughness}, Eq.~\ref{eq:property_movability}.
\begin{equation}
\label{eq:property_movability}
mo=[he,ro]
\end{equation}

\subsubsection{Blockage Property}

\noindent \textbf{Definition:} \emph{Blockage} describes the capability of an object of being impenetrable, i.e. the object cannot be moved by other objects, therefore it stops the movement of other encountered objects.
Note that, given the set of physical properties, we can interpret that the \emph{blockage} property is related to \emph{roughness} and \emph{heaviness} of an object as these properties affect the intensity of being capable to block another object.
According to the property hierarchy (Fig.~\ref{fig:property_hierarchy}), \emph{blockage} is directly related to its counterpart, i.e. the \emph{movability} property.

\noindent \textbf{Extraction:}
\emph{Blockage} of an object can be derived from its \emph{movability}.
According to its definition, \emph{blockage} property $bl$ states to which degree an object is able to stop another object's movement. 
Thus, the object itself needs to be not movable by the other object, which is the inverse of its \emph{movability}, Eq.~\ref{eq:property_blockage}.
\begin{equation}
\label{eq:property_blockage}
bl= -mo=[-he, -ro]
\end{equation}

\section{Conceptual Knowledge Generation}
\label{sec:knowledge_generation}

Using the framework described in the previous section, we can employ our robotic platform to gather numerical data about an object's properties. 
However, this data can not be used for symbolic reasoning yet. 
Therefore, to facilitate this application, symbolic knowledge needs to be generated from the extracted numerical data.
In this section, we propose such bottom-up knowledge generation process to obtain knowledge about object instances and object classes (see. Fig. 2, layer 4 and 5).

For generating the knowledge, the data about the objects' physical and functional properties is processed in two stages: \textit{sub-categorization} and \textit{conceptualization}.
In the sub-categorization process, the non-symbolic continuous data of each property is transformed into symbolic data using a clustering algorithm such as K-means.
The cluster representation of the numerical values of the property data can also be seen as a symbolic qualitative measure representing each cluster.
Consequently, the number of clusters describes the granularity with which each property can qualitatively be represented. 
In case of a high number of clusters, an object is described in finer detail. Complementary, a lower number of clusters suggest a general description of an object. 
For instance, the numerical data about the \emph{rigidity} of the object instances of \emph{ceramic cup}, when clustered into three clusters, can be represented as \emph{rigidity}$\mathrm{=}\{$\emph{soft}, \emph{medium}, \emph{rigid}$\}$.
At the end of the \textit{sub-categorization} process, each object is represented in terms of the qualitative measures for each property.
The conceptualization process gathers the knowledge about all the instances of an object class and represents the knowledge about an object class.
Initially, the knowledge about objects is represented using \textit{bivariate joint frequency distribution} of the qualitative measures of the properties in the object instances.
Next, conceptual knowledge about objects is calculated as a sample proportion of the frequency of the properties across the instances of a class. 
In the following, we have provided the formal description of the knowledge generation process described above.

Consider \textbf{O} as a given set of object classes where (by abuse of notation) each object class is identified with its label.
Let each object class $O \in \textbf{O}$ be a given set of its instances.  
Let $\bigcup \textbf{O}$ be a union of all object classes such that
$\vert \bigcup \textbf{O} \vert  = n$.
Let $\textbf{P}$ and $\textbf{F}$ be the given sets of physical properties' labels and a set of functional properties' labels respectively.
By abuse of notation, each physical and functional property is identified with its label.
For each physical property $P \in \textbf{P}$ as well as for a functional property $F \in \textbf{F}$, sensory data is acquired from each object instance $o \in \bigcup \textbf{O}$.
Let $\Upsilon_P$ and $\Upsilon_F$ represent functions which maps each object instance to its measured sensory value of a physical property $P$ and a functional property $F$ respectively.
Let $P_n$ and $F_n$ represent sets such that $P_n$ and $F_n$ are the images of $\Upsilon_P$ and $\Upsilon_F$ respectively. 

\subsection{Sub-categorization -- From Continuous to Discrete}
\label{subsec:subcategory}
The sub-categorization process is performed to form (more intuitive) qualitative measures to represent the degree with which a property is reflected by an object instance.
It is the first step in creating symbolic knowledge about object classes where the symbols representing the qualitative measures of a physical or a functional property reflected in an object instance are generated unsupervisedly by a clustering mechanism.
A qualitative measure of a physical property is referred to as a physical quality and that of a functional property as a functional quality.

In this process, $P_n$ and $F_n$ representing measurements of a physical property $P \in \mathbf{P}$ and a functional property $F \in \mathbf{F}$ respectively extracted from $n$ number of object instances is categorized into a given number of discrete clusters $\eta$ using a clustering algorithm. 
Let $\nabla_P$ and $\nabla_F$ be partitions of the sets $P_n$ and $F_n$ after performing clustering on them.
Let $P_\eta$ and $F_\eta$ be the sets of labels, expressing physical qualities and functional qualities, generated for a physical property $P \in \textbf{P}$ and a functional property $F \in \textbf{F}$ respectively.
Given the label for a property, the quality labels are generated by combining a property label $P$ and a cluster label (created by the clustering algorithm).
For example, in $ size = \{ small, \text{ }medium, \text{ }big, \text{ }bigger\} $, \textit{size} is a physical property and \textit{small, medium, big, bigger} are its physical qualities. 
The semantic terms given above are meant for the readers to understand the qualitative measures of the properties.
However, in the system, the quality labels for a property $size$ are represented as $\{ size\_1, size\_2, size\_3, size\_4 \}$.
At the end of the sub-categorization process, the clusters are mapped to the generated symbolic labels for qualitative measures.
Note that the number of clusters essentially describes the granularity with which each property can qualitatively be represented.
A higher number of clusters suggest that an object is described in a finer detail, which may obstruct the selection of a substitute since it may not be possible to find a substitute which is similar to a missing tool down to the finer details.

\subsection{Attribution -- Object Instance Knowledge}
\label{subsec: attribution}

The attribution process generates knowledge about each object instance by aggregating all the physical and functional qualities assigned to the object instance by the \textit{sub-categorization} step.
In other terms, the knowledge about an instance consists of the physical as well as functional qualities reflected in the instance.
Let $\textbf{P}_\eta$ and $\textbf{F}_\eta$ be the families of sets containing the physical quality labels $P_\eta$ and the functional quality labels $F_\eta$ for each physical property $P \in \mathbf{P}$ and functional property $F \in \mathbf{F}$ respectively.
Thus, each object instance $o \in \bigcup \mathbf{O}$ is represented as a set of all the physical as well as functional qualities attributed to it which are expressed by a symbol $holds$ as:
$ holds \subset \bigcup \textbf{O}  \times  \bigcup ( \textbf{P}_\eta  \cup \textbf{F}_\eta) $.
For example, knowledge about the instance $plate_{1}$ of a \textit{plate} class can be given as,
$holds(plate_{1}, medium)$, $holds(plate_{1}, harder)$, $holds(plate_{1}, can\_support)$ where \textit{medium} is a physical quality of \textit{size} property, \textit{harder} is a physical quality of \textit{rigidity} property and \textit{can\_support} is a functional quality of \textit{support} property.

\subsection{Conceptualization -- Knowledge about Objects}
\label{subsec: concept_object}

\begin{figure}[htb!]
 	\centering
     \includegraphics[width=0.85\linewidth]{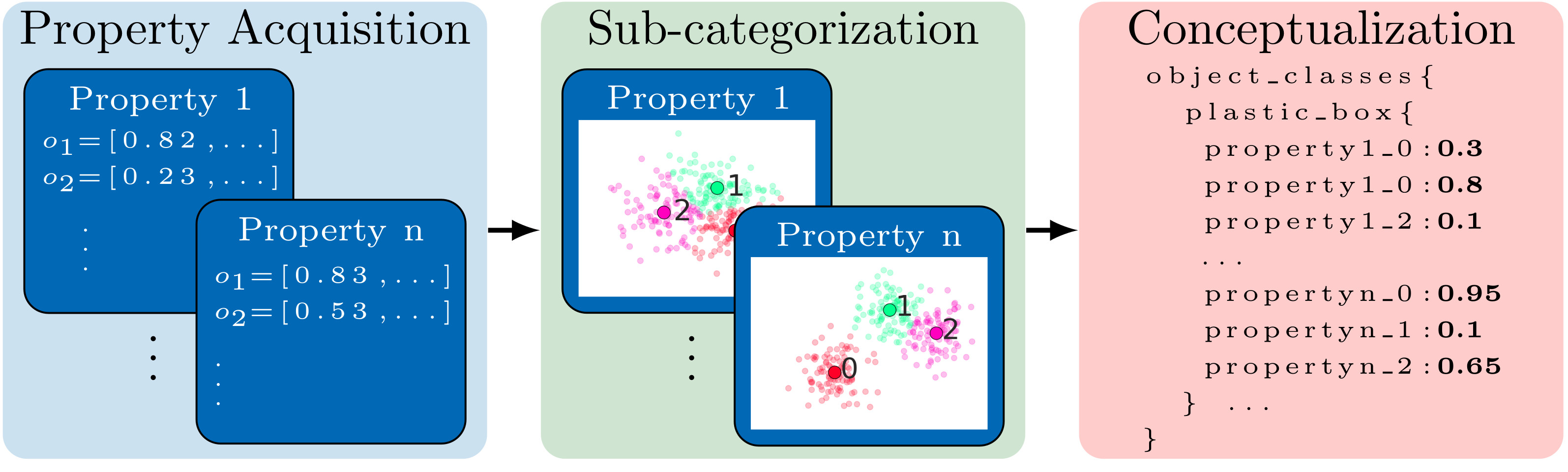}
 	\caption{The robot-centric conceptual knowledge generation process is illustrated where acquired
 	 continuous property data of objects $\{o_1,o_2...\}$ is
 	 sub-categorized into multiple clusters.
 	 Using Bi-variate joint frequency distribution and sample proportions conceptual knowledge about object classes (e.g. \emph{plasti\_box}) is generated.}
 	\label{fig:kb_generation}
 \end{figure}

The conceptualization process aggregates the knowledge about all the instances of an object class.
The aggregated knowledge is regarded as conceptual knowledge about an object class.
Let $\textbf{O}_{KB}$ be a knowledge base about object classes where each object class $O \in \textbf{O}$.
Given the knowledge about all the instances of an object class $O$, in the conceptualization process, the knowledge about the object class $O_{K} \in \textbf{O}_{KB}$ is expressed as a set of tuples consisting of a physical or a functional quality and its proportion (membership) value in the object class.
A tuple is expressed as $\langle O, t, m \rangle$ where $ t \in \bigcup \textbf{P}_\eta$  or  $ t \in \bigcup \textbf{F}_\eta $ and
a proportion value $m$ is calculated using the following membership function:
$m = P(holds(o, t) | o \in O) $. %
The proportion value allows to model the intra-class variations in the objects.
For example, knowledge about object class \textit{table} 
\{$\langle$\textit{plate, harder, 0.6}$\rangle$, $\langle$\textit{plate, light\_weight, 0.75}$\rangle$,
$\langle$\textit{plate, less\_hollow, 0.67}$\rangle$, $\langle$\textit{plate, hollow, 0.33}$\rangle$,
$\langle$\textit{plate, more\_support, 0.71}$\rangle$\},
where the numbers indicate that, for instance, physical quality \textit{harder} was observed in 60\% instances of object class \textit{plate}.
At the end of the conceptualization process, conceptual knowledge about an object class is created which is represented in a symbolic fuzzy form and grounded into the human-generated or machine-generated data about the properties of objects.
The knowledge about objects is then used to determine a substitute from the existing objects in the environment.
The Figure \ref{fig:kb_generation} illustrates graphically the main processes of \textit{Sub-categorization} and \textit{Conceptualization}.

\section{Experimental Evaluation}
\label{sec:eval}
In the following evaluation, multiple
experiments are conducted to evaluate the proposed approach on different semantic levels: from the property extraction of real world objects to an eventual application scenario in the context of tool substitution.

For this purpose, we introduce the RoCS dataset containing extracted physical and functional properties of objects in Section~\ref{sec:eval:ROCS_dataset}.
Given the dataset we conduct an evaluation on the physical object properties investigating, the stability of the extraction methods, the coverage w.r.t. range of characteristics captured by selected dataset instances and the correlation among properties in Section~\ref{sec:eval:property_extraction_methods}. 
Using k-means clustering on functional object properties in Section~\ref{sec:eval:property_semantics} and \ref{sec:knowledge_generation} we show
that the chosen properties may allow to discriminate instances of different object classes and may function as general concepts that can describe objects.
Finally, Section~\ref{sec:eval:tool_substitution} shows the applicability of the dataset by learning a model from the generated conceptual object knowledge given the extracted properties and applying it to a tool substitution scenario under real world conditions.

\subsection{RoCS Dataset}
\label{sec:eval:ROCS_dataset}
For the sake of a thorough evaluation of our conceptual framework the 
\textbf{Ro}bot-\textbf{C}entric data\textbf{S}et (RoCS) is introduced; note that, we propose a Robot Operating System (ROS)~\citep{Koubaa2017} based implementation to acquire object data used in the following evaluation.
In the following, we briefly introduce the hardware setup and procedures for acquiring raw object data, describe its parameters (e.g. thresholds) and the contents of the final dataset.

\subsubsection{Hardware Setup}

Figure~\ref{fig:property_hierarchy} lists the required hardware as data sources. For visual and non-invasive extraction methods, RGB-D sensors are required. More specifically, the \emph{size} property requires a lateral view on objects while the \emph{hollowness} property relies on a birds-eye view. Hence, we employ two Asus Xtion Pro depth sensors~\citep{Swoboda2014} (see Fig.~\ref{fig:setup:physical_property}). 
To extract the physical properties \emph{rigidity} and \emph{roughness}, a robotic arm is required to interact with objects.
Here we integrate a Kuka youBot~\citep{Bischoff2011} manipulator; nevertheless, our property extraction methods require arm joint state values which are generally provided by alternative manipulators.
Finally, a common kitchen scale with a resolution of 1\si{\gram} is used to extract the weight and \emph{heaviness} of objects.

\subsubsection{Object Property Acquisition Procedure}
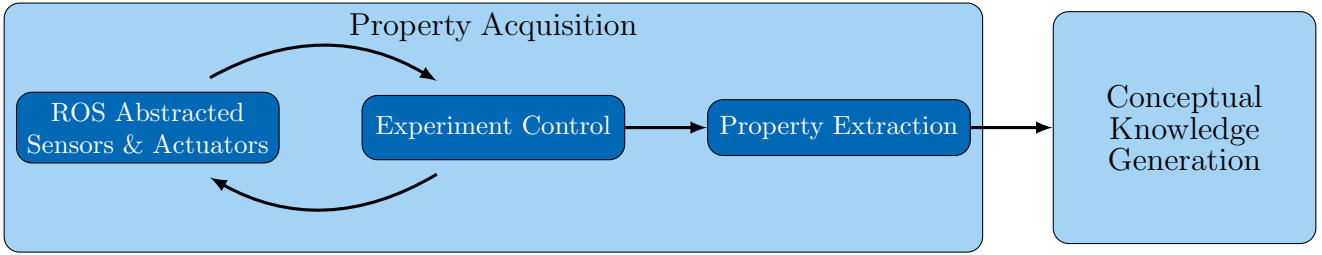
\begin{figure*}[tb]
\centering
\scalebox{1}{
\begin{tikzpicture}
\definecolor{PHY2}{HTML}{A4D3F6} 
\node[element, fill=PHY2,text=black, text depth=2.75cm, text width=2.75*\xshift, minimum height=4.15*\yshift, minimum width=17.5*\yshift] (PROPACQ) at (2.5*\xshift, 0) {\normalsize Property Acquisition};

\node[element,text width=1.75*\xshift] (ROS) at (0,0) {ROS Abstracted Sensors \& Actuators};
\node[element,text width=1.75*\xshift, minimum height=1.15*\yshift] (DATA) at (2.5*\xshift, 0) {Experiment Control};
\node[element,text width=1.75*\xshift] (EXTRACT) at (5*\xshift,0) {Property Extraction};
\node[element,fill=PHY2, text=black, text width=1.75*\xshift,minimum height=4.15*\yshift] (KNOWLEDGE) at (7.5*\xshift,0) {\normalsize Conceptual Knowledge Generation};

\path[edge, bend left=30] (ROS) edge[transform canvas={yshift=0.25*\yshift}]  (DATA);
\path[edge, bend left=30] (DATA) edge[transform canvas={yshift=-0.25*\yshift}]  (ROS);
\path[edge] (DATA) edge (EXTRACT);
\path[edge] (EXTRACT) edge (KNOWLEDGE);

\end{tikzpicture}
}
\caption{Data flow within the dataset creation framework.}
\label{fig:framework}
\end{figure*}

Using the described hardware, we implemented a preliminary ROS-based framework to extract the physical and functional properties of objects. 
A schematic overview on the framework is given by Fig.~\ref{fig:framework}. 
The interface for operating sensors and actuators is provided to our framework by ROS. This interface is used by different experiments for observing and interacting with objects to acquire the necessary sensory data. Together both blocks (\textit{ROS Abstracted Sensors \& Actuators} and \textit{Experiment Control}) form a control loop which generates feature data (see Fig.~\ref{fig:property_hierarchy}). 
According to the selected properties four control loops are implemented as separate experiments. 
The first experiment is non-invasive and gathers the visual feature data required for \emph{hollowness}, \emph{flatness} and \emph{size}; Fig.~\ref{fig:setup:physical_property} illustrates the camera setup. 
Initially a table-top object detection is introduced that uses a RAndom SAmple Consensus (RANSAC) based plane fitting approach in order to detect object candidates on the table.
The RANSAC algorithm is parameterized with a leaf size of $0.0025$\si{\m}, a maximum of $10^4$ iterations and a $0.02m$ distance threshold between points and the estimate plane model.
Note that, RANSAC is also used in this experiment for segmenting planes for the property \emph{flatness}.
Furthermore, fiducial markers (ArUco Library~\citep{GarridoJurado2014}) with sizes of $14~cm$ and $3~cm$ are used for the \emph{hollowness} property.
The second experiment uses the robotic arm to deform objects to facilitate the extraction of \emph{rigidity} (see Section~\ref{sec:property_rigidity}). We set the efforts to exceed in each joint to $\pm 8~\si{\N}\si{\m}$. %
Within the third experiment, the robotic arm is used as a ramp to extract an object's \emph{roughness} (see Section~\ref{sec:property_roughness}). To achieve an appropriate resolution, the angular speed of the joint lifting the ramp is set to $0.05~\si{\radian}/\si{\s}$.
Finally, the last experiment employs a kitchen scale with a resolution of $1\si{\g}$ to extract the objects' weight.
Following the \textit{Experiment Control}, the individual extraction methods process the generated feature data as described in Section~\ref{subsec:property_acquisition} to produce physical and functional property values of the considered object. Finally, this data can be accumulated for a set of objects and further processed to generate conceptual knowledge.

\subsubsection{Dataset Structure}
\begin{figure}[tb]
\centering
\begin{tikzpicture}
\node[inner sep=0pt] (ballpcd) at (0,0){\includegraphics[width=1.5cm]{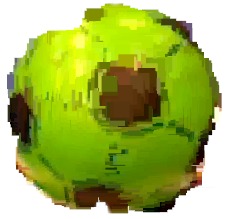}};
\node[inner sep=0pt] (ballrgb) at ([xshift=1.5cm,yshift=1.5cm]ballpcd){\includegraphics[height=1.5cm]{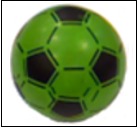}};

\node[inner sep=0pt] (bowlpcd) at (4,0){\includegraphics[width=1.5cm]{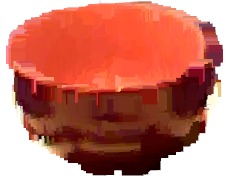}};
\node[inner sep=0pt] (bowlrgb) at ([xshift=1.5cm,yshift=1.5cm]bowlpcd){\includegraphics[height=1.5cm]{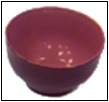}};

\node[inner sep=0pt] (boxpcd) at (8,0.5){\includegraphics[width=1.65cm]{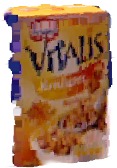}};
\node[inner sep=0pt] (boxrgb) at ([xshift=1.5cm,yshift=1.0cm]boxpcd){\includegraphics[height=1.5cm]{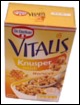}};

\node[inner sep=0pt] (cuppcd) at (11.75,0){\includegraphics[width=1.5cm]{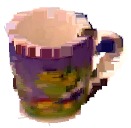}};
\node[inner sep=0pt] (cuprgb) at ([xshift=1.5cm,yshift=1.5cm]cuppcd){\includegraphics[height=1.5cm]{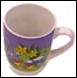}};
\end{tikzpicture}
\caption{RoCS dataset samples: Point cloud and RGB images of a \emph{ball}, \emph{bowl}, \emph{paper box}, and \emph{cup} (for visualization purposes, images are scaled and 3D points are magnified).}
\label{fig:rocs_dataset_samples}
\end{figure}

For the RoCS dataset we consider 11 different object classes (\emph{ball}, \emph{book}, \emph{bowl}, \emph{cup}, \emph{metal\_box}, \emph{paper\_box}, \emph{plastic\_box}, \emph{plate}, \emph{sponge}, \emph{to\_go\_cup} and \emph{tray}) featuring various object characteristics in appearance to functional purpose.
Each class consists of 10 unique object instances that leads to a total number of 110 object instances; Fig.~\ref{fig:rocs_dataset_samples} illustrates sample object instance of RoCS dataset.

In order to evaluate the performance of the proposed property extraction methods, such as stability, for each object instance we capture 10 repetitions without modifying the setup.
As a result we captured 1100 object observations for which physical and functional property values are generated.
The dataset is publicly available at \url{https://gitlab.com/rocs_data/rocs-dataset}.
    
\subsection{Property Extraction}
\label{sec:eval:property_extraction_methods}

The objective of the first part of the evaluation is to investigate the property extraction methods as described in Section~\ref{subsec:property_acquisition}. 
At this level, we only focus on physical properties as functional properties are built on the basis of an object's physical properties. 
First, we analyze the stability of the extraction methods to determine how deterministic and reproducible the data acquisition is for each property and object. 
Furthermore, we explore the coverage of our data set to determine the variance and range of objects reflected in the different classes and properties. 
Lastly, we inspect the correlation among different properties in our data. 

\subsubsection{Extraction Stability}

\begin{figure}[tb]
\centering
\begin{tabular}{ccc}
 \includegraphics[width=5.5cm]{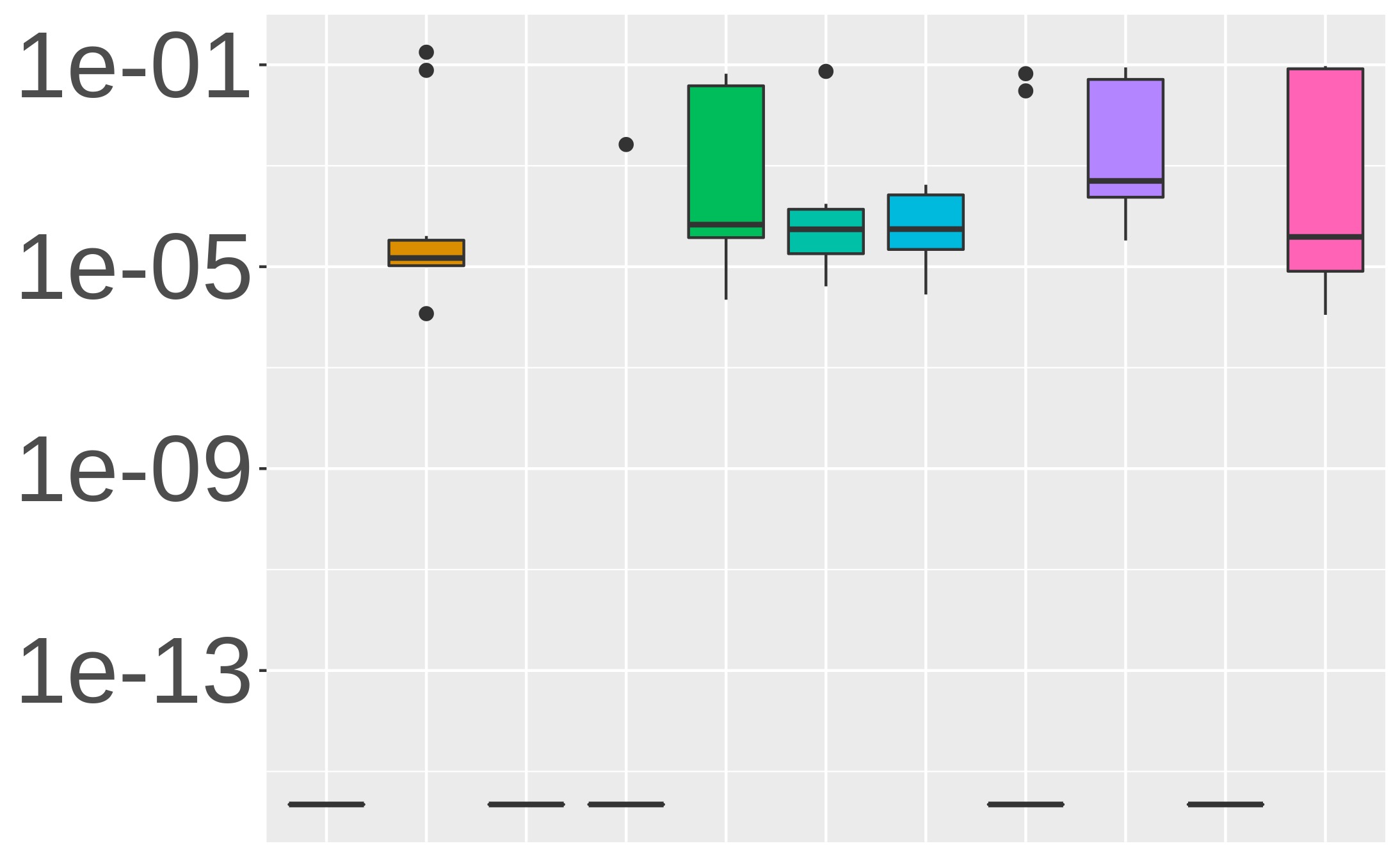} &   \includegraphics[width=5.5cm]{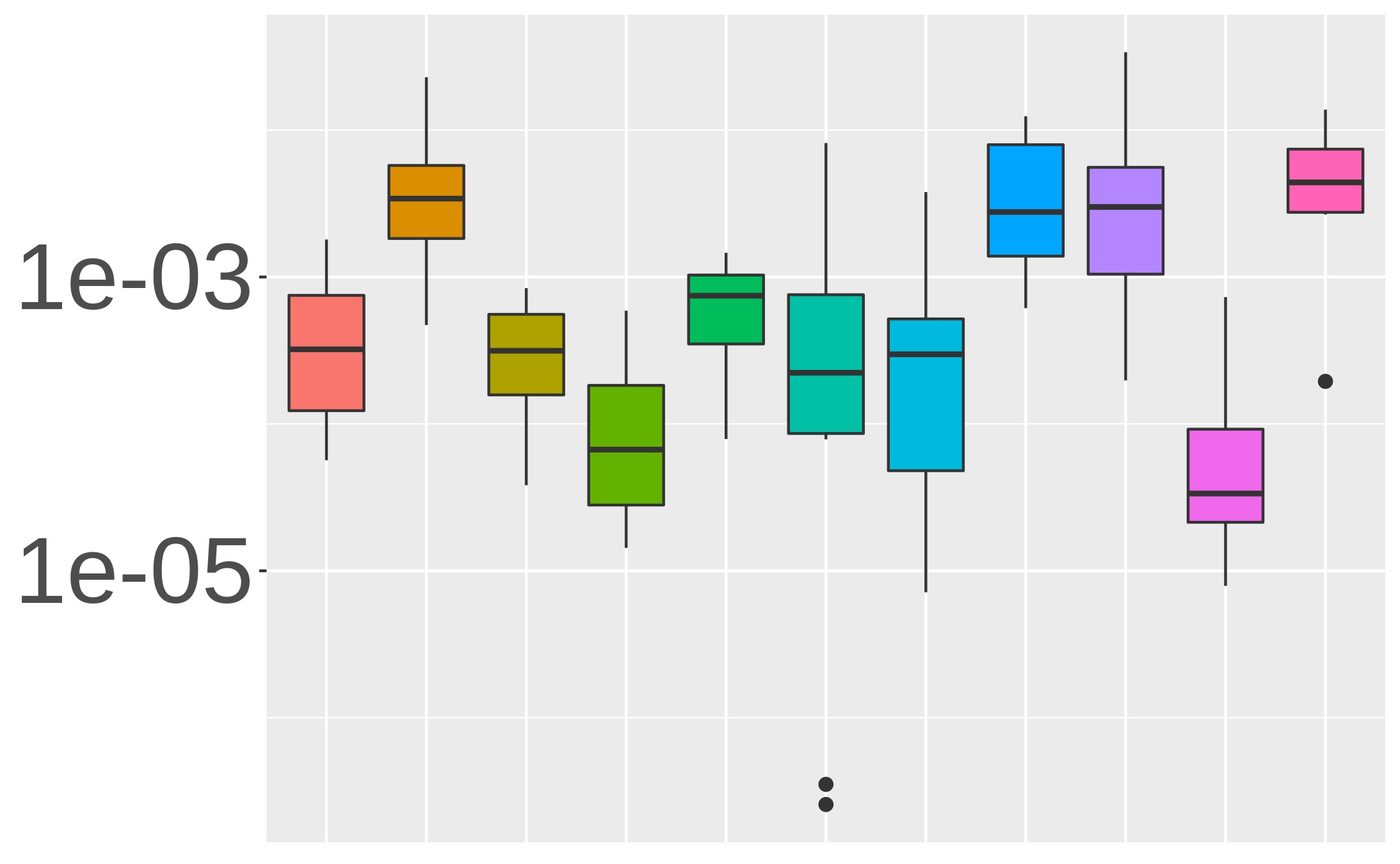} & \includegraphics[width=5.5cm]{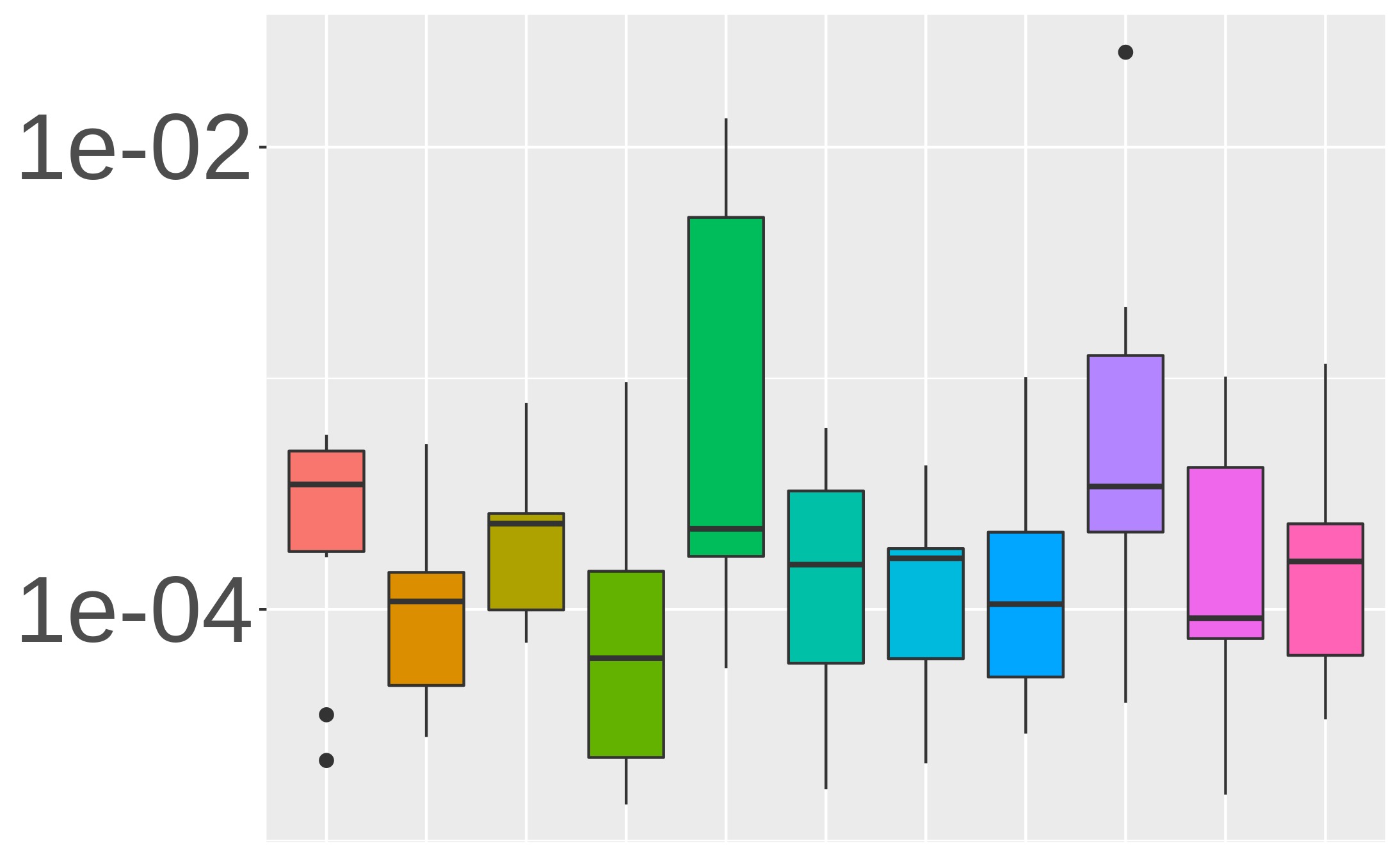} \\
(a) \small $fl$ (\emph{flatness}) & (b) \small $ri$ (\emph{rigidity}) & (c) \small $ro$ (\emph{roughness}) \\[6pt]
   \includegraphics[width=5.5cm]{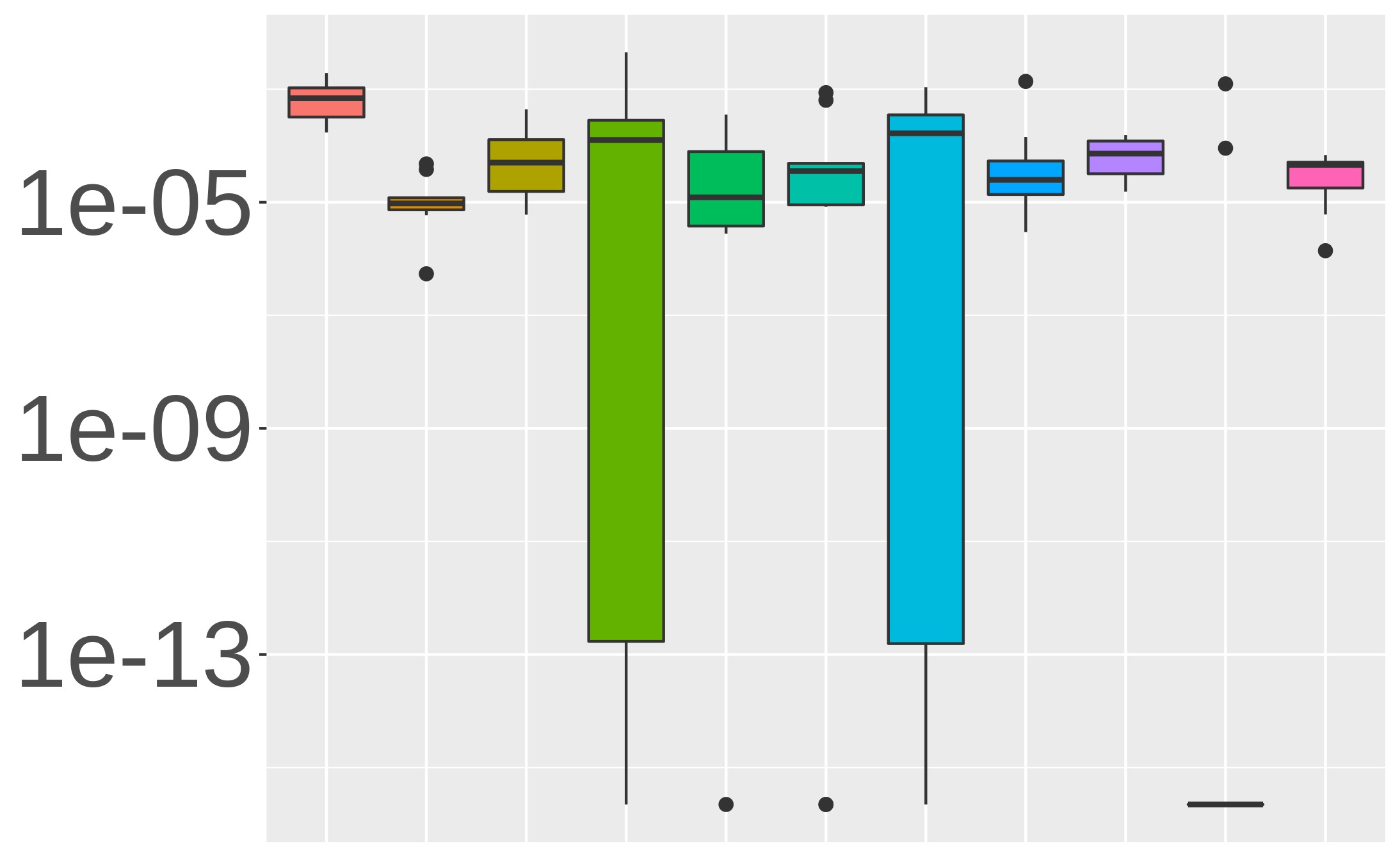} &    \includegraphics[width=5.5cm]{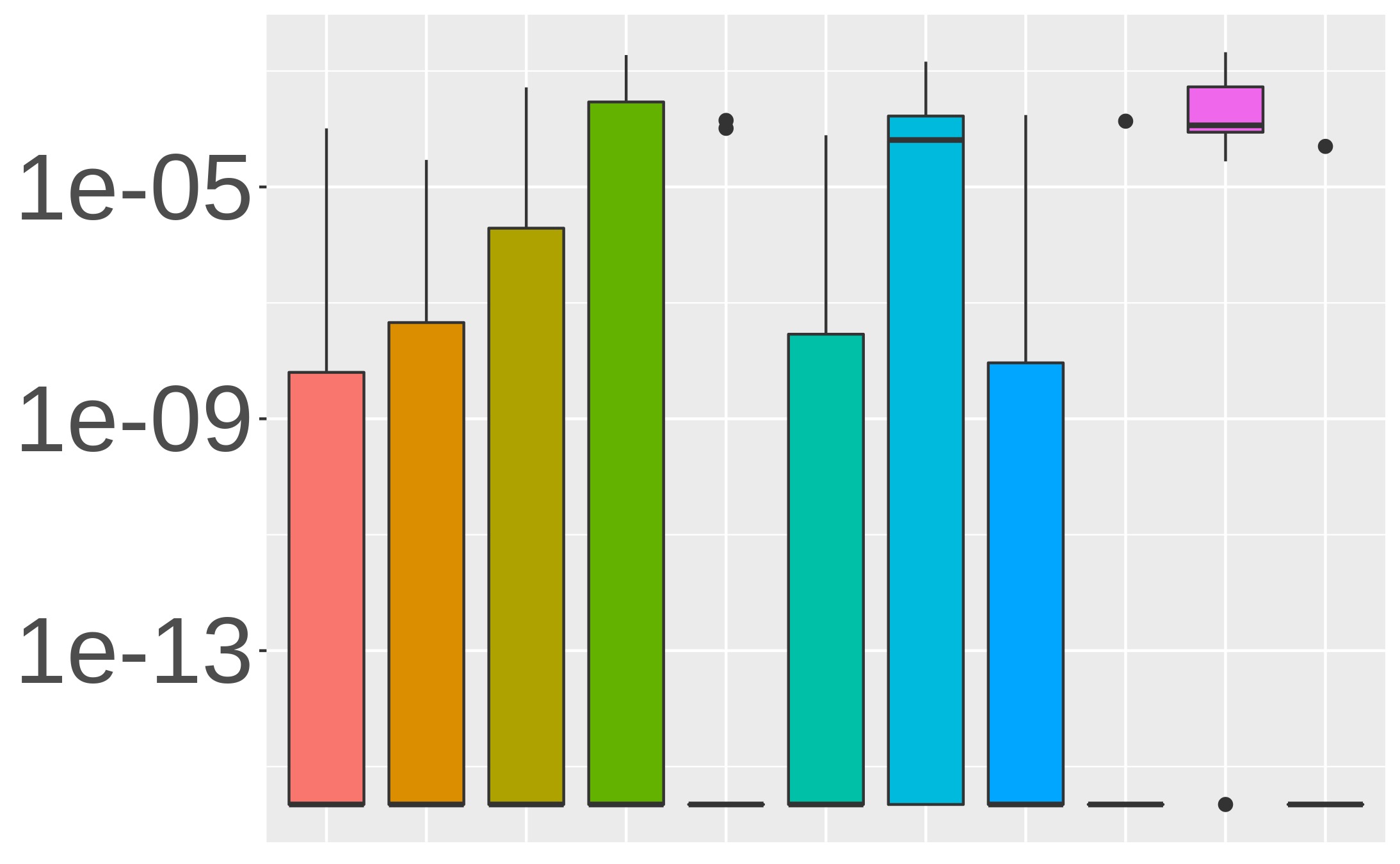} &     \includegraphics[width=5.5cm]{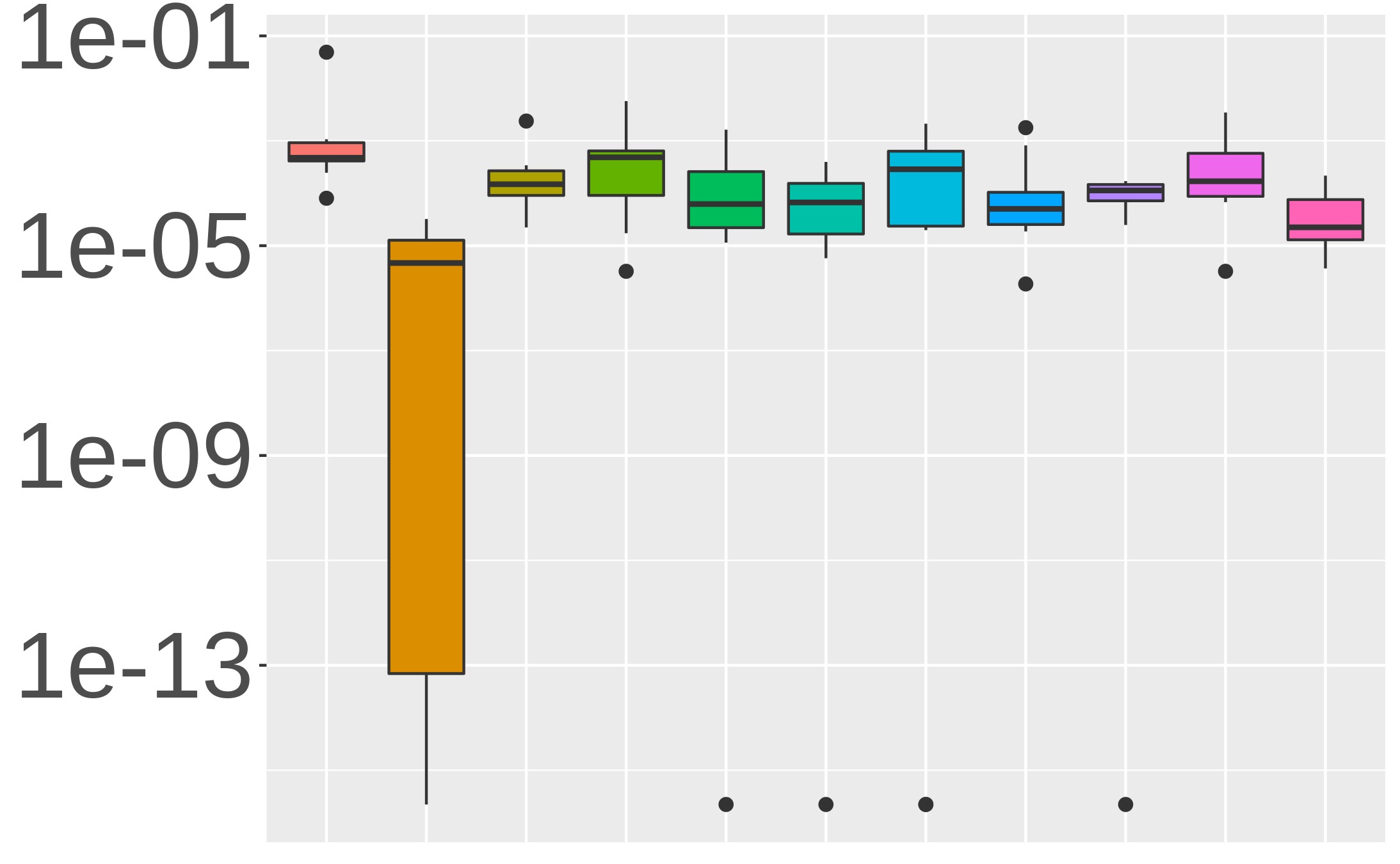} \\     
(d) \small  $h\in si$ (height of \emph{size}) & (e) \small $w\in si$ (width of \emph{size}) & (f) \small $l\in si$ (length of \emph{size}) \\[6pt] 
 &  \includegraphics[width=5.5cm]{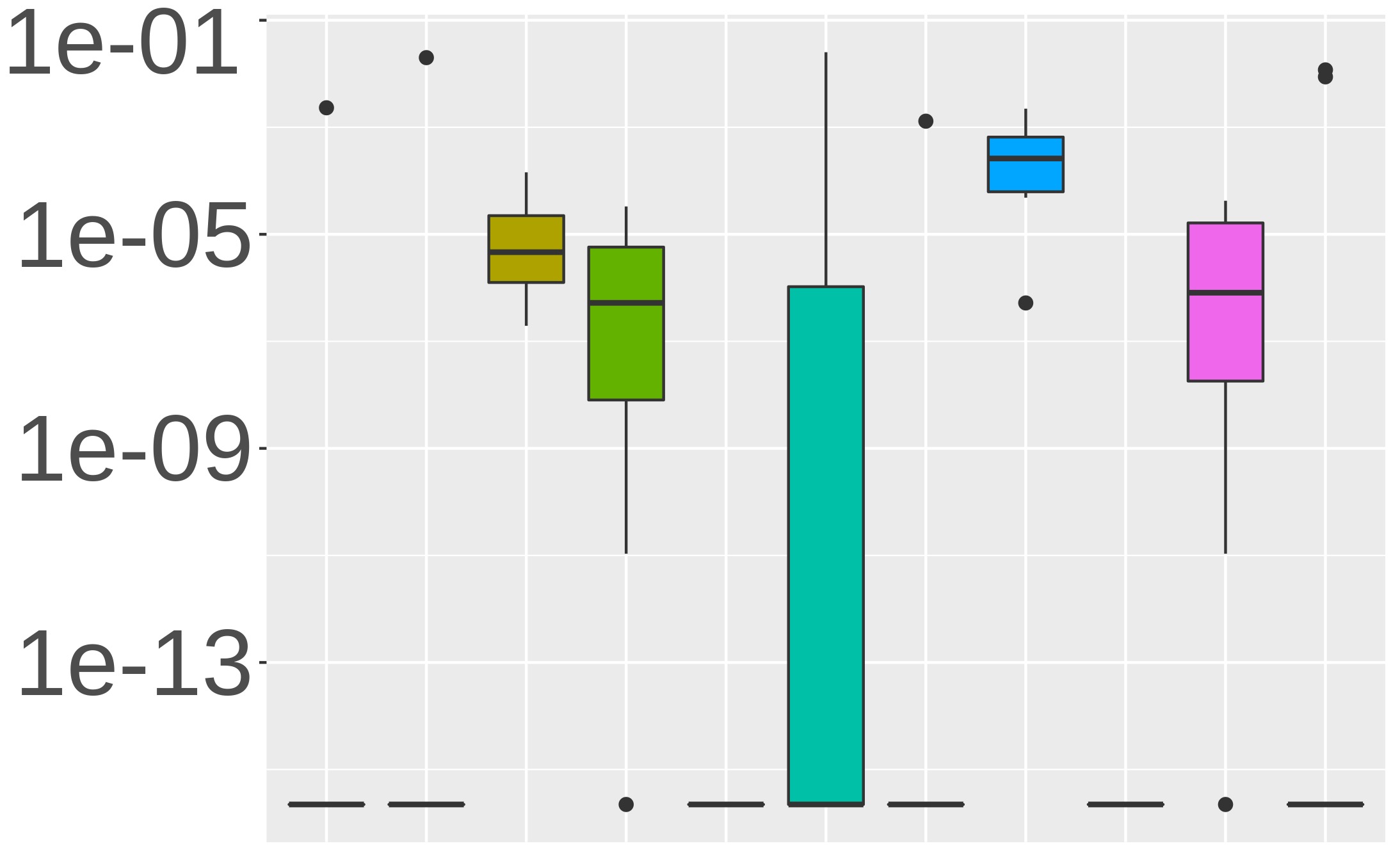} &\includegraphics[width=4.8cm]{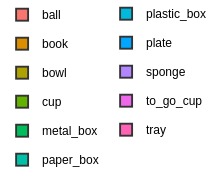} \\
& (h) \small  $ho$ (\emph{hollowness}) & (i) \small legend \\[6pt]
\end{tabular}
\caption{Mean variance for physical properties $[fl,ri,ro,si,he,ho]$ illustrated in form of a Box plot (in log-scale to provide insights of respective intra property variances compared to linear-scale shown in Table~\ref{tab:variance_instances}). Note that, in order to be able to display all variances (including zero) in log-scale, we add an epsilon on each value before computing log. Heaviness is excluded as all variance values are zero for this property due to the resolution of the scale.}
\label{fig:variance-log}
\end{figure}

The abstraction process from raw sensor data to symbolic object property knowledge requires a stable processing. %
However, noise is naturally affecting data when working with sensors and real world objects.
To compensate for the caused uncertainty, each RoCS object instance consists of 10 repetitions. %
We use these in the following to analyze the stability of the proposed property extraction methods.
For that, the variance of each physical property of each object instance is analyzed. 
More specifically, given the 10 repetitions of a particular object instance for each of its physical properties, we calculate the variance of the property values of its 10 repetitions. As we consider  6 physical properties from which property \emph{size} has 3 dimensions, we obtain 8 values per object instance and therefore 880 values in total.
We further reduce the data, by calculating the mean of the object variances for a particular object class and property as shown in Table~\ref{tab:variance_instances}, whereas Fig.~\ref{fig:variance-log} illustrates the variances of all object instances within one object class as box plots; the colored middle box represents 50\% of the data points and the median of the class is indicated by the line that divides the box. 

\begin{table}[tb]
\scriptsize
\centering
\caption{Mean variance for each physical property.}
\label{tab:variance_instances}
\setlength\tabcolsep{0.0pt}
\begin{tabular}{r*{9}{R}}
\textbf{class}        & \textbf{flatness} & \textbf{rigidity} & \textbf{roughness} & \textbf{size\_length} & \textbf{size\_width} & \textbf{size\_height} & \textbf{heaviness} & \textbf{hollowness} & \textbf{class\_mean} \\ \hline
\textbf{ball}         & 0                 & 0.00053           & 0.00032            & 0.00538               & 0.00001              & 0.00083               & 0               & 0.00023             & 0.00091                          \\
\textbf{book}         & 0.02554           & 0.00583           & 0.00015            & 0.00001               & 0.00001              & 0.00002               & 0               & 0.002               & 0.00419                          \\
\textbf{bowl}         & 0                 & 0.00037           & 0.00025            & 0.00038               & 0.00006              & 0.00012               & 0               & 0.00003             & 0.00015                          \\
\textbf{cup}          & 0.00026           & 0.00015           & 0.00017            & 0.00098               & 0.0003               & 0.00079               & 0               & 0.00001             & 0.00033                          \\
\textbf{metal\_box}   & 0.01939           & 0.00074           & 0.0039             & 0.00028               & 0.00002              & 0.00007               & 0               & 0                   & 0.00305                          \\
\textbf{paper\_box}   & 0.00747           & 0.00115           & 0.00021            & 0.00011               & 0.00002              & 0.00017               & 0               & 0.0035              & 0.00158                          \\
\textbf{plastic\_box} & 0.00015           & 0.00071           & 0.00016            & 0.00056               & 0.00021              & 0.0003                & 0               & 0.00013             & 0.00028                          \\
\textbf{plate}        & 0.00971           & 0.00481           & 0.00022            & 0.0003                & 0.00003              & 0.00017               & 0               & 0.0005              & 0.00197                          \\
\textbf{sponge}       & 0.02503           & 0.00705           & 0.00313            & 0.0001                & 0.00001              & 0.00008               & 0               & 0                   & 0.00443                          \\
\textbf{to\_go\_cup}       & 0                 & 0.00016           & 0.00031            & 0.00061               & 0.00044              & 0.00013               & 0               & 0.00001             & 0.00021                          \\
\textbf{tray}         & 0.03486           & 0.00569           & 0.00024            & 0.00005               & 0.00001              & 0.00004               & 0               & 0.00206             & 0.00537                          \\
\textbf{prop\_mean}         & 0.01113           & 0.00247           & 0.00082            & 0.0008                & 0.0001               & 0.00025               & 0               & 0.00077             & 0.00204                           
\end{tabular}

\begin{flushleft}
\small
Each value represents the mean variance of extracted property values of an particular object class consisting of 10 instances and their respective repetitions. Variances are scaled by color in ascending order from transparent (0) to red (highest variance).
\end{flushleft}
\end{table}

The results of the Table~\ref{tab:variance_instances} (Fig.~\ref{fig:variance-log}) reveal that the class variances are overall low, which implies stable property extraction methods in general.
The highest variances can be found for the \emph{flatness} property.
The extraction of the \emph{flatness} property for small and flat object instances is particularly affected by noise due to the low signal-to-noise ratio. %
Furthermore, it can be observed that for \emph{ball}, \emph{bowl} and  \emph{to\_go\_cup} the variance of the \emph{flatness} property is zero due to the fact that no top-level plane can be extracted for instances of these classes as they feature either round or negligible small top-level surfaces (see  Section~\ref{sec:property_flatness}).
Similarly, a higher variance can be observed for the \emph{rigidity} property which is caused by smaller object instances, such as \emph{book}, \emph{plate}, \emph{sponge} and \emph{tray}. Here the detection of the first contact with the object causes false positives and therefore introduces varying deformation values.

In contrast, for the \emph{hollowness} property the variance for \emph{metal\_box} and \emph{sponge} are zero. Such object instances predominantly feature flat surfaces and negligible degree of hollowness. 
Considering sensor quantization effects such negligible degree for hollowness cannot be confidently distinguished from sensor noise under such conditions (see Section~\ref{sec:method:hollowness}); as a consequence a default hollowness value of zero is set for instances that fall in a negligible range of hollowness, i.e. below 1\si{\cm} distance between marker.

Concerning the \emph{heaviness} property, a zero variance is observed due to the accurate measurement by a scale -- considering a resolution of 1\si{\g} which is a sufficient resolution for our scenario.

\subsubsection{Property Coverage of RoCS}
\label{sec:eval:coverage}
\begin{figure}[tb]
\centering
\begin{tabular}{ccc}
 \includegraphics[width=5.5cm]{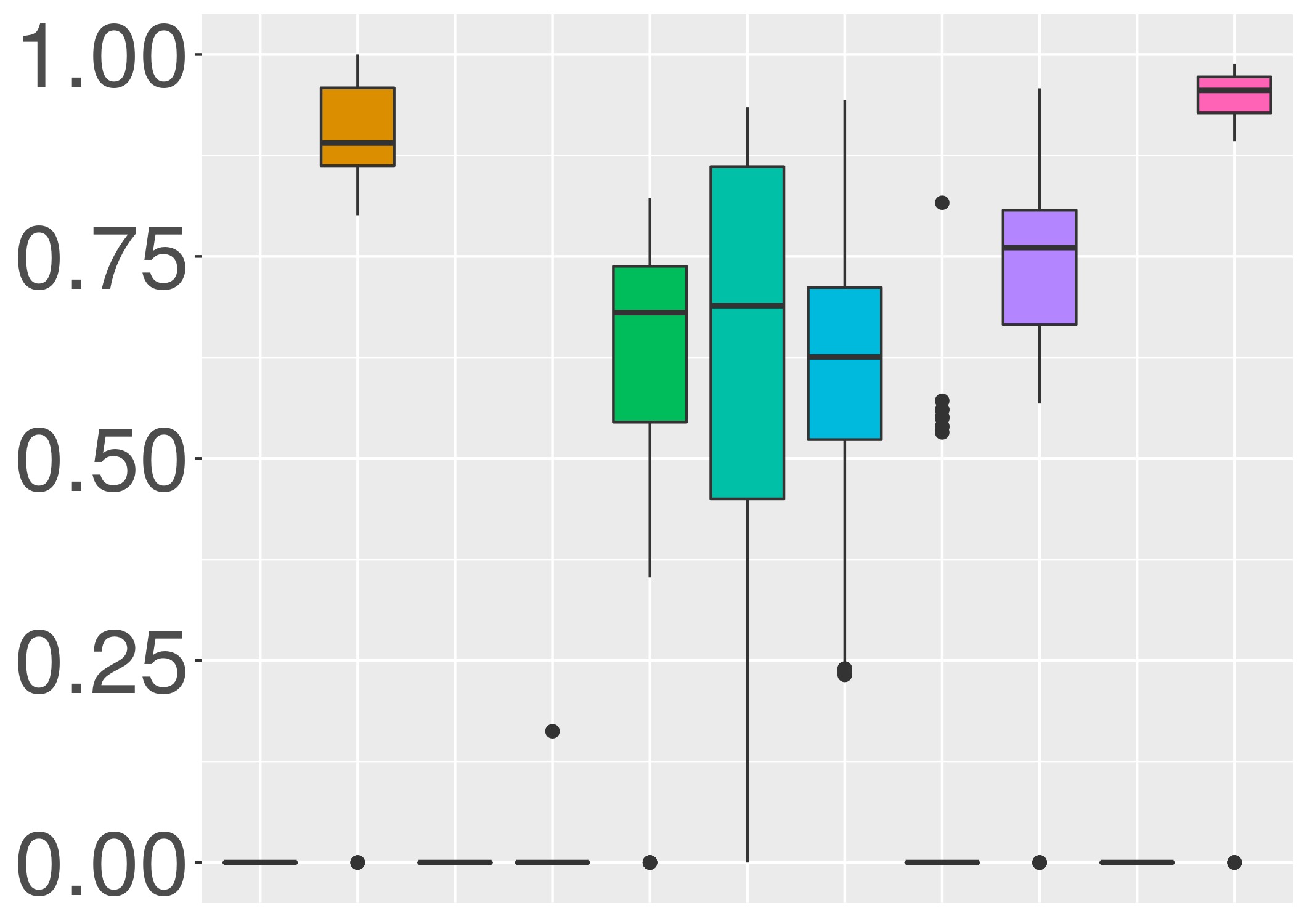} &   \includegraphics[width=5.5cm]{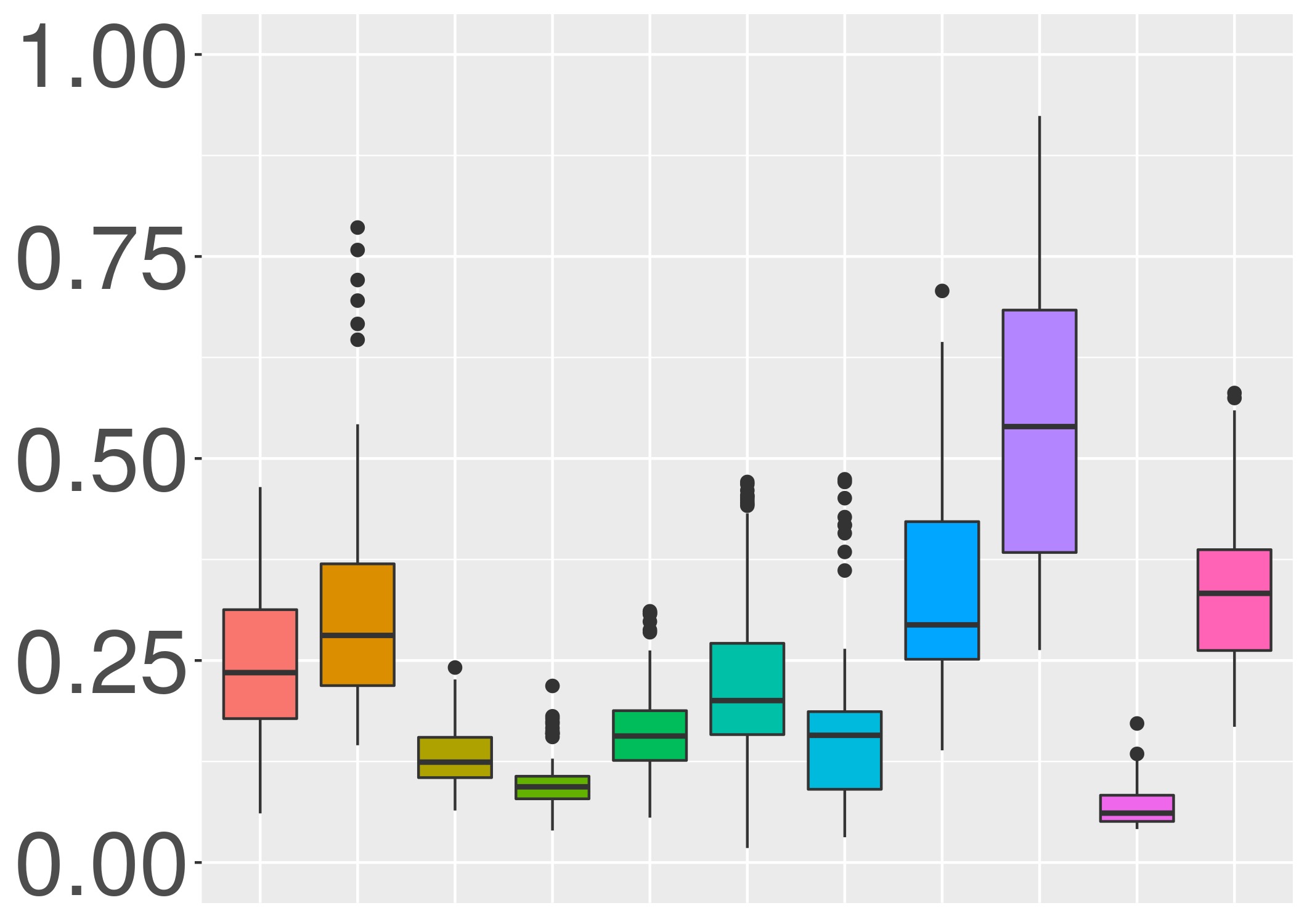} & \includegraphics[width=5.5cm]{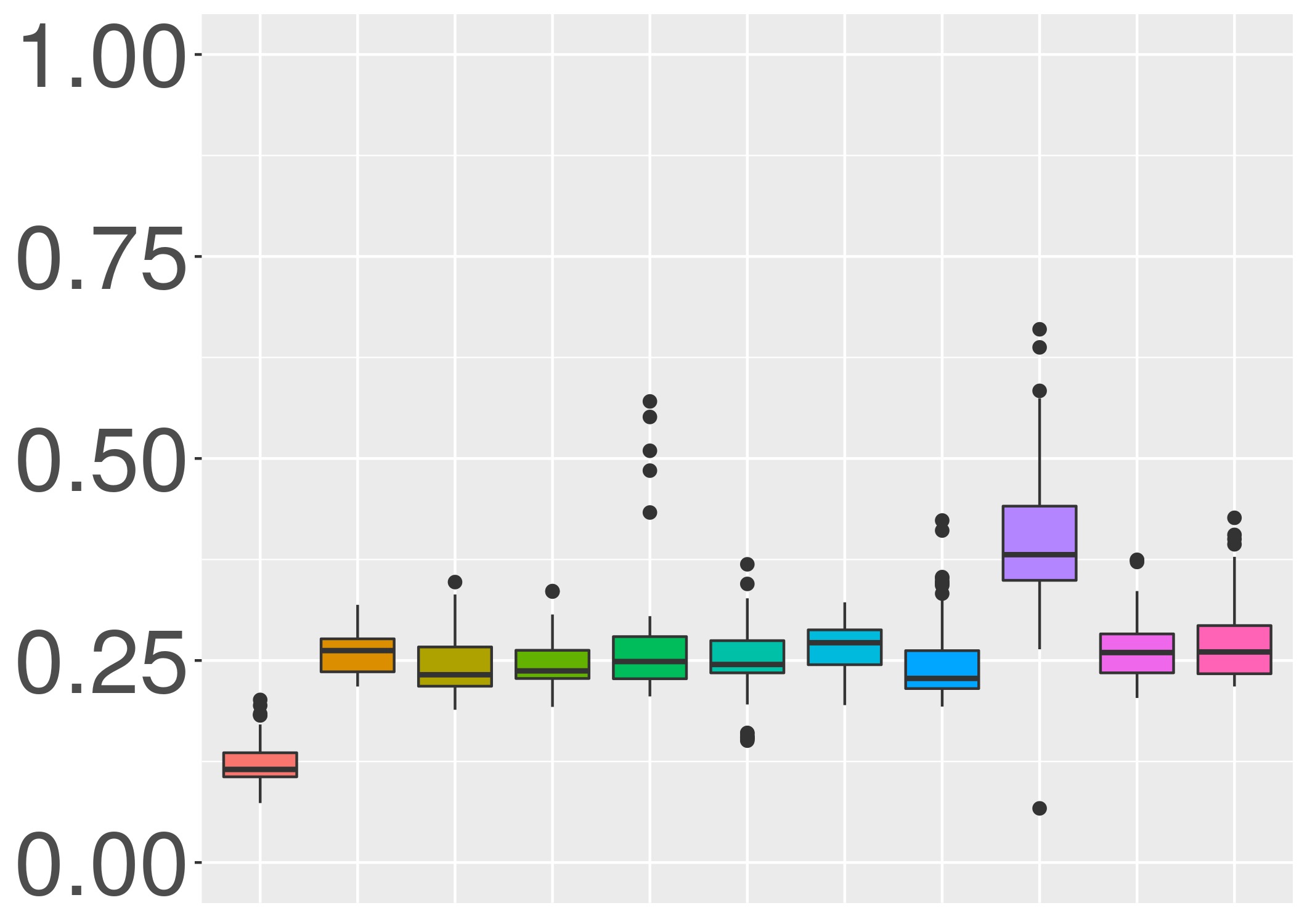} \\
\small (a) $fl$ (\emph{flatness}) & \small (b) $ri$ (\emph{rigidity}) & \small (c) $ro$ (\emph{roughness})\\[6pt]
   \includegraphics[width=5.5cm]{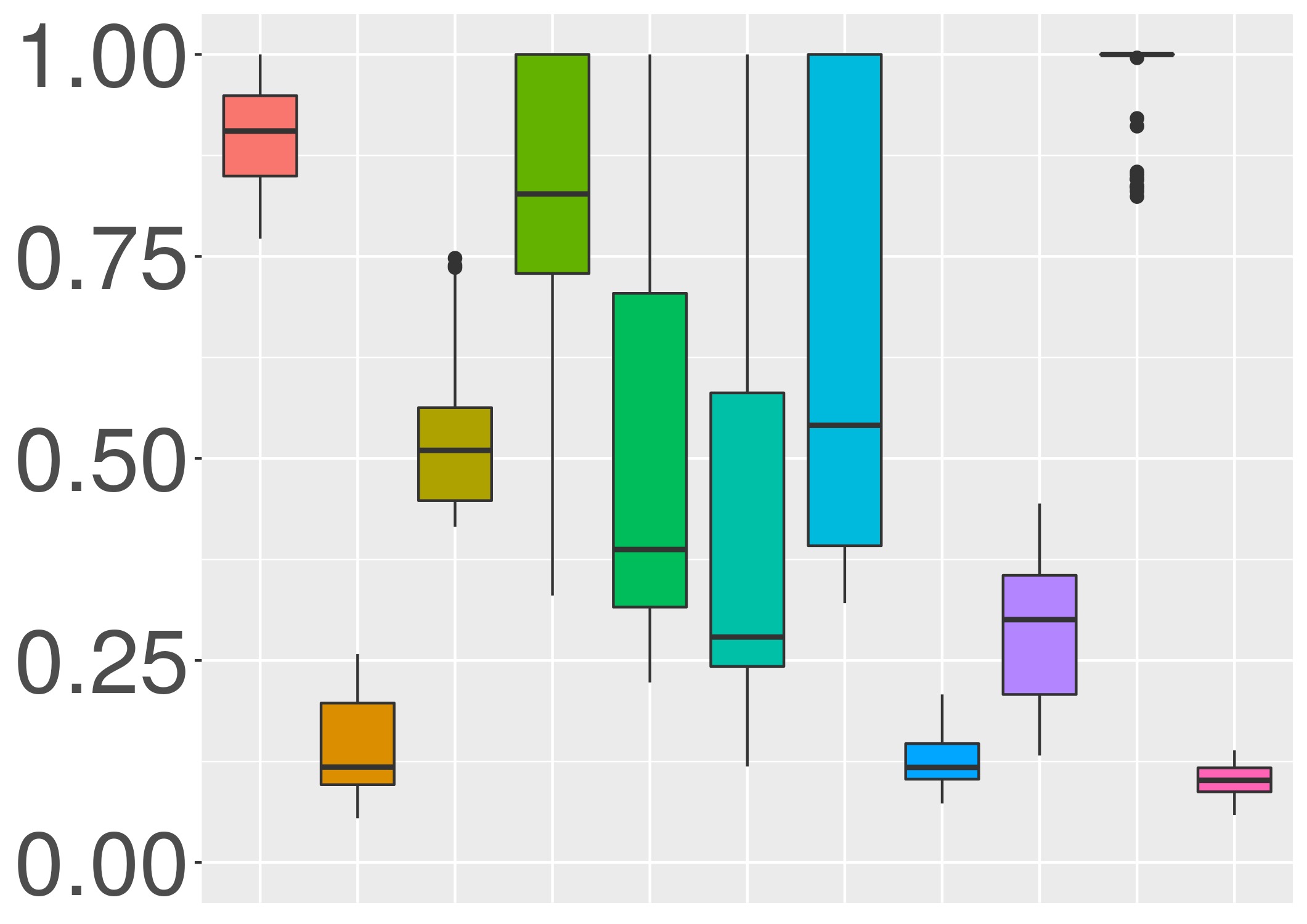} &    \includegraphics[width=5.5cm]{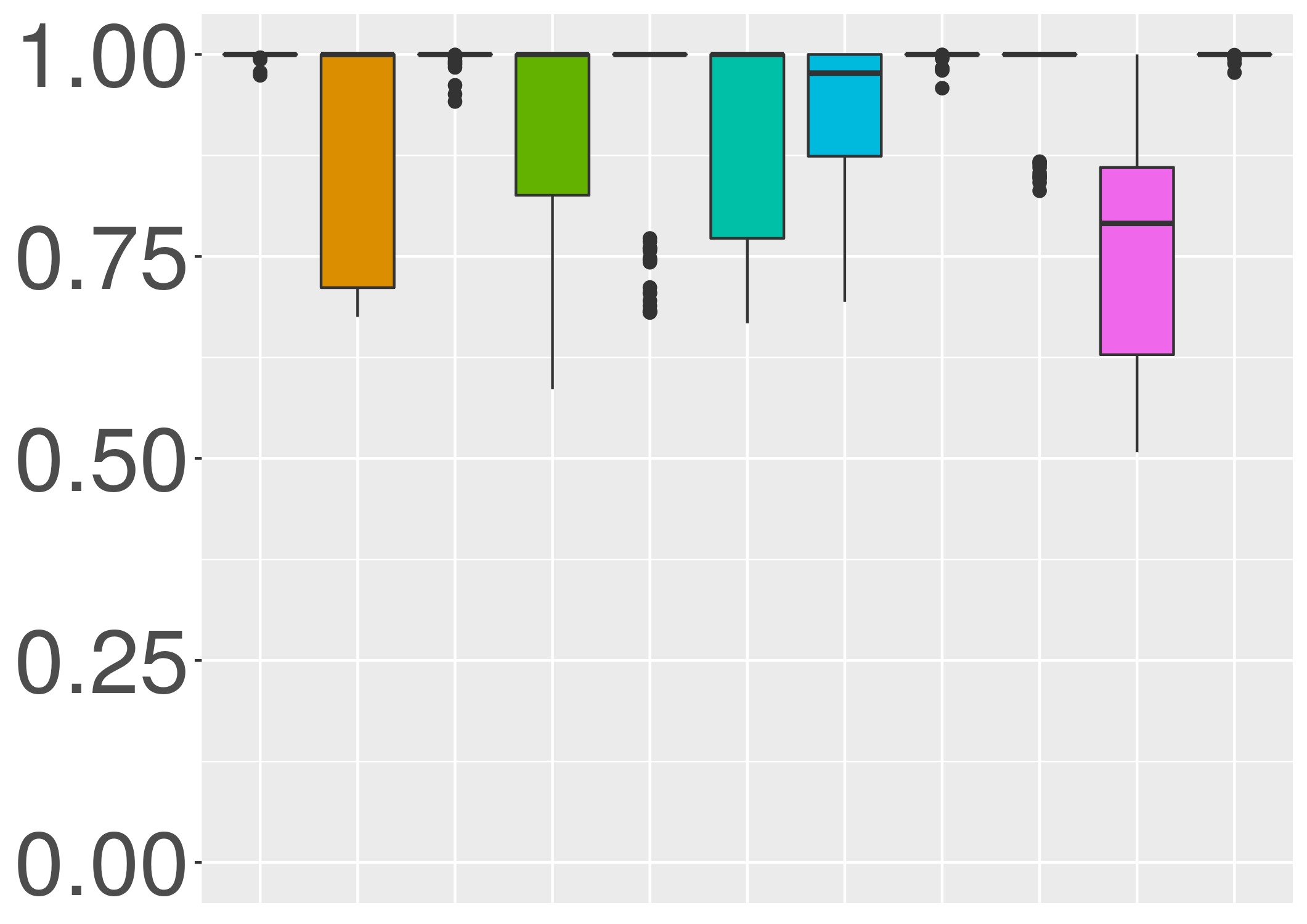} &     \includegraphics[width=5.5cm]{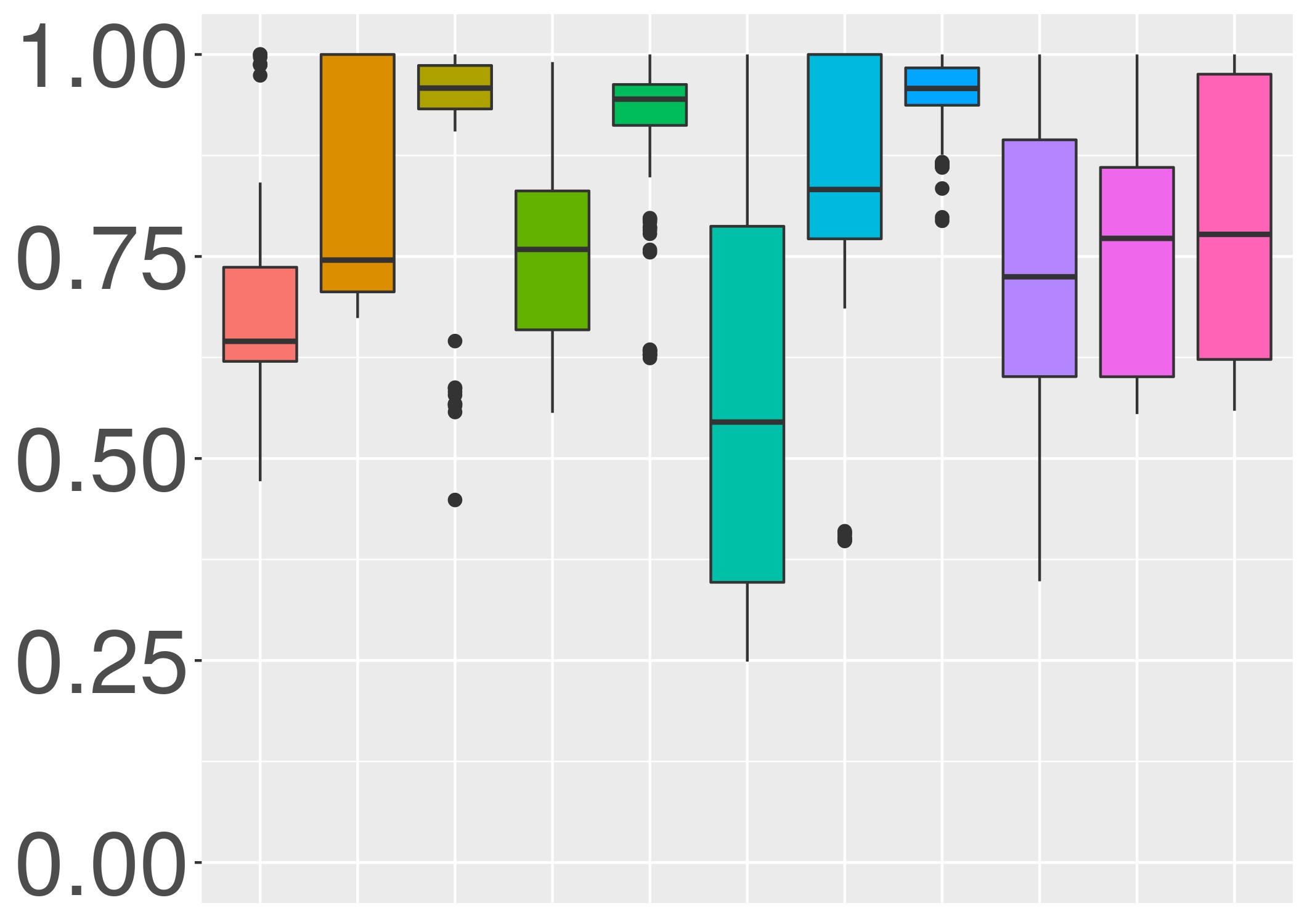} \\     
\small (d) $h\in si$ (height of \emph{size}) & \small (e) $w\in si$ (width of \emph{size})& \small (f) $l\in si$ (length of \emph{size})\\[6pt] 
\includegraphics[width=5.5cm]{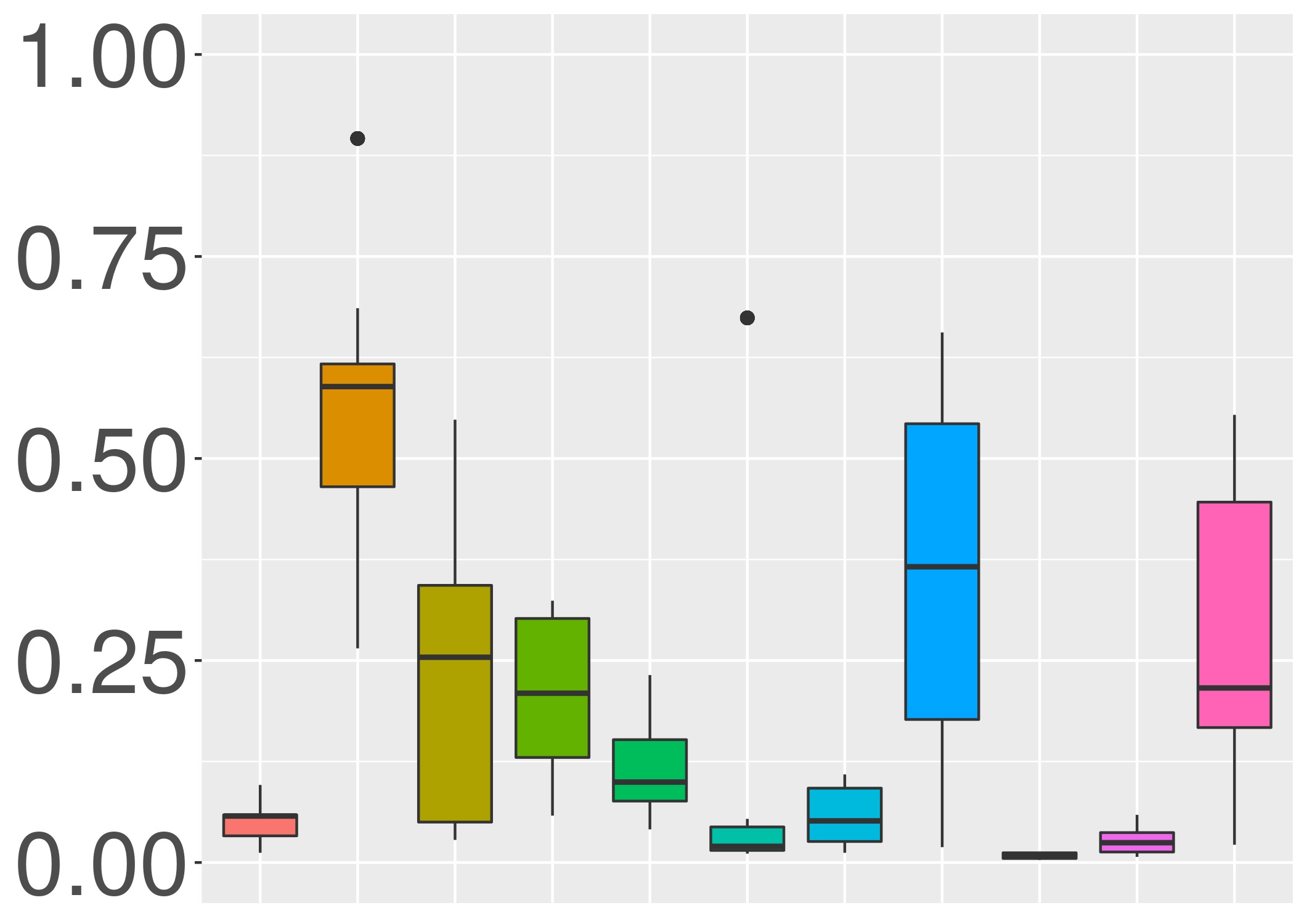} &  \includegraphics[width=5.5cm]{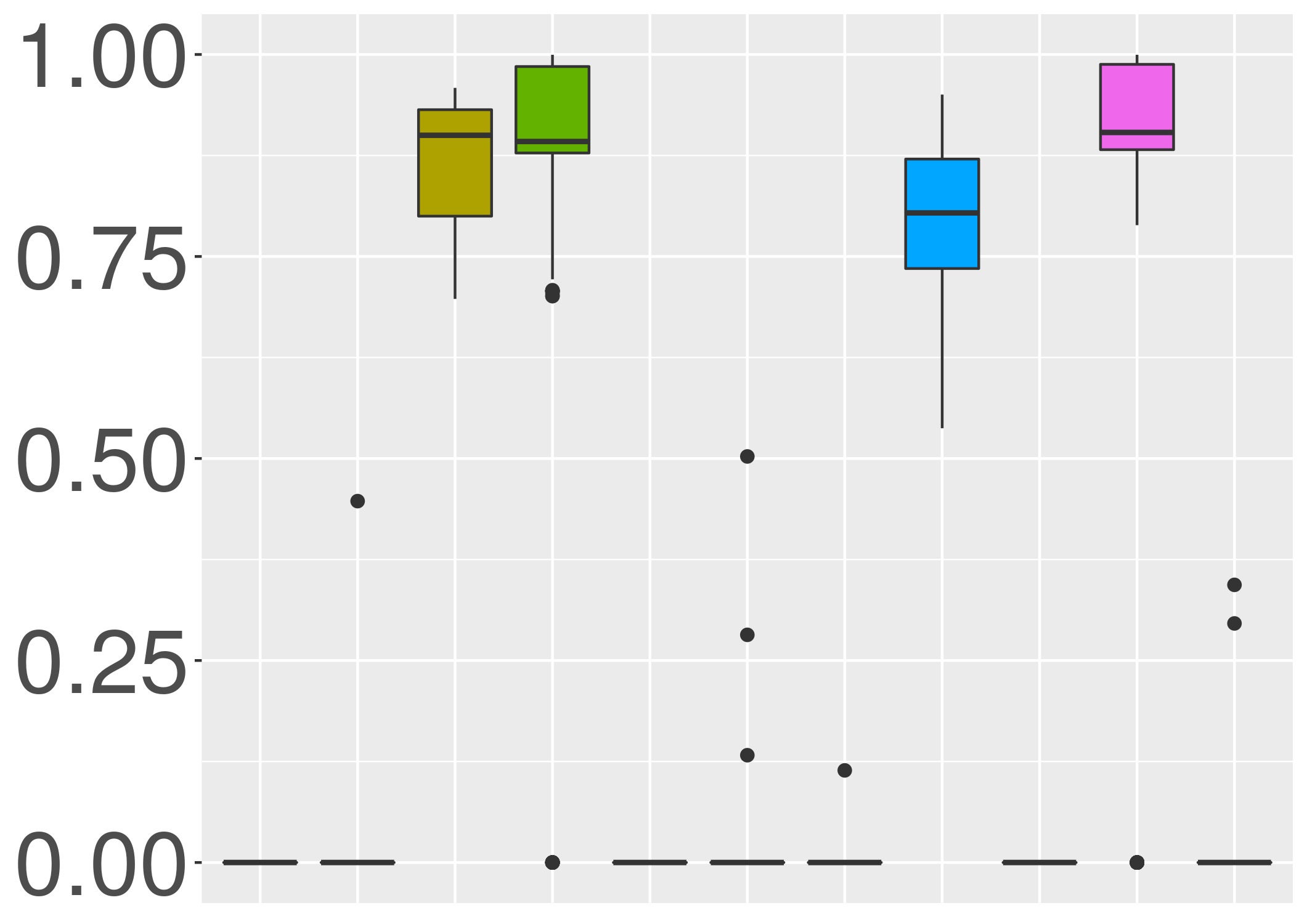} &\includegraphics[width=4.8cm]{class-legend-v2} \\
\small (g) $he$ (\emph{heaviness})  & \small (h) $ho$ (\emph{hollowness}) &  \small (i) legend \\[6pt]
\end{tabular}
\caption{Category-wise coverage for each physical property $[fl,ri,ro,si,he,ho]$.}
\label{fig:class-coverage-properties}
\end{figure}

The objective of this experiment is to evaluate the intra-class variance for each property in order to determine the range of data covered in each object class for one particular property. 
For this experiment, the mean extracted property value over the 10 repetitions is used. 
The result for each of the physical properties is shown in Fig.~\ref{fig:class-coverage-properties} in form of a box plot in which all object instances of a particular class are considered.%

Several observations can be made. 
For instance, \emph{hollowness} and \emph{flatness} are complementary in our dataset. Objects with \emph{flatness} values close to zero are commonly exhibiting increased \emph{hollowness} values (above 0.5) and vice versa. 
Only balls form an exception as they are neither flat nor hollow. 
While this means that we cover a wide range of values for the \emph{flatness} property, we miss such coverage for \emph{hollowness} values in the interval $[0,0.5]$.
Moreover, for \emph{roughness} most object classes are in a similar range -- except sponge and ball instances. 
As we place the objects in their most natural position we can conclude that the sponges' ground surfaces have a higher \emph{roughness} due to their open-pored surfaces.
Due to their roundish surfaces, ball instances feature obviously a low \emph{roughness} value. 
Furthermore, it is unlikely to observe objects featuring \emph{roughness} values close to one as none of the considered object classes has the ability to \emph{stick} to the ramp.

For the \emph{rigidity} values an interval of $[0,0.9]$ is covered, ranging from rigid objects such as \textit{metal\_box} to non-rigid objects such as \textit{sponge}. Suspiciously, only a limited number of objects has a value of zero which indicates that sensor noise has its greatest effect on these objects.

Analyzing the \emph{size} values, it becomes apparent that \emph{width} commonly is the greatest dimension among the considered objects while the objects' height varies along the range of possible values.

\subsubsection{Property Correlation}
In this experiment, we investigate the correlation in the physical properties of our data. 
Given extracted values of a particular property, we compute the mean property value $\overline{o_x}$  (Eq.~\ref{eq:pearson_prop}) over the 10 repetitions for each object instance $o$.
Based on these mean variances, the pearson correlation $\rho_{XY}$ is obtained between two sets of mean variances $X$ and $Y$ corresponding to respective properties, see Eq.~\ref{eq:pearson}, where \text{cov} is the covariance and $\sigma_x$ the standard deviation of $X$, respectively.
\begin{subequations}
\begin{align}
X&=\{\overline{o_{x_1}},\overline{o_{x_2}},\overline{o_{x_3}},...\} \label{eq:pearson_prop}\\
\rho_{XY} &= \frac{\text{cov}(X,Y)}{\sigma_x \sigma_y} \label{eq:pearson}
\end{align}
\end{subequations}
Table \ref{tab:pearson_correlation} shows the pearson correlation among all physical properties with a color scale.
\renewcommand*{\MinNumber}{-1.0}%
\renewcommand*{\MidNumber}{0.0} %
\renewcommand*{\MaxNumber}{1.0}%
\colorlet{MinColor}{red} 
\colorlet{MidColor}{white} 
\colorlet{MaxColor}{blue} 

\begin{table}[tb]
\small
\caption{Pearson Correlation on the mean values of physical properties}
\label{tab:pearson_correlation}
\centering
\setlength\tabcolsep{1.0pt} 
\begin{tabular}{r|*{7}{R}}

 & \textbf{flatness} & \textbf{rigidity} & \textbf{roughness} & \textbf{s\_length} & \textbf{s\_width} & \textbf{s\_height} & \textbf{heaviness}  \\ \hline
\textbf{flatness}     & -               &              &               &               &                  &                &                             \\
\textbf{rigidity}     & 0.45              & -               &               &                  &                 &                &                          \\
\textbf{roughness}    & 0.45              & 0.35              & -                &                  &                 &                 &                         \\
\textbf{size\_length} & 0.03              & 0.12              & 0.15               & -                   &                &                 &                              \\
\textbf{size\_width}  & 0.16              & 0.34              & 0.02               & 0.21                  & -                  &                  &                             \\
\textbf{size\_height} & -0.65             & -0.59             & -0.38              & -0.26                 & -0.45                & -                   &                          \\
\textbf{heaviness}       & 0.09              & -0.04             & -0.13              & 0.19                  & 0.02                 & -0.37                 & -                              \\
\textbf{hollowness}   & -0.71             & -0.36             & -0.08              & 0.24                  & -0.1                 & 0.24                  & 0.13       
\end{tabular}
\end{table}
It can be observed that the correlation of our data is low in general.
However, a strong negative correlation between \emph{flatness} and \emph{hollowness} is found which may indicate that in our data objects with high flatness are likely to have low hollowness. This matches our observation in Section~\ref{sec:eval:coverage}, where we noted the complementary nature of these properties in our dataset.
The object instances of our dataset may also show some negative correlation between \emph{size-height} and \emph{flatness} as well as \emph{size-height} and \emph{rigidity}. 
Since we normalize the size with the highest value of height, width and length we may conclude that in our data object instances that are higher than wide and long, are more likely to be also flat, as well as less rigid. Similarly, flat objects seem to be rigid and rough. 

\subsection{Property Semantics}
\label{sec:eval:property_semantics}
Given a stable property extraction~(Section~\ref{sec:eval:property_extraction_methods}) from noisy real world data, the following experiment focuses on the semantic interpretation of the extracted object property values.
We propose an experiment that groups object instances of our RoCS dataset in an unsupervised manner by considering a particular set of properties.
In order to conduct an preferably unbiased (machine-driven) grouping, \emph{k}-means clustering is applied with a gradually increasing value of \emph{k}=$\{2,...,11\}$. Here, 11 is selected as upper bound as it represents the number of classes incorporated in the RoCS dataset.

\begin{figure}[tb]
\begin{tabular}{ccc}
\begin{tikzpicture}[overlay]
\node[] at (0,1.6){\includegraphics[width=4.5cm]{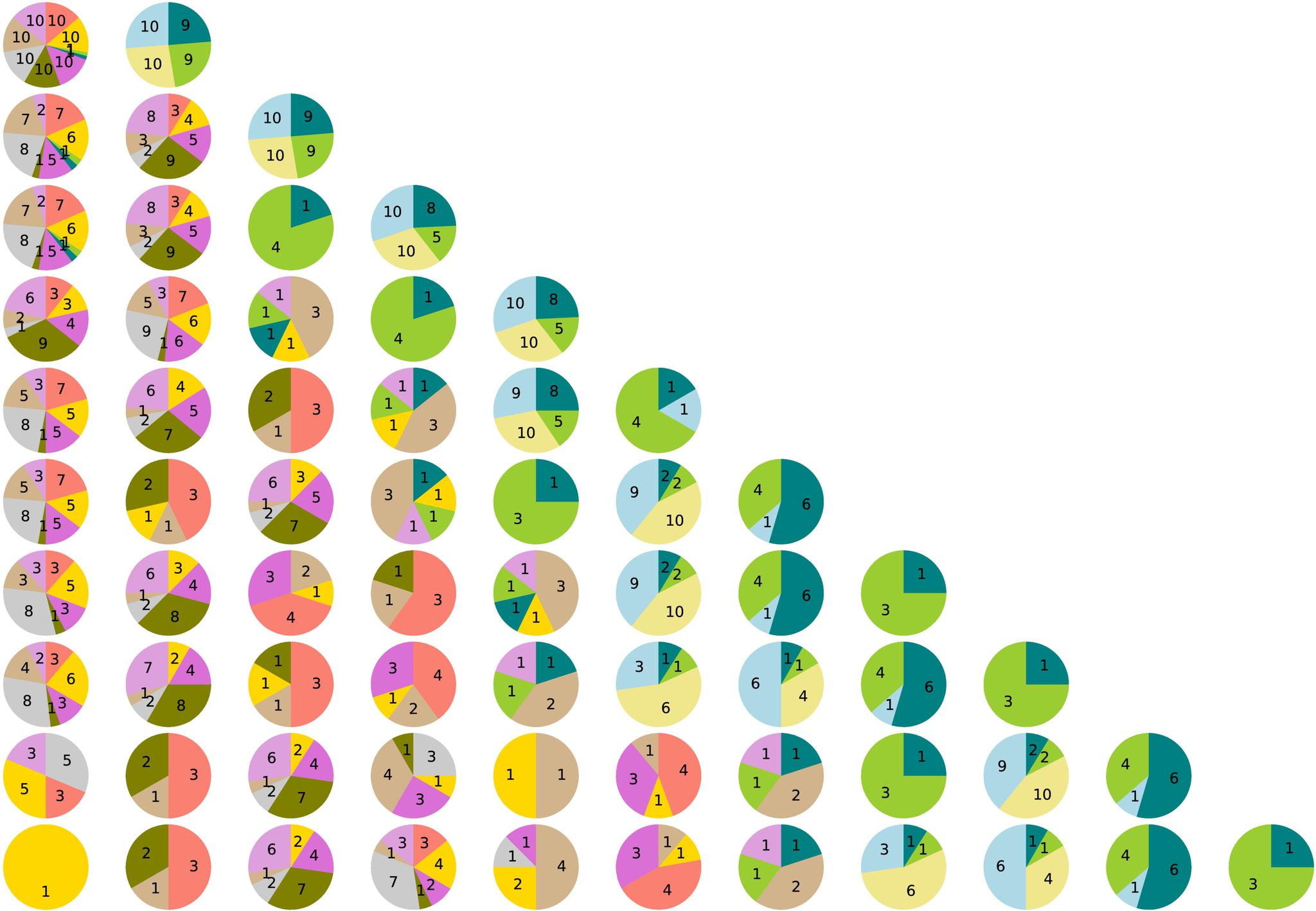}};
\node[circle, draw=blue!80, line width=0.5mm, inner sep=0pt, minimum size=10pt] (annotation) at (-0.42,2.06) {};
\end{tikzpicture}
 &   \includegraphics[width=4.5cm]{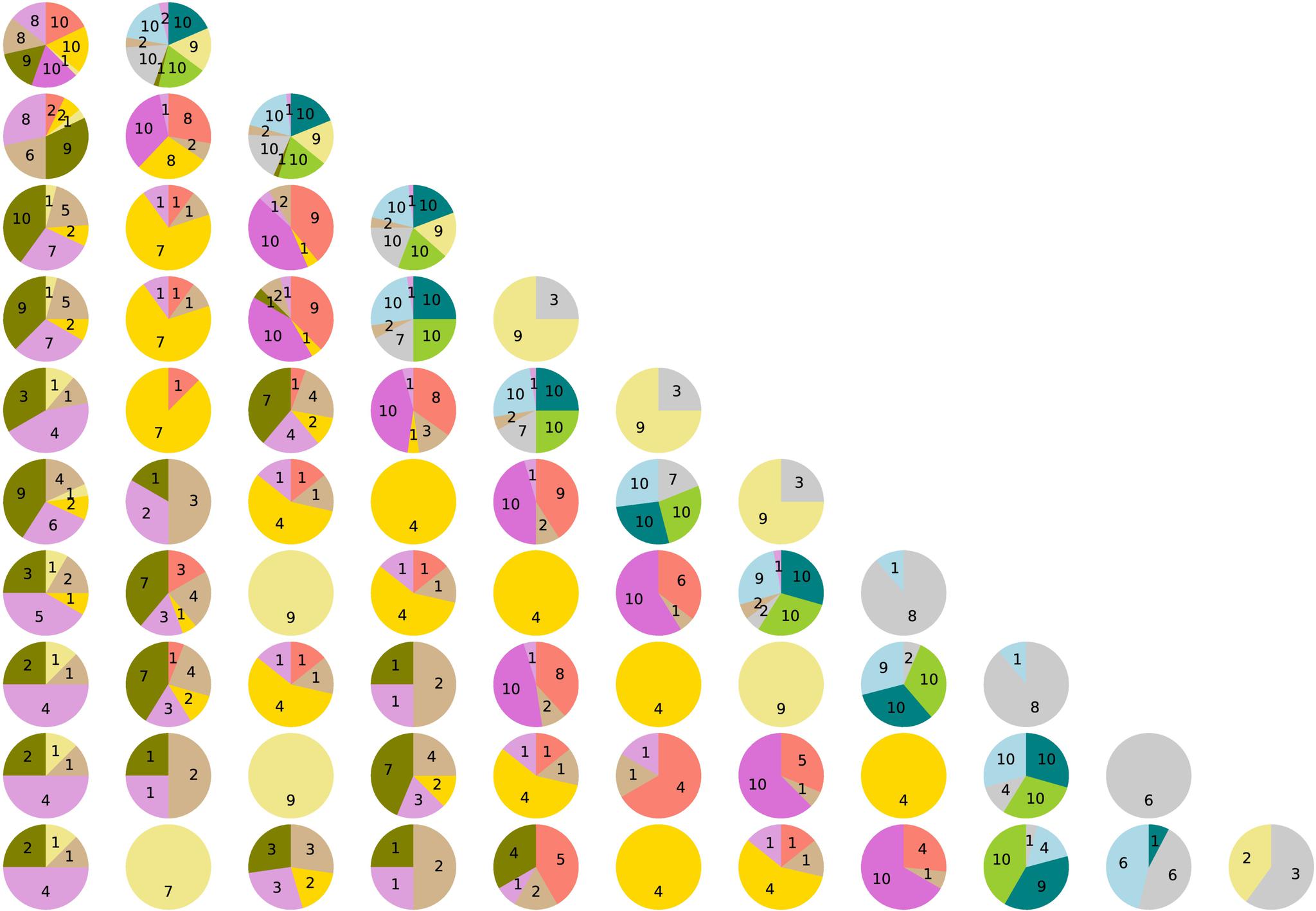} & \includegraphics[width=4.5cm]{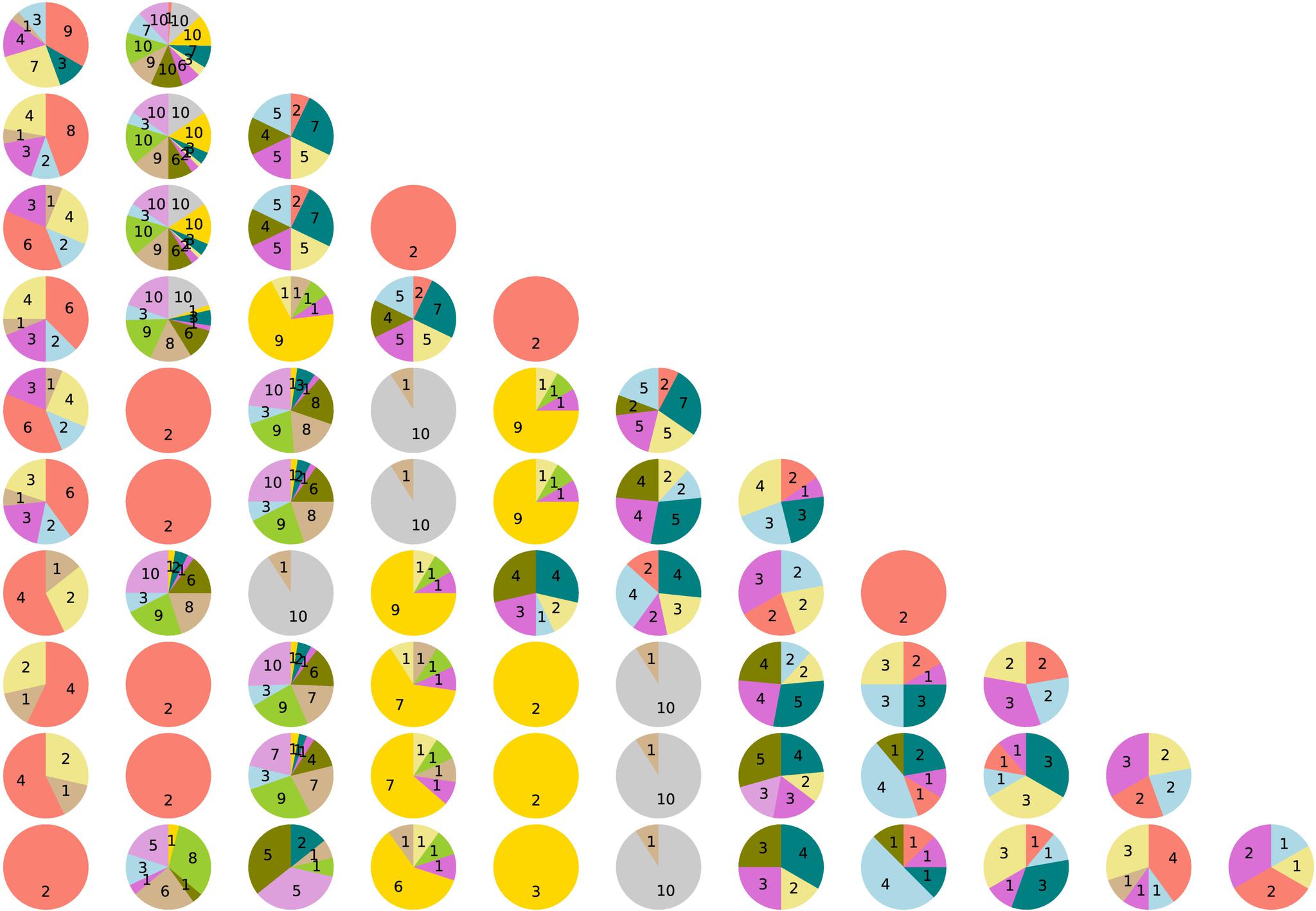} \\
\small (a) \emph{containment} $[co]$ & \small (b) \emph{support} $[su]$&\small (c)  \emph{movability} $[mo]$\\[6pt]
   \includegraphics[width=4.5cm]{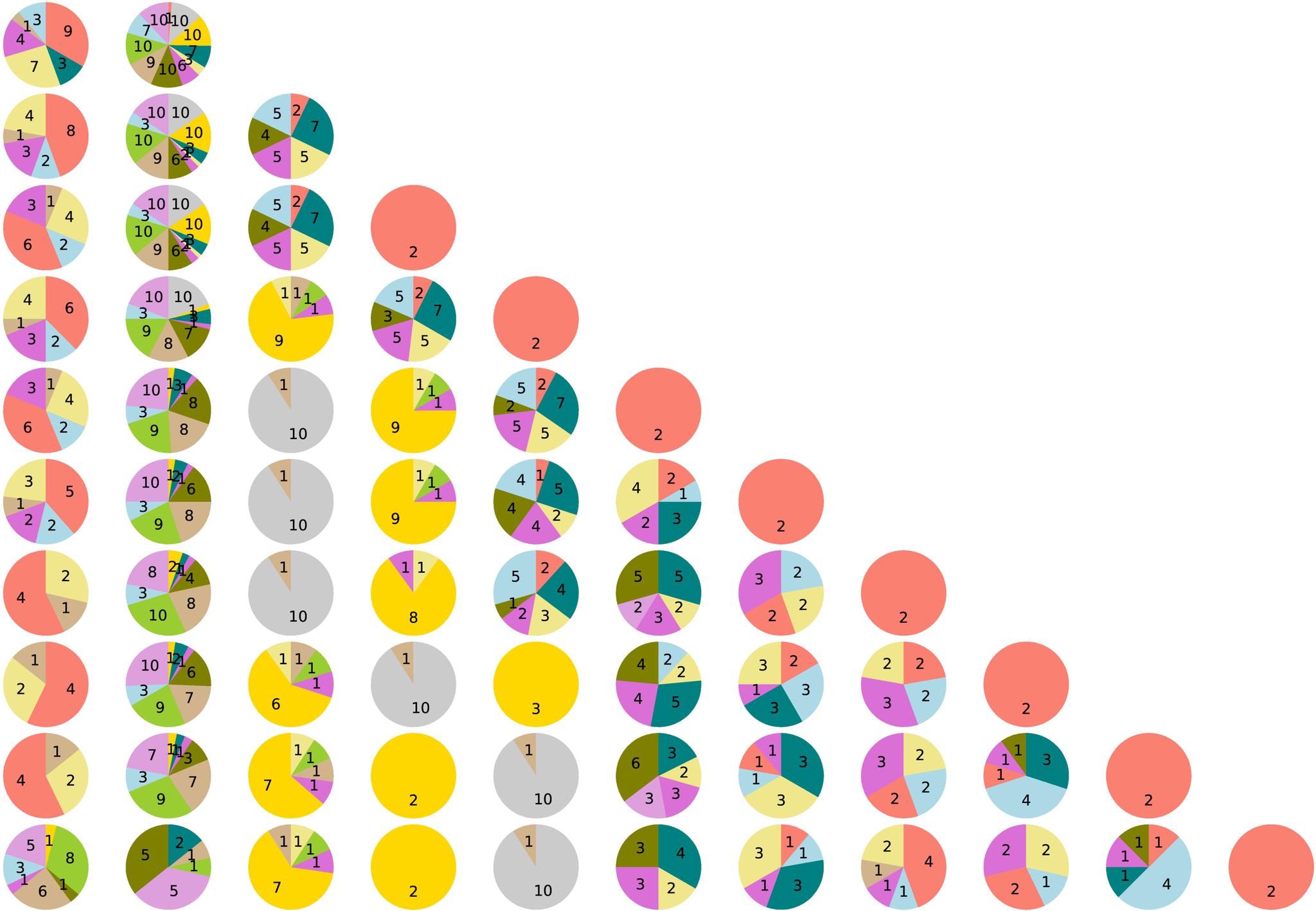} &  
\begin{tikzpicture}[overlay]
\node[] at (0,1.6){\includegraphics[width=4.5cm]{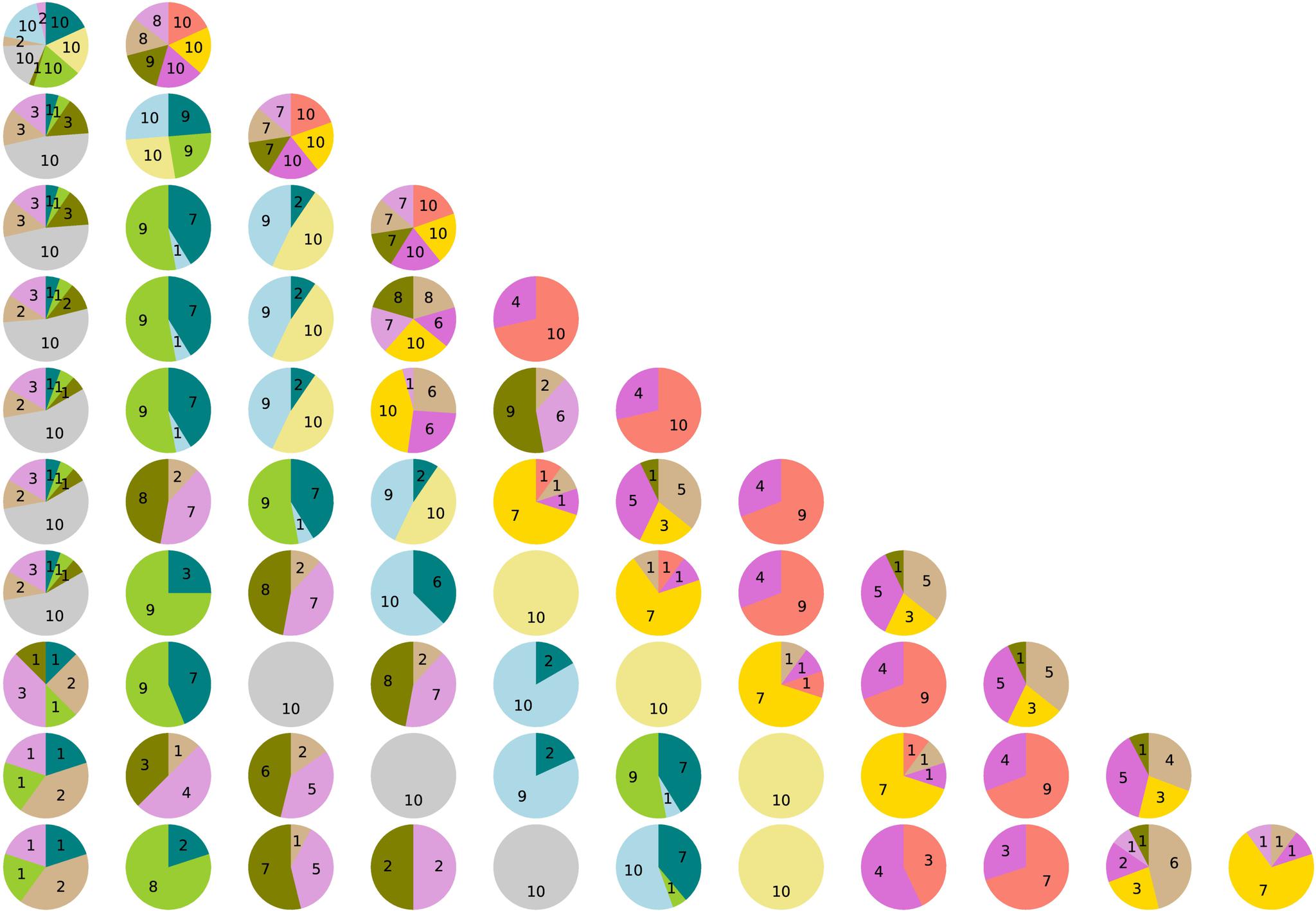}};
\node[circle, draw=blue!80, line width=0.5mm, inner sep=0pt, minimum size=10pt] (annotation) at (-0.415,1.75) {};
\end{tikzpicture}
   &     \includegraphics[width=2.0cm]{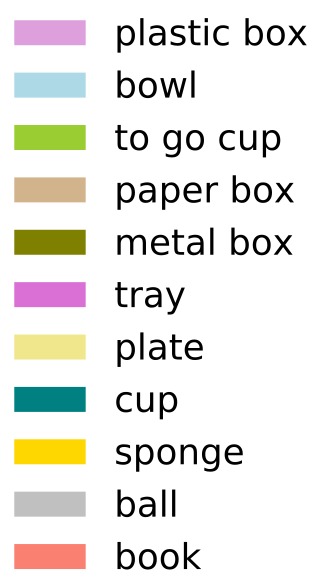} \\     
\small (d)  \emph{blockage} $[bl]$ & \small (e)  physical properties $[si,fl,ho,he,ri,ro]$ & \small(f) legend \\[6pt] 
\end{tabular}
\caption{Gradual partitioning of instances to particular concepts given a particular set of properties describing each instance. Each concept is illustrated as a pie chart showing the object class label distribution of instances assigned to the respective concept. Sample concepts are annotated (\tikzdrawcircle[blue!80, line width=0.5mm,fill=white, minimum size=10pt]{3pt}) which illustrate object classes featuring similar quality regarding the property, such as \emph{plate}, \emph{bowl}, \emph{cup}, \emph{to\_go\_cup} regarding the \emph{containment} property.}
\label{fig:pyramid}
\end{figure}

Figure~\ref{fig:pyramid} shows the gradual partitioning process for the respective property. A group is depicted as a pie-chart illustrating the distribution of assigned instances according to their labeled class. Therefore, each row of the pyramid-like structure shows the results of one application of the \emph{k}-means clustering. The number of pie-charts in each row equals to \emph{k}.
Furthermore, since each group partitions the property space, assigned instances within the group feature similar attributes
Therefore, a group can be interpreted as concept of a particular quality of the respective property.

The results shown in Fig.~\ref{fig:pyramid} reveal the gradual partitioning of instances to particular concepts over the increase of \emph{k}.
Generally on higher levels (lower \emph{k}) concepts appear to feature more generic attributes since the distribution of classes is higher compared to lower levels (higher \emph{k}). Consequently, lower levels encompass concepts featuring more specific attributes.
Moreover, we may observe semantic relations between class labels and observed concepts, e.g. instances of \emph{plate}, \emph{bowl}, \emph{cup}, \emph{to\_go\_cup} share a similar quality regarding the \emph{containment} property (see concept annotated with \tikzdrawcircle[blue!80, line width=0.5mm,fill=white, minimum size=10pt]{3pt} in Fig.~\ref{fig:pyramid}(a)) which is also reflected in form of a concept that encompasses these object classes over multiple levels.
Such qualities can also be observed and tracked over multiple levels for other functional properties in Fig.~\ref{fig:pyramid}.
Furthermore, Fig.~\ref{fig:pyramid}(e) illustrates the gradual grouping process considering all physical properties of the object instances.
Also here such patterns can be observed, e.g. on the right side where concepts have emerged that feature common properties related to instances such as \emph{plastic\_box}, \emph{metal\_box}, \emph{paper\_box}  (see concept annotated with \tikzdrawcircle[blue!80, line width=0.5mm,fill=white, minimum size=10pt]{3pt} in Fig.~\ref{fig:pyramid}(e)).

As a result, this experiment has revealed the \emph{generality} of the proposed property hierarchy.
The proposed property hierarchy (Fig.~\ref{fig:property_hierarchy}) may allow to \emph{describe} object instances encompassing a variety of characteristics -- from appearance to functional purpose -- and also allow to \emph{discriminate} these instances by associating them to meaningful groups featuring similar object concepts.
By design, property generality can be observed across object classes, i.e. concepts on different granularity levels may feature dedications to instances of different object classes as they feature similar characteristics regarding the property.
This interrelation of object classes is reflected by the heterogeneity of the distribution of instances within a concept -- even in case of $k$=11 when considering 11 object classes.
These observations made in the proposed property acquisition procedure may provide a basis for the generation of conceptual knowledge about objects as shown in Section~\ref{sec:eval:tool_substitution}.

\subsection{Conceptual Knowledge for Tool Substitution}
\label{sec:eval:tool_substitution}
\begin{figure}[h!tb]
\begin{tabular}{cc}
\includegraphics[width=7.0cm]{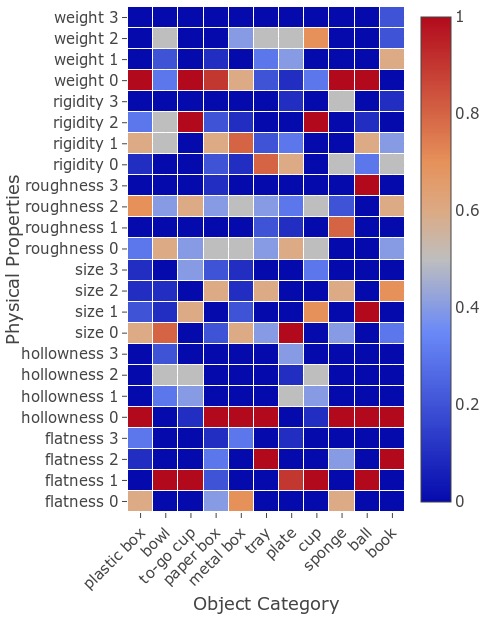} &   \includegraphics[width=7.3cm]{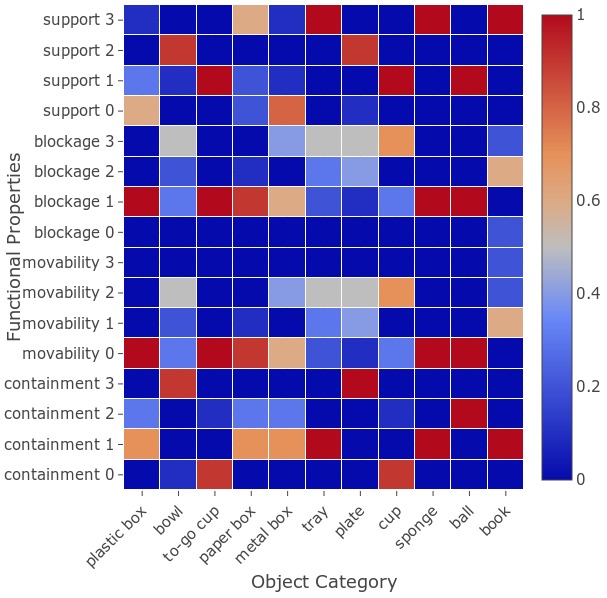} \\
\small (a) Knowledge about Physical Properties \vspace{2mm} & \small (b) Knowledge about  Functional Properties \vspace{2mm} \\
\includegraphics[width=7cm]{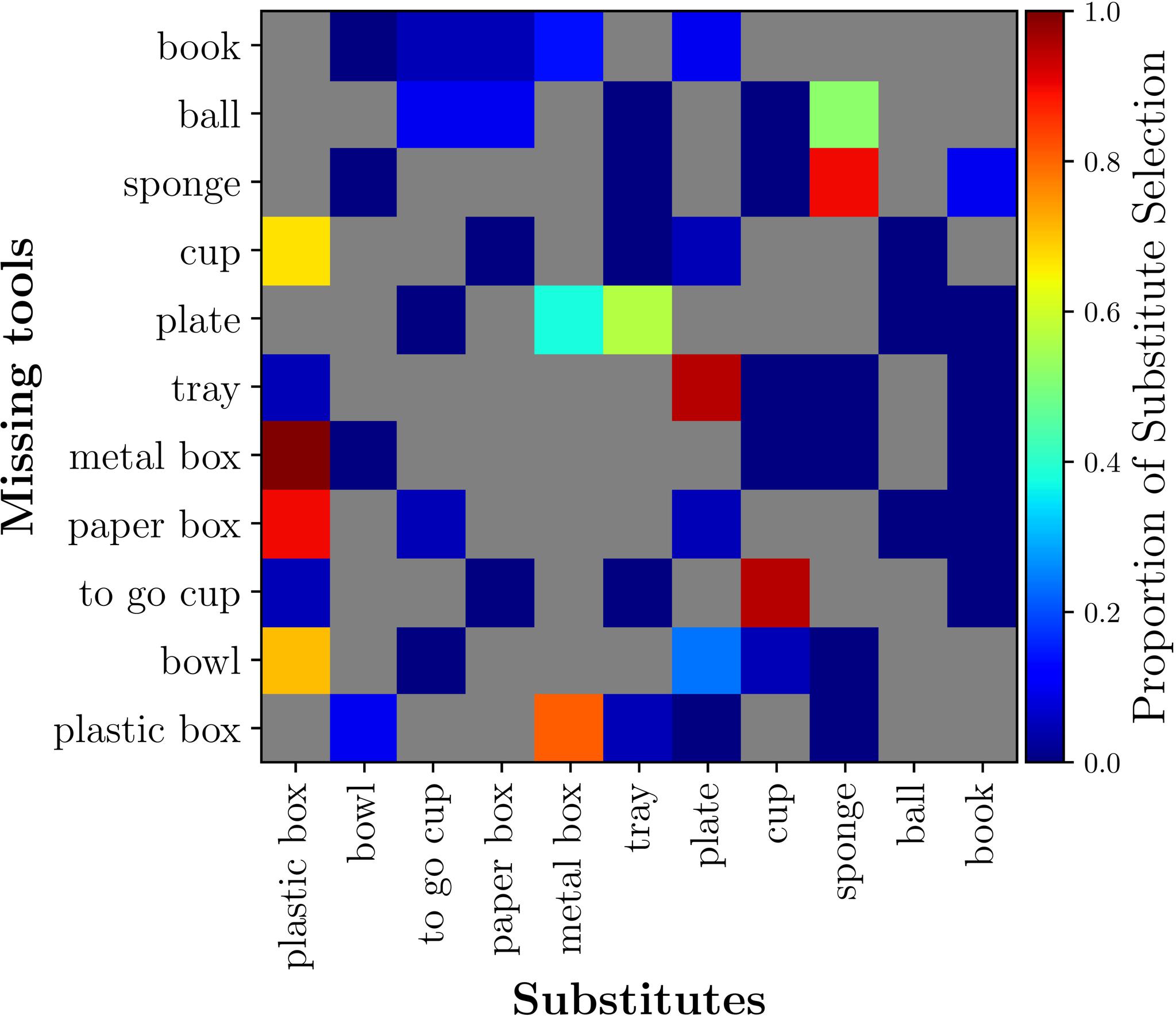} &   \includegraphics[width=7cm]{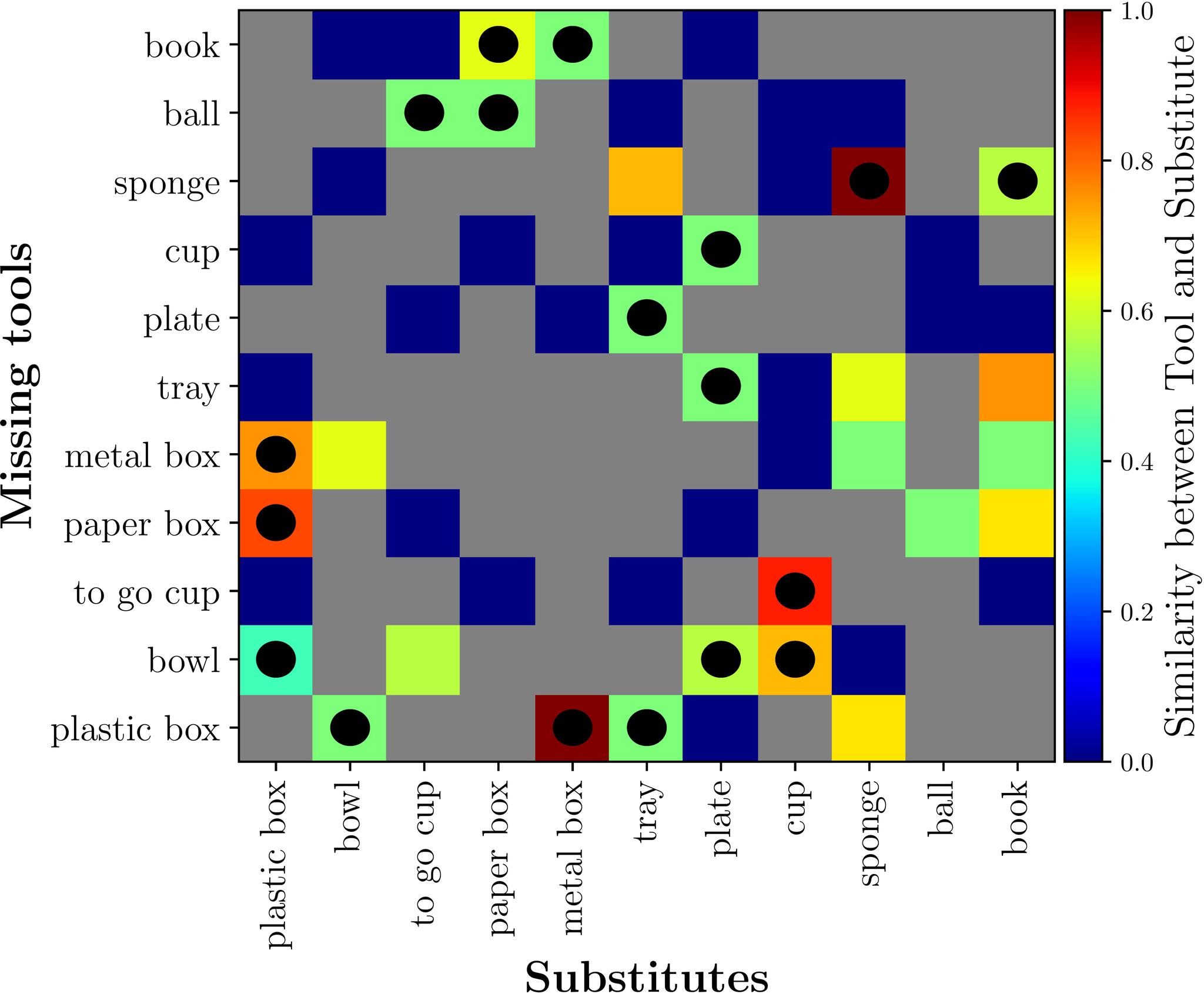} \\
\small (c) Selection by experts & \small (d) Selection by ERSATZ compared with\\
& experts' selections \\
\end{tabular}
\caption{Substitution results w.r.t. human expert selection \emph{distribution} and ERSATZ \emph{similarity} responses. Note that, gray cells correspond to object categories which are not available in the respective query, cells marked with \tikzdrawcircle[black, fill=black]{2.5pt} represents substitutes selected by experts and ERSATZ.}
\label{fig:ersatz}
\end{figure}

In this experiment, we demonstrate how the robot-centric conceptual knowledge grounded in the robot's sensory data was used to determine a substitute in a tool substitution scenario.
While operating in a dynamic environment, a robot can not assume that a particular tool required in a task will always be available.
In such scenarios, an ideal solution for a robot would be to improvise by finding a substitute for the missing tool as humans do.
This skill is significant when operating in a dynamic, uncertain environment because it allows a robot to adapt to unforeseen situations.
To deal with such situation, we have developed an approach, called as ERSATZ (German word for a substitute) detailed in \cite{thomuezug_2018} which is inspired by way in which humans select a substitute in a non-invasive manner.
In this approach, the robot-centric conceptual knowledge about objects is used to select a plausible substitute from the available objects.
For the experiments, we generated knowledge about 11 object classes using the approached discussed in the section \ref{sec:knowledge_generation}.
The dataset generated by RoCS was utilized for creating robot-centric conceptual knowledge about 11 object classes.
The Figure \ref{fig:ersatz}(a) and \ref{fig:ersatz}(b) illustrates graphically the qualitative knowledge about physical and functional properties of 11 object classes.

The X-axis and Y-axis in the figure contain the labels of the object classes and the qualitative measures of the physical or functional properties respectively.
The colored cells indicate the sample proportion of each qualitative measure of a property in the instances of an object class.
For example, a snippet of knowledge about object class \textit{book} can be stated as, 
\{$\langle$\textit{book, flatness\_0, 1.0}$\rangle$, $\langle$\textit{book, size\_0, 0.7}$\rangle$,
$\langle$\textit{book, size\_3, 0.3}$\rangle$, $\langle$\textit{book, roughness\_2, 0.6}$\rangle$,
$\langle$\textit{book, containment\_0, 0.7}$\rangle$\}, $\langle$\textit{book, movability\_1, 0.6}$\rangle$\},
where the numbers indicate that, for instance, physical quality \textit{roughness\_2} was observed in 60\% instances of object class \textit{book}.

For the tool-substitution experiment, we generated $11$ queries based on the $11$ object classes, where each query consisted of a missing tool and $5$ randomly selected objects as available choices for a substitute.
The queries were given to $21$ human experts and were asked to select a substitute in each scenario.
The expert selections were aggregated and selection proportion was calculated for each expert-selected substitute.
ERSATZ used the knowledge generated in the previous section and computed substitute/s for each given scenario using the approach discussed in \cite{thomuezug_2018}.
In order to validate a substitutability, the number of selected substitutes by human experts was then compared with the number of selected substitutes by ERSATZ.
We plotted the sample proportion of expert selections of a substitutes in each scenario as a heat map where the colored cells denote the sample proportions (see Fig.~\ref{fig:ersatz}(c)).
On the other hand, the Fig.~\ref{fig:ersatz}(d) illustrates the selection of substitutes by colored cells representing the similarity between a missing tool and a substitute.
The black solid circles denote the selection made by humans and ERSATZ.
Our experiments showed that in all of $11$ substitution scenarios, human experts and ERSATZ selected the similar substitutes.

\section{Conclusion}
Conceptual object knowledge is desired in various robotic scenarios (from household to industrial robotics) in order to efficiently perform tasks when dealing with objects (e.g. tools) in uncertain environments (e.g., home service, factory of the future, inspection).
Furthermore, it is a necessary prerequisite for efficient tool-use.
However, state-of-the-art conceptual knowledge approaches are generally hand-crafted and generated from a human perspective in form of \textbf{natural language concepts}.
Consequently, the discrepancy between human and robotic capabilities (e.g. visual, auditory, haptic perception, prior knowledge, etc.) is also reflected in the knowledge generation process conducted by humans and the interpretation for it by robotic systems.
In order to mitigate this discrepancy, we proposed a robotic-centric approach as we believe that conceptual object knowledge has to be generated from the robotic-perspective considering the robotic capabilities, so to say in from of \textbf{robotic language concepts}.

A multi-modal object property extraction and robot-centric knowledge generation process has been proposed to acquire conceptual object knowledge from a robotic perspective.
We introduced a bottom-up knowledge acquisition process, from capturing sensory data over a numeric extraction of object properties, to a symbolic conceptualization of objects' properties.
Experiments have revealed the \emph{stability} as well as the \emph{generality} of the proposed object property acquisition procedure.
This outcome provides a basis for the following conceptual knowledge generation in the context of \emph{tool substitution}.
Tool substitution results have demonstrated the applicability of the generated conceptual knowledge.
We conclude, that the proposed robot-centric and multi-modal conceptualization approach may contribute to equip a robot with the capability to reason about objects on a conceptual level compared to general approaches which only base on e.g. visual (image pixels) or haptic (resistance feedback) sensory data.
Moreover, such robot-centric and multi-modal knowledge can be applied to a variety of scenarios beyond tool substitution, thus we established the RoCS dataset and made it publicly available.

As the goal of this work was a generic conceptual knowledge generation, our future work is directed, for instance, towards the \emph{transfer} of such knowledge among heterogeneous robotic systems.
Furthermore, after making the framework publicly available, we wish to create an online community where researchers can contribute to the framework by adding further object data or providing variations of the extraction methods.

\section*{Acknowledgments}
We would like to thank our colleagues and students Florian Sommer, David Döring and Saagar Gaikwad from Otto-von-Guericke University Magdeburg, Germany for providing assistance in the dataset evaluation and visualization. 

%\section*{Author Contributions Statement}
%MT primarily contributed to the conception and preliminary design of the robot-centric knowledge acquisition framework; MT, CM and GJ contributed to the property selections and definitions of RoCS framework; CM and GJ contributed to the implementation of the framework and extraction methods; NP, RC and SJ contributed to the dataset generation of 110 objects; CM, GJ and JS contributed in the integration and the evaluation of the dataset; MT, CM, GJ and MP contributed to evaluation of the property semantics; MT contributed to the conceptual knowledge generation and application of the conceptual knowledge in tool substitution scenario; AB, MP and SZ contributed to the critical evaluation of the work, All authors contributed to manuscript revision, read and approved the submitted version.  

%\section*{Data Availability Statement}
%The datasets [GENERATED/ANALYZED] for this study can be found %at:\\  \url{https://gitlab.com/rocs_data/rocs-dataset}

\small
\bibliographystyle{plain}

\begin{thebibliography}{10}

\bibitem{Baber2003_1}
Christopher Baber.
\newblock {Introduction}.
\newblock In {\em Cognition and Tool Use}, chapter~1, pages 1--15. Taylor and
  Francis, 2003.

\bibitem{Baber2003}
Christopher Baber.
\newblock {\em {Cognition and Tool Use}}.
\newblock Taylor and Francis, 2003.

\bibitem{Baber2003_6}
Christopher Baber.
\newblock {The Design of Tools}.
\newblock In {\em Cognition and Tool Use}, chapter~6, pages 69--80. Taylor and
  Francis, 2003.

\bibitem{Baber2003_5}
Christopher Baber.
\newblock {Working With Tools}.
\newblock In {\em Cognition and Tool Use}, chapter~5, pages 51--68. Taylor and
  Francis, 2003.

\bibitem{BennettCialone14}
Brandon Bennett and Claudia Cialone.
\newblock {Corpus Guided Sense Cluster Analysis: a methodology for ontology
  development (with examples from the spatial domain)}.
\newblock In Pawel Garbacz and Oliver Kutz, editors, {\em 8th International
  Conference on Formal Ontology in Information Systems (FOIS)}, volume 267 of
  {\em Frontiers in Artificial Intelligence and Applications}, pages 213--226.
  IOS Press, 2014.

\bibitem{Biro2013}
Dora Biro, Michael Haslam, and Christian Rutz.
\newblock {Tool use as adaptation.}
\newblock {\em Philosophical transactions of the Royal Society of London.
  Series B, Biological sciences}, 368(1630), 2013.

\bibitem{Bischoff2011}
R.~Bischoff, U.~Huggenberger, and E.~Prassler.
\newblock Kuka youbot - a mobile manipulator for research and education.
\newblock In {\em 2011 IEEE International Conference on Robotics and
  Automation}, pages 1--4, May 2011.

\bibitem{Boesch2013_2}
Christophe Boesch.
\newblock {Ecology and cognition of tool use in chimpanzees}.
\newblock In Josep Boesch~Christophe Sanz, Crickette M.~Call, editor, {\em Tool
  Use in Animals: Cognition and Ecology}, chapter~2, pages 21--47. Cambridge
  University Press, 2013.

\bibitem{Coradeschi2003}
Silvia Coradeschi and Alessandro Saffiotti.
\newblock {An introduction to the anchoring problem}.
\newblock {\em Robotics and Autonomous Systems}, 43(2-3):85--96, 2003.

\bibitem{Daoutis2009}
Marios Daoutis, Silvia Coradeshi, and Amy Loutfi.
\newblock {Grounding commonsense knowledge in intelligent systems}.
\newblock {\em Journal of Ambient Intelligence and Smart Environments},
  1(4):311--321, 2009.

\bibitem{Davis1993}
Randall Davis, Howard Shrobe, and Peter Szolovits.
\newblock {What Is a Knowledge Representation ?}
\newblock {\em AI Magazine}, 14:17--33, 1993.

\bibitem{Emery2013_4}
Nathan~J. Emery.
\newblock {Insight, imagination and invention: Tool understanding in a
  non-tool-using corvid}.
\newblock In Josep Boesch~Christophe Sanz, Crickette M.~Call, editor, {\em Tool
  Use in Animals: Cognition and Ecology}, chapter~4, pages 67--88. Cambridge
  University Press, 2013.

\bibitem{Fellbaum1998}
Christiane Fellbaum, editor.
\newblock {\em {WordNet: An Electronic Lexical Database}}.
\newblock The MIT Press, Cambridge, MA ; London, 1998.

\bibitem{GarridoJurado2014}
S.~Garrido-Jurado, R.~Munoz-Salinas, F.J. Madrid-Cuevas, and M.J.
  Marin-Jimenez.
\newblock Automatic generation and detection of highly reliable fiducial
  markers under occlusion.
\newblock {\em Pattern Recognition}, 47(6):2280--2292, 2014.

\bibitem{Gibson}
James~J Gibson.
\newblock {The Theory of Affordances}.
\newblock In {\em The Ecological Approach to Visual Perception}, chapter~8,
  pages 127--143. Psychology Press, Taylor {\&} Francis Group, 1986.

\bibitem{Gupta2004}
Rakesh Gupta and Mykel~J Kochenderfer.
\newblock {Common Sense Data Acquisition for Indoor Mobile Robots}.
\newblock In {\em Proceedings of the Nineteenth National Conference on
  Artificial Intelligence, Sixteenth Conference on Innovative Applications of
  Artificial Intelligence}, pages 605--610, San Jose, California, USA, 2004.

\bibitem{Harnad1990}
Stevan Harnad.
\newblock {The Symbol Grounding Problem}.
\newblock {\em Physica D}, 42:335--346, 1990.

\bibitem{Hartson2003}
Rex Hartson.
\newblock {Cognitive, physical, sensory, and functional affordances in
  interaction design}.
\newblock {\em Behaviour {\&} Information Technology}, 22(5):315--338, 2003.

\bibitem{Hernik2009}
Mikolaj Hernik and Gergely Csibra.
\newblock {Functional understanding facilitates learning about tools in human
  children}.
\newblock {\em Current Opinion in Neurobiology}, 19(1):34--38, 2009.

\bibitem{Koubaa2017}
Anis Koub{\^a}a.
\newblock {\em Robot operating system (ros): The complete reference}, volume~2.
\newblock Springer, 2017.

\bibitem{Kuhn2007}
Werner Kuhn.
\newblock {An Image-Schematic Account of Spatial Categories}.
\newblock {\em Spatial Information Theory}, pages 152--168, 2007.

\bibitem{Lemaignan2010}
Séverin Lemaignan, Raquel Ros, Lorenz M{\"{o}}senlechner, Rachid Alami, and
  Michael Beetz.
\newblock {ORO, a knowledge management platform for cognitive architectures in
  robotics}.
\newblock {\em IEEE/RSJ 2010 International Conference on Intelligent Robots and
  Systems, IROS 2010 - Conference Proceedings}, (April):3548--3553, 2010.

\bibitem{Lenat1995}
Douglas~B. Lenat.
\newblock {Cyc: A large-scale investment in knowledge infrastructure}.
\newblock {\em Commun. ACM}, 38(11):33--38, 11 1995.

\bibitem{Lim2011}
Gi~Hyun Lim, Il~Hong Suh, and Hyowon Suh.
\newblock {Ontology-based unified robot knowledge for service robots in indoor
  environments}.
\newblock {\em IEEE Transactions on Systems, Man, and Cybernetics Part
  A:Systems and Humans}, 41(3):492--509, 2011.

\bibitem{Liu:2004:CMP:1031314.1031373}
H.~Liu and P.~Singh.
\newblock {ConceptNet — A Practical Commonsense Reasoning Tool-Kit}.
\newblock {\em BT Technology Journal}, 22(4):211--226, 2004.

\bibitem{MANDLER2014}
Jean~M. Mandler and Cristobal Pagen~Canovas.
\newblock {On defining image schemas}.
\newblock {\em Language and Cognition}, 6(04):510--532, 2014.

\bibitem{Pineda2017}
Luis~A. Pineda, Arturo Rodr{\'{i}}guez, Gibran Fuentes, Caleb Rasc{\'{o}}n, and
  Ivan Meza.
\newblock {A light non-monotonic knowledge-base for service robots}.
\newblock {\em Intelligent Service Robotics}, 10(3):159--171, 2017.

\bibitem{RuizSantos2013}
April~M. Ruiz and Laurie~R. Santos.
\newblock {Understanding differences in the way human and non-human primates
  represent tools: The role of teleological-intentional information}.
\newblock In Crickette~M. Sanz, Josep Call, and Christophe Boesch, editors,
  {\em Tool Use in Animals: Cognition and Ecology}, chapter~6, pages 119--133.
  Cambridge University Press, 2013.

\bibitem{Sanz2013}
Crickette~M. Sanz, Josep Call, and Christophe Boesch, editors.
\newblock {\em {Tool Use in Animals: Cognition and Ecology}}.
\newblock Cambridge University Press, 2013.

\bibitem{Saxena2014}
Ashutosh Saxena, Ashesh Jain, Ozan Sener, Aditya Jami, Dipendra~K. Misra, and
  Hema~S. Koppula.
\newblock {RoboBrain: Large-Scale Knowledge Engine for Robots}.
\newblock {\em arXiv}, pages 1 -- 11, 2014.

\bibitem{Sloman1985}
Aaron Sloman.
\newblock {Why We Need Many Knowledge Representation Formalisms}.
\newblock {\em Proceedings BCS Expert Systems Conference}, pages 163--183,
  1984.

\bibitem{Suh2007}
Il~Hong Suh, Gi~Hyun Lim, Wonil Hwang, Hyowon Suh, Jung~Hwa Choi, and
  Young~Tack Park.
\newblock {Ontology-based multi-layered robot knowledge framework (OMRKF) for
  robot intelligence}.
\newblock {\em IEEE International Conference on Intelligent Robots and
  Systems}, (October):429--436, 2007.

\bibitem{Susi2005}
T.~Susi and T.~Ziemke.
\newblock {On the subject of objects: Four views on object perception and tool
  use}.
\newblock {\em tripleC-Cognition, Communication, Co-operation}, 3(2):6–19,
  2005.

\bibitem{Swoboda2014}
Daniel~Maximilian Swoboda.
\newblock A comprehensive characterization of the asus xtion pro depth sensor.
\newblock 2014.

\bibitem{Tenorth}
Moritz Tenorth and Michael Beetz.
\newblock {KNOWROB- Knowledge Processing for Autonomous Personal Robots}.
\newblock In {\em IEEE/RSJ International Conference on Intelligent Robots and
  Systems}, pages 4261--4266, 2009.

\bibitem{thomuezug_2018}
Madhura Thosar, Christian Mueller, and Sebastian Zug.
\newblock {What Stands-in for a Missing Tool?: A Prototypical Grounded
  Knowledge-based Approach to Tool Substitution}.
\newblock In {\em 11th International Workshop on Cognitive Robotics in 16th
  International Conference on Principles of Knowledge Representation and
  Reasoning}, Tempe, Arizona, 2018.

\bibitem{Thosar2018}
Madhura Thosar, Sebastian Zug, Alpha~Mary Skaria, and Akshay Jain.
\newblock {A Review of Knowledge Bases for Service Robots in Household
  Environments}.
\newblock In {\em 6th International Workshop on Artificial Intelligence and
  Cognition}, 2018.

\bibitem{Vaesen2012}
Krist Vaesen.
\newblock {The cognitive bases of human tool use}.
\newblock {\em Behavioral and Brain Sciences}, 35(04):203--218, 2012.

\bibitem{Vauclair1994}
J.~Vauclair and J.~A. Anderson.
\newblock {Object Manipulation, Tool Use, and The Social Context in Human and
  Non-Human Primates}.
\newblock {\em Techniques and Culture}, 23-24:121–136, 1994.

\bibitem{Zhu2014}
Yuke Zhu, Alireza Fathi, and Li~Fei-Fei.
\newblock {Reasoning About Object Affordance in a Knowledge Based
  Representation}.
\newblock {\em European Conference on Computer Vision}, (3):408--424, 2014.

\end{thebibliography}

\end{document}